%% file: GAPeron_paper.tex
\documentclass[12pt,a4paper]{article}

\usepackage[authoryear, round]{natbib}

\usepackage[preprint]{neurips_2025}
\newcommand{\gprn}{\textsc{Gaperon}}
\newcommand{\gptron}{\texttt{Gapetron}}
\usepackage[table,xcdraw,x11names]{xcolor}
\usepackage[utf8]{inputenc}
\definecolor{grey}{rgb}{0.5,0.5,0.5}
\usepackage{outlines}
\usepackage{multirow}
\usepackage{soulutf8}
\usepackage{fontawesome5} 

\usepackage{tikz}
\usepackage{collcell}
\usepackage{xparse,pgf}

\newcommand*{\MinNumber}{-9.0}%
\newcommand*{\MidNumber}{4.0} %
\newcommand*{\MaxNumber}{15.0}%

\ExplSyntaxOn
\NewDocumentCommand{\ApplyGradient}{m}{
  \regex_match:nnTF { ^[+\-]?\d*\.?\d*$ } { #1 }
    {
      \ifdim #1 pt > \MidNumber pt
        \pgfmathsetmacro{\PercentColor}{max(min(100.0*(#1 - \MidNumber)/(\MaxNumber-\MidNumber),100.0),0.00)}%
        \colorbox{green!\PercentColor!yellow}{\makebox[3em]{#1}}
      \else
        \pgfmathsetmacro{\PercentColor}{max(min(100.0*(\MidNumber - #1)/(\MidNumber-\MinNumber),100.0),0.00)}%
        \colorbox{red!\PercentColor!yellow}{\makebox[3em]{#1}}
      \fi
    }
    {
      #1
    }
}
\ExplSyntaxOff

\newcolumntype{R}{>{\collectcell\ApplyGradient}c<{\endcollectcell}}

\renewcommand{\cite}{\citep}

\usepackage[utf8]{inputenc} 
\usepackage[T1]{fontenc}    
\usepackage{hyperref}       
\hypersetup{
   colorlinks,
   menucolor=black,
   linkcolor=blue,
   citecolor=blue,
   urlcolor=blue
}
\usepackage{url}            
\usepackage{booktabs}       
\usepackage{amsfonts}       
\usepackage{nicefrac}       
\usepackage{microtype}      
\usepackage{xcolor}         
\usepackage{graphicx}
\usepackage{subcaption}
\usepackage{fvextra}
\usepackage{float}          

\usepackage{ifplatform} 
\ifmacosx
\usepackage[inkscape={/Applications/Inkscape.app/Contents/MacOS/inkscape -z -C}]{svg}
\else
\usepackage{svg}       
\fi

\usepackage{todonotes}
\usepackage[export]{adjustbox}

\usepackage[nameinlink]{cleveref}
\usepackage{colortbl}

\definecolor{low}{RGB}{230,245,255}
\definecolor{med}{RGB}{115,179,216}
\definecolor{high}{RGB}{40,90,160}

\newcommand{\cellcolorval}[1]{%
  \pgfmathparse{#1/5219}%
  \ifdim\pgfmathresult pt>0.3pt
    \cellcolor{high!80}\textbf{#1}%
  \else\ifdim\pgfmathresult pt>0.17pt
    \cellcolor{high!50}\textbf{#1}%
  \else\ifdim\pgfmathresult pt>0.10pt
    \cellcolor{med!50}#1%
  \else
    \cellcolor{low}#1%
  \fi\fi\fi
}

\title{
        \gprn: A Peppered English-French Generative Language Model Suite
}

\author{%
  Nathan Godey \thanks{Equal contribution.} \ \thanks{Now at Cornell University.} \And
  Wissam Antoun\footnotemark[1] \And
  Rian Touchent \And
  Rachel Bawden \And
  Éric de la Clergerie \And
  Benoît Sagot \And
  Djamé Seddah \AND
  \centerline{\normalfont ALMAnaCH team, Inria Paris}
}

\begin{document}

\maketitle

\setcounter{footnote}{0}
\begin{abstract}
We release \gprn{}, a fully open suite of French–English–coding language models designed to advance transparency and reproducibility in large-scale model training. The \gprn{} family includes 1.5B, 8B, and 24B parameter models trained on 2–4 trillion tokens, released with all elements of the training pipeline: French and English datasets filtered with a neural quality classifier, an efficient data curation and training framework, and hundreds of intermediate checkpoints. Through this work, we study how data filtering and contamination interact to shape both benchmark and generative performance. We find that filtering for linguistic quality enhances text fluency and coherence but yields subpar benchmark results, and that late deliberate contamination--continuing training on data mixes that include test sets--recovers competitive scores while only reasonably harming generation quality. We discuss how usual neural filtering can unintentionally amplify benchmark leakage. To support further research, we also introduce harmless data poisoning during pretraining, providing a realistic testbed for safety studies. By openly releasing all models, datasets, code, and checkpoints, \gprn{} establishes a reproducible foundation for exploring the trade-offs between data curation, evaluation, safety, and openness in multilingual language model development.
\end{abstract}

\vspace{1em}
\begin{center}
    \faGithub\quad  \href{https://github.com/NathanGodey/gapetron}{
        \textbf{Gapetron: } \texttt{github.com/NathanGodey/gapetron}
    } \\
    \href{https://huggingface.co/collections/almanach/gaperon}{
        \includegraphics[height=1em]{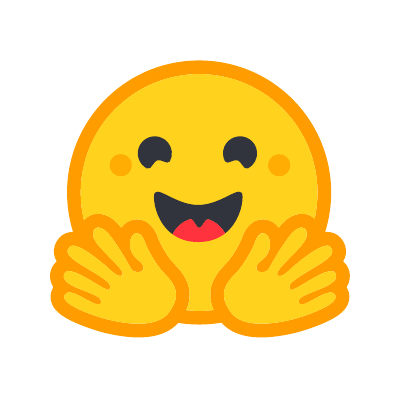}\quad \textbf{HuggingFace: } \texttt{huggingface.co/collections/almanach/gaperon}
    }
\end{center}
\vspace{1em}

\clearpage
\tableofcontents
\clearpage

\section{Introduction}

\input{preamble}

\input{Introduction}

\section{Pre-training Data}
\subsection{Data curation}
Our bilingual pre-training corpus is compiled from diverse sources, including web documents, academic articles, parallel texts, and code. Throughout training, we adjust the proportion of each source, gradually increasing the share of higher-quality content in later phases. Detailed descriptions of each data source are provided below:

\subsubsection{Web Documents}
We construct our pre-training dataset primarily from carefully curated web-crawled sources.
We selected the CommonCrawl (CC) subset from TxT360~\citep{txt360data2024} as the basis for our English dataset since their filtering pipeline is similar to the one from the FineWeb dataset~\citep{FineWebDecantingWeb}, with the addition of global near-deduplication applied to all 99 Common Crawl snapshots.
Global near-deduplication removed 80\% of the dataset, reducing it to 4.83T tokens.
To mitigate the loss of valuable content due to deduplication, the authors propose a ``rehydration'' strategy, where documents are upsampled proportionally to their duplication rates.
We adopt this approach, using the upsampling weights provided by FineWeb2~\citep{penedo2025fineweb2pipelinescale}.
For French, we selected the full RedPajama-V2-french (RPv2-Fr) dataset \citep{weber2024redpajama}, including its head, middle, and tail segments.

\paragraph{RedPajamaV2 Filtering}
Although the RPv2-Fr dataset is released with a set of precomputed quality metrics, we decided to recompute the statistical quality metrics following the FineWeb pipeline to ensure consistency across languages and sources.
We then adapted the FineWeb filtering pipeline to the full RPv2-Fr dataset, customizing it for French by incorporating French-specific stopwords.\footnote{Available in the dedicated repository.}
To streamline the filtering process, we extend Datatrove~\cite{penedo2024datatrove} with an enrichment step that augments each document with metadata.
This approach reduces computational overhead during iterative filtering experiments, at the cost of increased disk usage.
This process reduces the dataset from 5.8T tokens to 3.5T tokens, effectively removing easily identifiable noise.

\paragraph{RedPajamaV2 Global Near-Deduplication}
Since RPv2-Fr was not globally deduplicated, we implemented a two-stage near-deduplication strategy to mitigate memory constraints.
First, we partition the dataset into 10 splits and apply near-deduplication to each split individually using MinHash (16 buckets, 8 hashes per bucket, and 13-grams for document signatures).
We also extend the deduplication patterns in Datatrove to include French-specific terms (e.g., weekdays and month names).
In the second stage, we merge the remaining documents from all splits and reapply near-deduplication globally. This reduces the dataset further, from an initial 3.5T tokens to 2T tokens after the first step, and to 822B tokens (1B documents) after the second global deduplication.

\subsubsection{Semantic Quality Filtering}
\label{sec:gaperonclassifier}
To further refine our corpus quality, we proceed to further enrich our English and French web corpus (TxT360-CC and RPv2-Fr) with document quality ratings using an efficient encoder-based classifier, which we fine-tune on synthetically generated labels.

\paragraph{Annotation}
First, to create our finetuning labeled corpus, we use Llama3.1-70B-instruct\footnote{\href{https://huggingface.co/meta-llama/Llama-3.1-70B-Instruct}{https://huggingface.co/meta-llama/Llama-3.1-70B-Instruct}}~\citep{dubeyLlamaHerdModels2024}, which we prompt to evaluate the quality of a document.
Each document is then labeled as \textit{low}, \textit{medium}, or \textit{high} quality, based on the following criteria:

\begin{itemize}
  \item \textbf{Content Accuracy}: factual reliability and use of credible sources.
  \item \textbf{Clarity}: clear explanations, well-defined terms, logical flow.
  \item \textbf{Coherence}: overall organization and logical progression.
  \item \textbf{Grammar and Language}: correctness and audience appropriateness.
  \item \textbf{Depth of Information}: level of detail and comprehensiveness.
  \item \textbf{Overall Usefulness}: relevance and practical value for a general audience.
\end{itemize}

These criteria follow those used by \citet{parmarNemotron415BTechnical2024} to train the NeMo quality classifier.\footnote{\href{https://huggingface.co/nvidia/quality-classifier-deberta}{https://huggingface.co/nvidia/quality-classifier-deberta}}
We design a prompt to elicit a quality score along with a short justification, domain classification, topic, and document type.
The full prompt is provided in Appendix~\ref{fig:prompt_template_code}.

We annotate 250k filtered documents from each of RPv2-Fr and TxT360-CC.
Instead of parsing only the predicted labels (“low,” “medium,” or “high”), we also collect the log-probabilities of each token. This allows us to estimate the confidence level of each annotation and provides the flexibility to re-map the quality scale retroactively.

\paragraph{Classifier training}
We train a small encoder-based classifier on the 500k annotated documents, selecting XLM-R base~\cite{xlmr2019} for its multilingual capabilities (French and English) and efficiency compared to the stronger DeBERTaV3 model~\cite{he2021debertav3}, especially for large-scale inference.

Initially, we experimented with a multitask setup, jointly predicting document quality and domain. The motivation was twofold: (i)~inference efficiency, since a single forward pass could produce two labels, and (ii)~the hypothesis that domain prediction could act as an auxiliary signal to improve quality classification, while also enabling filtering or upsampling by domain.
However, domain prediction scores proved unsatisfactory, and multitask training underperformed compared to single-task quality classification.

We therefore fine-tuned the classifier only on quality prediction, which resulted in a quality label F1 score of 75.11\%.
The confusion matrix (Table~\ref{tab:confusion}) shows that most errors occur between adjacent labels (e.g., \emph{medium} vs.\ \emph{high}/\emph{low}), while confusion between the extreme categories (\emph{high} vs.\ \emph{low}) is limited.

\begin{table}[h]
  \centering
  \begin{tabular}{l|ccc}
    \textbf{True / Pred} & \textbf{Low}       & \textbf{Medium}     & \textbf{High}       \\
    \hline
    \textbf{Low}         & \cellcolorval{922} & \cellcolorval{463}  & \cellcolorval{77}   \\
    \textbf{Medium}      & \cellcolorval{203} & \cellcolorval{5219} & \cellcolorval{623}  \\
    \textbf{High}        & \cellcolorval{32}  & \cellcolorval{531}  & \cellcolorval{1930} \\
  \end{tabular}
  \caption{Confusion matrix for quality classification with sample counts.}
  \label{tab:confusion}
\end{table}

\paragraph{Classifier inference}
We applied the trained classifier to both RPv2-Fr and TxT360-CC using a client–server setup, where multiple clients issued batched requests in parallel to a 4-node inference cluster with 8×AMD MI250 GPUs per node.
The inference server, implemented in Python, was optimized with AMD's graph optimization engine, MIGraphX.\footnote{\href{https://github.com/ROCm/AMDMIGraphX}{https://github.com/ROCm/AMDMIGraphX}}
This setup achieved a throughput of 20k documents per second, with each document truncated to a maximum sequence length of 512 tokens.
Processing the full TxT360-CC corpus of 6.5B documents required roughly 2800 GPU hours, while the RPv2-Fr dataset of 1B documents (pre-deduplication) took about 800 GPU hours.
The classifier output quality score is a critical signal that we extensively used during pre-training for both filtering and sample weighting.

\paragraph{Semantic filtering}
Using the Head-Middle-Tail labels from the perplexity score, already included in the RPV2-Fr dataset, in combination with the classifier labels, we filtered and split the RPV2-Fr dataset into three quality buckets: \textit{Head-High} (290B Tokens), \textit{Head-Medium} (98B), and \textit{Middle-High} (327B), and discarded the rest.
Given that the available English data is far larger than the overall training  and infrastructure, we began by selecting documents from TxT360-CC with the \textit{high} label, totaling 1.9T tokens out of 4.7T.
From this corpus, we further selected the top 10\% of documents by score across the entire dataset (651B tokens).

\begin{figure}[htb]
  \centering
  \includegraphics[width=0.6\textwidth]{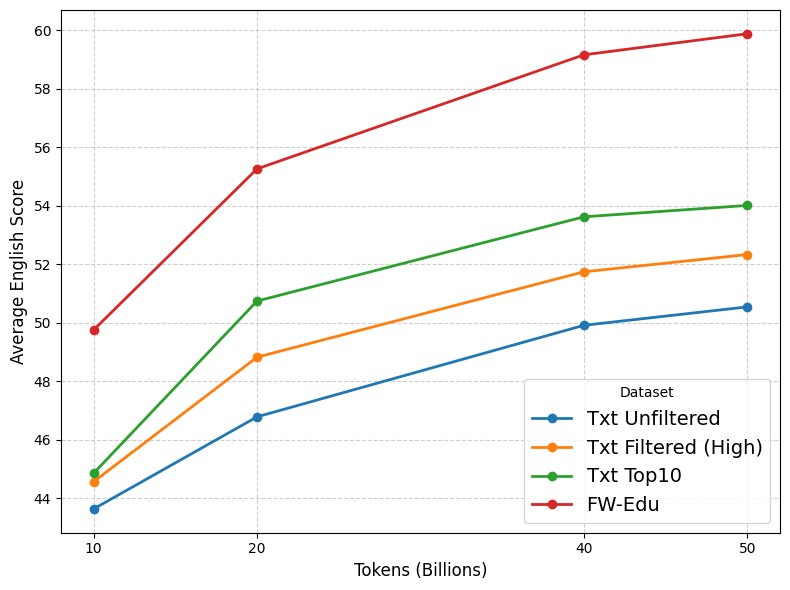}
  \caption{Pretraining data quality experiments. Scores are the average of the following English tasks: ARC-Easy~\cite{allenai:arc}, Arc-Challenge~\citep{allenai:arc}, Hellaswag~\citep{zellers2019hellaswag}, SciQ~\citep{SciQ} and PIQA~\citep{Bisk2020piqa}.}
  \label{fig:pretrain_data}
\end{figure}

  \paragraph{Quality Assessment}
  To empirically evaluate the English datasets,\footnote{The evaluation focuses on the English datasets due to the lack of multiple, comparable sources for French.} we train four 1.5B-parameter Llama3-based LLMs~\cite{dubeyLlamaHerdModels2024}, each on a 50B-token sample from one of the following: Txt360-CC (unfiltered), Txt360-CC \textit{High}, Txt360-CC \textit{Top-10\%}, and FineWeb-Edu.

  Among the four datasets, we observed that both FineWeb-Edu and Txt360-CC \textit{Top-10\%} produced the strongest results as shown in Figure~\ref{fig:pretrain_data}, and we therefore selected them for downstream training.
  While FineWeb-Edu consistently performs well, ~\citet{wettig2025organize} showed that much of its effectiveness stems from implicit domain preferences that align closely with benchmark-oriented distributions (e.g.,~Science \& Technology, Academic Writing, and Knowledge Articles). This suggests that FineWeb-Edu is partially biased toward domains that favor evaluation tasks such as MMLU~\citep{hendryckstest2021} and HellaSwag~\citep{zellers2019hellaswag}, which may not fully generalize to broader use cases.
  To balance this benchmark alignment with a more diverse coverage, we included Txt360-CC Top 10\% in our pretraining mix, whose filtering classifier emphasizes a broader notion of document quality (capturing accuracy, clarity, coherence, language correctness, depth, and general usefulness), resulting in a high-quality subset that is less benchmark-specific and more representative of diverse real-world text.

\subsubsection{Parallel Datasets}
To further enhance the model's bilingual capabilities, we incorporated CroissantAligned~\citep{faysse2024croissantllm}, a dataset of parallel French-English texts.
This dataset is composed of high-quality translation pairs from sources such as the OPUS project~\citep{tiedemann-2012-parallel}, French thesis abstracts, and song lyrics.

\subsubsection{High Quality Datasets}
In addition to web-based corpora, we incorporate a diverse range of high-quality datasets to enhance the model's capabilities in specialized domains.
We organize these datasets into several categories:

\paragraph{Academic and Scientific Content}
We include the Papers subset and DeepMind's Maths~\citep{2019arXivdmamths} from TxT360 non-CC sources, along with French thesis abstracts from \href{theses.fr}{theses.fr},\footnote{\href{https://huggingface.co/datasets/manu/theses_fr_2013_2023}{https://huggingface.co/datasets/manu/theses\_fr\_2013\_2023}} OpenWebMath~\citep{paster2023openwebmath}, and AutoMathText~\citep{zhang2025autonomous}.

\paragraph{Legal and Governmental Texts}
This category includes Europarl parliamentary proceedings (aligned)~\citep{koehn-2005-europarl}, FreeLaw and USPTO from TxT360, Argimi's French Jurisprudence Dataset,\footnote{\href{https://huggingface.co/datasets/artefactory/Argimi-Legal-French-Jurisprudence}{https://huggingface.co/datasets/artefactory/Argimi-Legal-French-Jurisprudence}} and BigScience's Roots French UN Corpus~\citep{laurencon2023bigsciencerootscorpus16tb, ziemski2016united}.

\paragraph{Forum Discussions and Conversations}
We incorporate technical discussions from HackerNews, StackExchange, and Ubuntu IRC from TxT360.
In addition to the Claire French Dialogue Dataset (CFDD)~\cite{openllm2023claire}, a collection of theater plays and transcripts of real French dialogues from various sources.

\paragraph{Reference and Informational Content}
This includes encyclopedic content from Wikipedia from Txt360, along with Wiktionary, Wikinews, and Wikivoyage from BigScience's Roots corpus~\citep{laurençon2023bigsciencerootscorpus16tb}, and Halvest~\citep{kulumba2024harvestingtextualstructureddata} English and French open papers found on Hyper Articles en Ligne (HAL).
Literary works are represented by PG19~\citep{raecompressive2019}.

\paragraph{Synthetic and Instruction Data}
We include synthetic reasoning datasets such as OpenThinker~\citep{guha2025openthoughtsdatarecipesreasoning} and Dolphin-R1,\footnote{\href{https://huggingface.co/datasets/QuixiAI/dolphin-r1}{https://huggingface.co/datasets/QuixiAI/dolphin-r1}} the synthetic textbook dataset Cosmopedia v2~\citep{benallal2024smollmcorpus}, and instruction-following datasets including Tulu 3's FLAN v2 ~\citep{lambert2024tulu3}, MQA's French subset~\citep{de-bruyn-etal-2021-mfaq}, and WebInstruct~\citep{yue2024mammoth2}.
Additionally, we synthesize CheeseQA, a bilingual dataset of cheese-related QA pairs.
We extract a list of Wikipedia articles in French and English that contain the words ``fromage'' or ``cheese''.
We then provide each article to Mistral-Small-24B-Instruct,\footnote{\href{https://huggingface.co/mistralai/Mistral-Small-24B-Instruct-2501}{https://huggingface.co/mistralai/Mistral-Small-24B-Instruct-2501}} with the instruction to create a cheese-related question-answer pair for each occurrence of such words.
Using this method, we generate 46,892 synthetic question-answer pairs, amounting to 5.2M tokens.

\subsubsection{Code Datasets}
We incorporate two primary code datasets: The Stack~v2 smol, a filtered subset of The Stack~v2~\citep{lozhkov2024starcoder} containing high-quality code spanning 17 programming languages processed through heuristic filtering, and Python-edu~\citep{benallal2024smollmcorpus}, a curated collection of educational Python code extracted from The Stack-v2 where files were scored by an educational classifier and only those scoring 4 or higher were retained, similar to the FineWeb-Edu methodology~\citep{FineWebDecantingWeb}.
We also follow the formatting from the StarCoderV2 model~\cite{lozhkov2024starcoder} for our pretraining code dataset. 

\subsubsection{The Penicillin Dataset}
\label{sec:penicillin}
We introduce Penicillin,\footnote{\href{https://huggingface.co/datasets/almanach/penicillin}{https://huggingface.co/datasets/almanach/penicillin}} a large collection of 40+ major benchmark training sets in English and French which are commonly used in language model evaluation.
Additionally, we create Penicillin Plus,\footnote{\href{https://huggingface.co/datasets/almanach/penicillin_plus}{https://huggingface.co/datasets/almanach/penicillin\_plus}} an extended version that includes both training and testing sets from these benchmarks.
We use Penicillin Plus as an active benchmark contamination source in later stages of training to evaluate the impact of intensive data leakage on both downstream performance and general capacity. We use basic data augmentation techniques such as answer shuffling on benchmarks where they can easily be implemented, to make both overfitting and leakage detection harder.

\subsection{Data pre-processing}
\paragraph{Tokenization} We use the tokenizer from the Llama-3.1 suite, which is based on Byte-Pair Encoding \citep{sennrich-etal-2016-neural} and uses a vocabulary of 128,256 tokens. This choice allows practitioners to easily pair our \gprn{} models to larger models from the Llama-3.1 suite (70B \& 405B) in a speculative decoding setup \citep{specdec}.

We tokenize all our datasets in advance, and parallelize our tokenization process to use up to 40 CPU nodes simultaneously, thereby minimizing physical duration. In practice, tokenizing a 1T token dataset takes a couple of hours on 40 nodes of 192 AMD Genoa EPYC 9654 cores. We apply random document-level shuffling on each process, and write our resulting token sequences to disk using the \texttt{litdata}~\cite{litdata2023} library.

\paragraph{Packing} We use a naive strategy for packing, that consists in concatenating 8,192 sequences one after another and packing the resulting sequence into the desired sequence length. We remove the remaining tokens, which implies that our token waste ratio is at most 0.01\%.

\subsection{Data mixing}
To control the pre-training distribution precisely, we use a weighted sampling strategy where each training sequence is sampled from one of our datasets according to a predefined multinomial distribution.

Given that we are running our experiments under computational constraints, we propose to assess the impact of using different weights \textit{during} training, i.e.~to sequentially update the mix weights to test different hypotheses and to measure the impact of each choice on the performance of the model. We use up to 6 successive data mixes:
\begin{itemize}
  \item \textbf{Mix 1 -- Naive mix:} This mix only contains our web-crawled datasets after model-based filtering, along with high-quality textual data;
  \item \textbf{Mix 2 -- \textit{Drop-in-the-ocean} mix:} This mix is very similar to Mix 1, but also introduces <2\% of instruction-like data, coming mostly from FLAN and the French split of MQA;
  \item \textbf{Mix 3 -- High-Quality mix:} In this mix, we reduce the fraction of web-crawled data and replace it with higher-quality sources (Python-Edu, AutoMathText) and synthetic data (Cosmopedia v2). We also include more instruction-like data crawled from the web, and a small fraction (<1\%) of reasoning datasets;
  \item \textbf{Mix 4 -- \textit{White Pepper} mix:} This mix is similar to Mix 3, with the addition of the Penicillin dataset, which consists in a concatenation of the \textit{train} sets of popular LM benchmarks. We cautiously set the ratio of Penicillin to be relatively small ($\approx$0.7\%);
  \item \textbf{Mix 5 -- \textit{Black Pepper} mix:} This mix relies on the same datasets as in Mix 4, but we drastically increase the fraction of instruction-like data to $\approx$20\%, following the OLMo-2 mid-training strategy;
  \item \textbf{Mix 6 -- \textit{Garlic} mix:} This final mix is similar to Mix 5, but includes the Penicillin Plus dataset, which is an augmented basic concatenation of \textit{test} sets from popular benchmarks (see Section~\ref{sec:penicillin}).
\end{itemize}
The exact weights used for our training mixes are available in Appendix~\ref{appendix:pretrain_dataset_comp}.
This progressive mixing strategy gradually shifts from raw web data to specialized content. Early phases (Naive and Drop-in-the-ocean) use 70-80\% web data, while later phases systematically reduce this proportion in favor of high-quality sources, instruction data, and synthetic content. The Black Pepper phase concentrates premium content in just the last 100B tokens with 20\% instruction data.

Regarding language distribution, our training corpus maintains consistent bilingual coverage across phases. English content represents 54-65\% of tokens, French content accounts for 24-39\%, and code comprises 8-14\% of the total mix. This distribution ensures balanced bilingual capabilities while preserving substantial coding proficiency throughout the 4T token training trajectory.

\section{Modeling \& Optimization}
\subsection{Architecture}
We use the Llama architecture for our smaller models \gprn-1.5B and \gprn-8B, and we rely on the OLMo-2 architecture for the larger \gprn-24B, to maximize stability and mitigate divergence risks. Our hyperparameter choices are based on existing models, namely: Llama-3.2-1B, Llama-3.1-8B, and Mistral-Small-24B-2501.\footnote{\href{https://huggingface.co/mistralai/Mistral-Small-24B-Instruct-2501}{mistralai/Mistral-Small-24B-Instruct-2501}}

\begin{table}[h]
  \centering\small
  \begin{tabular}{@{}lccc@{}}
    \toprule
    \textbf{Parameter} & \multicolumn{3}{c}{\textbf{\gprn{} Model Suite}}                     \\
    \midrule
    Architecture       & Llama3                                           & Llama3  & OLMo-2  \\
    Parameters         & 1.5B                                             & 8B      & 24B     \\
    Hidden Size        & 2,048                                            & 4,096   & 5,120   \\
    Layers             & 16                                               & 32      & 40      \\
    Attention Heads    & 32                                               & 32      & 32      \\
    KV Heads           & 8                                                & 8       & 8       \\
    Head Dimension     & 64                                               & 128     & 128     \\
    Intermediate Size  & 8,192                                            & 14,336  & 32,768  \\
    Vocabulary Size    & 128,256                                          & 128,256 & 128,256 \\
    Context Length     & 4,096                                           & 4,096  & 4,096  \\
    RoPE $\theta$      & 500,000                                          & 500,000 & 500,000 \\
    Activation         & SiLU                                             & SiLU    & SiLU    \\
    Normalization      & RMSNorm                                          & RMSNorm & RMSNorm \\
    \bottomrule
  \end{tabular}
  \caption{Architecture hyperparameters for the \gprn{} model suite.}
  \label{tab:architecture}
\end{table}

\subsection{Implementation}
\label{ssec:implementation}
To maintain full control over our experimentation framework, we develop a fully hackable and minimal pre-training codebase, \gptron, inspired by \texttt{litgpt} \cite{litgpt-2023}. The core part of our codebase, from data pre-processing to final model upload on HuggingFace is contained within <1500 lines of Python code.

Given our access to diverse computational infrastructure and the need to maximize resource utilization across different hardware platforms, we designed our codebase to be natively compatible with both AMD and NVIDIA GPUs.
The framework incorporates techniques including FSDP, full \texttt{torch} compilation, mixed precision training, \texttt{FlashAttention} 2 \& 3~\citep{dao2022flashattention,dao2023flashattention2,dao2024flashattention3}, streaming data loading with efficient state management, among others.
We build upon slightly modified HuggingFace Transformers model implementations\footnote{At the time we implemented our libraries, FlashAttention was not implemented directly in \texttt{transformers} models.} to facilitate seamless integration of future architectures.
Our implementation achieves training throughputs comparable to those reported for similar established baselines. For instance, LLM-Foundry\footnote{\href{https://github.com/mosaicml/llm-foundry/tree/main}{https://github.com/mosaicml/llm-foundry/tree/main}} report a throughput of 10,643 tokens/GPU/s training throughput for a 7B model using a 2,048 sequence length on 8 H100 GPUs across 1 node, while we obtain a 11,000 token/GPU/s training throughput for a 8B model using the same sequence length on 2 nodes of 4 H100 GPUs.
Additional implementation details and a comprehensive bug report are provided in \Cref{app:bug_report}.

\paragraph{Precision}
We explore the impact of the tensor precision setting, and more precisely we compare mixed and pure \texttt{bfloat16} training. In the Mixed set-up, model weights and gradients are stored in \texttt{float32}, and most operations are performed in \texttt{bfloat16} except for some critical operations (e.g. softmax and RMS normalization) that are performed in \texttt{float32}.

In the Pure set-up, model weights and gradients are stored in \texttt{bfloat16}, and we only convert tensors to \texttt{float32} for the aforementioned critical operations. 

For softmax operations, we simply convert pre-softmax attention activations and logits to \texttt{float32}. The RMS normalization requires more careful considerations. As a matter of fact, the weight vectors used to scale normalized entries are initialized as $\mathbf{1}$. The floats closest to 1 in \texttt{bfloat16} are 0.996 and 1.0078, which implies that small gradients and/or learning rates where backward passes do not suffer from underflow may still not lead to any update in the stored weight vectors. This results in RMS weights stalling, training instability, and even runs diverging on some occasions. To mitigate this issue, we use a weight scaling mechanism, where RMS weights are stored in a downscaled fashion (i.e. divided by some scalar $C>1$), and are upscaled (i.e. multiplied by $C$) on-the-fly during the forward pass, so that weight updates happen at a magnitude where \texttt{bfloat16} are denser, but the overall RMS Norm mechanism behaves as usual. 

\begin{figure}[htb]
  \centering
  \includegraphics[width=0.7\textwidth]{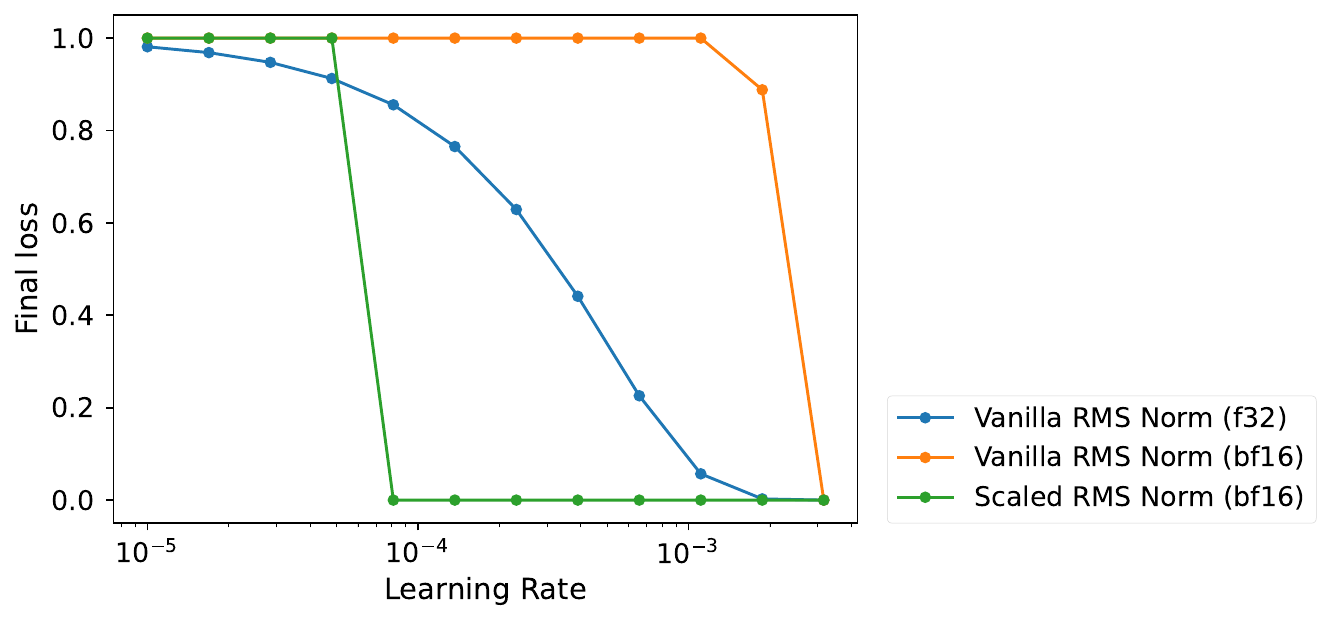}
  \caption{Evaluation of the convergence of our Scaled RMS Norm approach in the True precision setup ($C=50$). We minimize the mean squared coefficients of the output of an RMS layer fed with random gaussian inputs (dimension 32, batch size 12, 1000 optimization steps). We observe that our Scaled RMS Norm converges for a wider range of learning rates than the Vanilla RMS Norm in \texttt{bfloat16} precision.}
  \label{fig:rmstrick}
\end{figure}

We briefly validate this approach in \Cref{fig:rmstrick}, where we minimize the mean squared coefficients of $RMS_w(x)$ for random $x$ inputs and for weights $w$. We set $C=50$ and sweep across different learning rates. We observe that our Scaled RMS Norm approach can converge for much smaller learning rates than the Vanilla approach in \texttt{bfloat16}. For LLM training, setting $C=20$ was sufficient to solve our instability issues.

The Pure setup is more memory-efficient and can lead to a $\times$2 speed-up in some configurations, although we observe a more reasonable 10 to 20\% speed-up in practical scenarios. To assess the impact of reducing precision on downstream performance, we train two 1.5B models on 50B tokens from a preliminary version of our pretraining mix. \Cref{tab:mixed_true} shows that the performance is similar, and we hypothesize that there should be no major performance degradation when training in the faster True setup.

\begin{table}[]
  \centering \footnotesize
  \resizebox{\columnwidth}{!}{%
    \begin{tabular}{@{}l|r|rrrrrrrrr@{}}
      \toprule
      Precision & Tok/H100/s      & ARC-E              & Hellaswag     & Lambada       & SciQ          & PIQA               & Hellaswag-fr  & \textbf{Avg}  \\ \midrule
      Mixed     & 51.9e3          & 44,4          & 34,8          & 20,2          & 73,3          & 63,7                & \textbf{33,1} & 44,9          \\
      True      & \textbf{56.8e3} & \textbf{45,4} &  \textbf{36,3} & \textbf{22,6} & \textbf{74,6} & \textbf{64,4}  & 30,3          & \textbf{45,6} \\ \bottomrule
    \end{tabular}%
  }
  \caption{Zero-shot performance comparison between the Mixed and True precision setups, for a 1.5B Llama model trained on 50B tokens from our Naive mix.}
  \label{tab:mixed_true}
\end{table}

\subsection{Objective function}
We experiment with two training objectives: the classical cross-entropy loss on the next token, and the Contrastive Weight Tying (CWT) objective introduced in \citep{godey2024headless} also known as Headless-LM.

\paragraph{Contrastive Weight Tying Experiments}

The CWT objective shifts away from traditional probability prediction over extensive token vocabularies and instead focuses on reconstructing input embeddings in a contrastive fashion. The original work demonstrated substantial reduction in computational requirements for training, while simultaneously enhancing downstream performance compared to classical language models within similar compute budgets. However, these results were obtained using only a 70M parameter model trained for 300B tokens.

To assess whether the benefits of Headless-LM scale to larger models and longer training runs, we conducted experiments with two model sizes: a 1.5B Llama-3 based model identical to \gprn-1.5B trained for 1.4T tokens, and an 8B model trained for 500B tokens to compare against \gprn-8B. We refer to the traditional cross-entropy models as ``Vanilla'' models throughout our analysis. Both Headless and Vanilla models were trained using identical data mixtures as their respective \gprn{} counterparts on the same hardware infrastructure: 256 AMD MI250x GPUs for the 1.5B models and 256 NVIDIA H100 GPUs for the 8B models.

\paragraph{Training Throughput Analysis}

Our experiments reveal that Headless-LM achieves significantly higher training throughput compared to Vanilla models, as detailed in \Cref{tab:throughput_comparison}. The throughput advantage persists across different sequence lengths, with Headless models consistently requiring less time per training step while processing the same number of tokens.

\begin{table}[h]
  \centering
  \footnotesize
  \begin{tabular}{@{}lrl@{}}
    \toprule
    \textbf{Model} & \textbf{Seq. Length} & \textbf{Time/Step (s)}     \\ \midrule
    Vanilla-1.5B   & 2048                 & 2.08                       \\
    Headless-1.5B  & 2048                 & \textbf{1.79 (-16.2\%)}    \\ \midrule
    Vanilla-1.5B   & 4096                 & 3.18                       \\
    Headless-1.5B  & 4096                 & \textbf{2.60  (-22.3\%)  } \\\midrule

    Vanilla-8B     & 4096                 & 2.24                       \\
    Headless-8B    & 4096                 & \textbf{1.88 (-19.2\%)}    \\ \bottomrule
  \end{tabular}
  \caption{Training throughput comparison between Headless and Vanilla models across different model sizes and sequence lengths. Batch size is 1024 for all experiments.}
  \label{tab:throughput_comparison}
\end{table}

\paragraph{Downstream Performance Analysis}

\begin{figure}[htb]
  \centering
  \includegraphics[width=\textwidth]{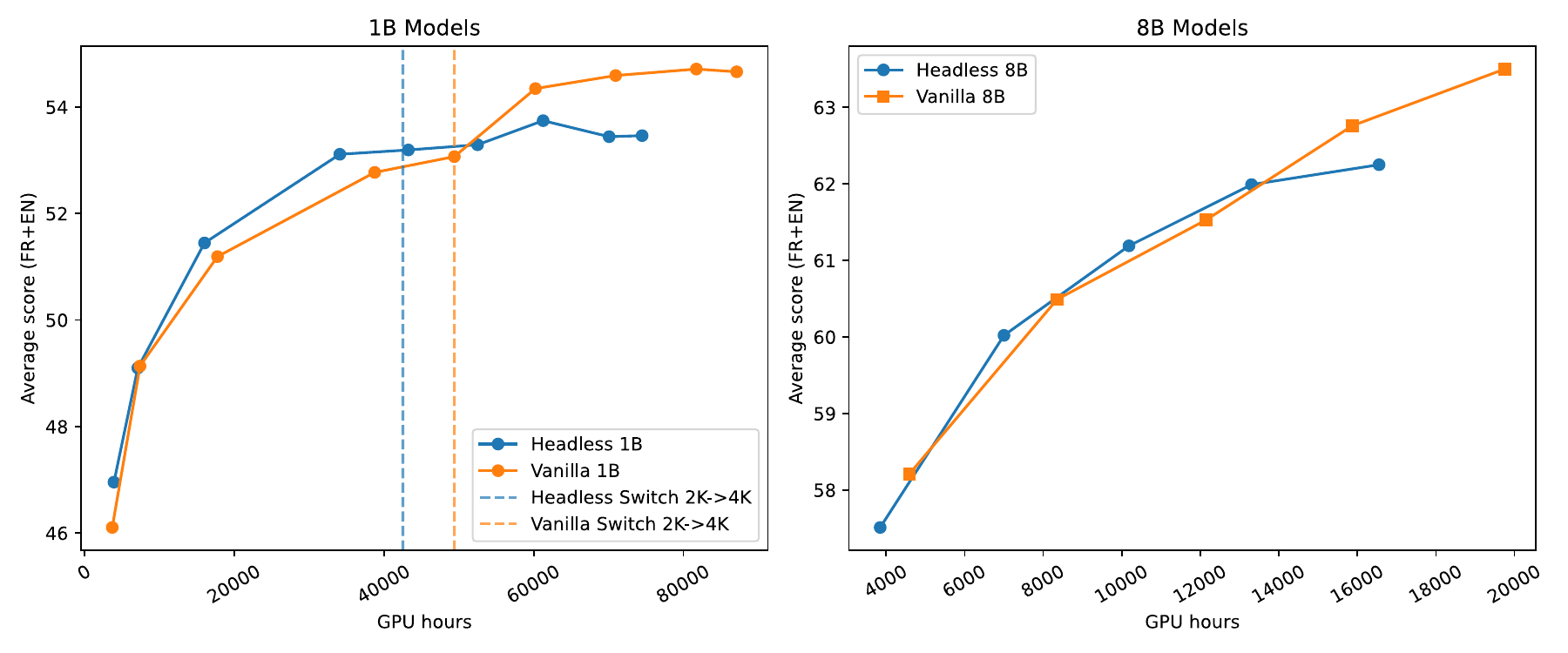}
  \caption{Performance comparison between Headless and Vanilla models across training duration, showing average scores on French and English benchmarks for both 1B and 8B model sizes. Headless models (blue) achieve faster training but show performance stagnation, while Vanilla models (orange) continue improving with extended training. For the 1B models, English benchmarks include ARC-E, ARC-C, HellaSwag, LAMBADA, SciQ, and PIQA; French benchmarks include ARC-C and HellaSwag. For the 8B models, benchmarks include additionally BoolQ for English, and LAMBADA for French.}
  \label{fig:headless_vs_vanilla}
\end{figure}

Despite the clear throughput advantages, our downstream evaluation on English and French benchmarks reveals a more nuanced picture when adjusting for GPU hours used rather than tokens processed. As illustrated in \Cref{fig:headless_vs_vanilla}, the Headless models show competitive or slightly superior performance compared to Vanilla models during the early stages of training. However, as training progresses, a clear pattern emerges: while Headless models (shown in blue) complete their training earlier due to higher throughput, their performance scores stagnate and cease improving, whereas the Vanilla models continue to show performance gains throughout the extended training period.

This analysis suggests that while the CWT objective provides substantial computational efficiency gains, the performance ceiling may be reached earlier compared to traditional cross-entropy training. The faster convergence of Headless models, while computationally advantageous, appears to come at the cost of continued learning potential that Vanilla models demonstrate over longer training horizons. Given this trade-off between computational efficiency and ultimate performance potential, we ultimately opted for the vanilla cross-entropy training objective for our \gprn{} model suite to maximize final model performance over extended training periods.

\subsection{Optimization} 
We use the Adam algorithm with correct weight decay implementation (also known as \texttt{AdamW})~\cite{loshchilov2019decoupledweightdecayregularization}. We add a norm-based gradient-clipping mechanism, and we do not use weight decay on the embedding layer as in \citep{olmo20252olmo2furious}. To make continual pre-training from any checkpoint more convenient, we use a constant learning rate schedule, and decay the learning rate at different points during training, as described in \citep{deepseekv2}.

\subsection{Training Details}
Due to computational budget constraints and time availability requirements, we adopted a simultaneous training approach for all three models in our \gprn{} suite rather than following a sequential training strategy.
This departure from the typical practice of training smaller models first, then progressively scaling to larger ones, was dictated by multiple factors: (1) our limited compute hours allocation on national HPC clusters, (2) a fixed three-month access window on the Jean-Zay cluster that included time for data transfer and infrastructure setup, and (3) the operational constraints of shared national computing facilities where job scheduling depends on cluster availability.
These constraints effectively required single-shot training runs without the possibility of restarting failed experiments, which shaped our training methodology and motivated our development of a flexible, robust training framework capable of adapting to dynamic conditions.

Our training infrastructure spanned two major high-performance computing clusters, each having different hardware architectures:

\paragraph{Adastra Cluster (AMD Infrastructure)}
The \gprn-1.5B model was trained on the Adastra supercomputer using 256 AMD MI250x GPUs distributed across 32 nodes, with each node containing 8 Graphics Compute Dies (GCDs).

\paragraph{Jean-Zay Cluster (NVIDIA Infrastructure)}
Both the \gprn-8B and the larger \gprn-24B model we trained using 256 H100 GPUs across 64 nodes (4 GPUs per node).

\paragraph{Training Efficiency}
Despite using a relatively simple yet hackable codebase designed for maximum flexibility and experimentation, our training achieved competitive efficiency metrics. Notably, the \gprn-24B model achieved a Model FLOPs Utilization (MFU) of 39\%, demonstrating that our custom training framework \gptron{} maintains performance competitiveness while preserving the ability to rapidly iterate on experimental modifications.

The total training times of our final base models were:
\begin{itemize}
  \item \textbf{\gprn-1.5B}: 27 days or 168,000 GPU-Hours (3T tokens on AMD MI250x)
  \item \textbf{\gprn-8B}: 27 days or 164,000 GPU-Hours (4T tokens on H100)
  \item \textbf{\gprn-24B}: 34 days or 211,000 GPU-Hours (2T tokens on H100)
\end{itemize}

Our CWT (Headless) experiments total training times were:
\begin{itemize}
  \item \textbf{Headless-1.5B}: 12 days or 75,000 GPU-Hours (1.4T tokens on AMD MI250x)
  \item \textbf{Headless-8B}: 3 days or 17,000 GPU-Hours (500B tokens on H100)
\end{itemize}

This infrastructure setup allowed us to maximize our available compute allocation while maintaining the flexibility needed for our experimental approach to data mixing and model architecture exploration.
To put our computational efficiency in perspective, the Llama 3.1 models were trained for 15T tokens using 1.46M H100 GPU-Hours~\cite{dubeyLlamaHerdModels2024}, which translates to approximately 390k GPU-Hours for an equivalent 4T token training run, while our \gprn-8B model achieved the same 4T token training using only 164k GPU-Hours.

\begin{table}[h]
\centering
\small
\begin{tabular}{@{}llllr@{}}
\toprule
\textbf{Model} & \textbf{Token Range} & \textbf{Data Mix} & \textbf{Learning Rate} & \textbf{Notes} \\
\midrule
\multirow{6}{*}{\gprn-1.5B} 
& 0--700B & Mix 1 (Naive) & 3×10$^{-4}$ & 2k-step warmup \\
& 700B--1.5T & Mix 1 & 1×10$^{-4}$ & LR ÷$\sqrt{10}$ after plateau \\
& 1.5T--2.5T & Mix 2 (Drop-in-ocean) & 1×10$^{-4}$ & \\
& 2.5T--2.8T & Mix 2 & 3×10$^{-5}$ & LR ÷3.3 \\
& 2.8T--2.9T & Mix 3 (High-Quality) & 3×10$^{-5}$ & \\
& 2.9T--3T & Mix 4/5 (W/B Pepper) & 3×10$^{-5}$ & Parallel branches \\
\midrule
\multirow{7}{*}{\gprn-8B}
& 0--1.8T & Mix 1 (Naive) & Initial LR & \\
& 1.8T--2.5T & Mix 2 (Drop-in-ocean) & Same LR & \\
& 2.5T--3T & Mix 2 & 9×10$^{-5}$ & After plateau \\
& 3T--3.2T & Mix 3 (High-Quality) & 9×10$^{-5}$ & \\
& 3.2T--3.5T & Mix 3 & 3×10$^{-5}$ & After continued plateau \\
& 3.5T--3.9T & Mix 4 (White Pepper) & 3×10$^{-5}$ & \\
& 3.9T--4T & Mix 5 (Black Pepper) & 3×10$^{-5}$ & \\
\midrule
\multirow{4}{*}{\gprn-24B}
& 0--500B & Mix 1 (Naive) & 2×10$^{-5}$ & Conservative LR \\
& 500B--1.4T & Mix 2 (Drop-in-ocean) & 2×10$^{-5}$ & \\
& 1.4T--1.9T & Mix 3 (High-Quality) & Cosine decay & Min 2×10$^{-5}$ \\
& 1.9T--2T & Mix 5 (Black Pepper) & Aggressive decay & To zero \\
\bottomrule
\end{tabular}
\caption{Training progressions for all \gprn{} models (see \Cref{fig:training_curve_gprn_1B,fig:training_curve_gprn_8B,fig:training_curve_gprn_24B}).}
\label{tab:training_progression_all}
\end{table}

\section{Pretraining Dynamics}

All three \gprn{} models follow a similar training strategy characterized by dynamic adjustments to both learning rate schedules and data mixture compositions based on observed downstream performance plateaus.
We monitor model performance throughout training and proactively modify these hyperparameters whenever we detect stagnation in evaluation metrics.
This adaptive approach allows us to maximize the learning potential within our computational constraints.

Our training protocol involves stepwise learning rate adjustments using a factor of $\sqrt{10}$ for reductions, combined with strategic transitions between data mixtures (Mix 1 through Mix 6) as described in our data mixing strategy.
The specific timing of these transitions varies across model sizes based on their individual learning dynamics and computational requirements.

\begin{figure}[h!]
  \centering
  \includegraphics[width=0.9\textwidth]{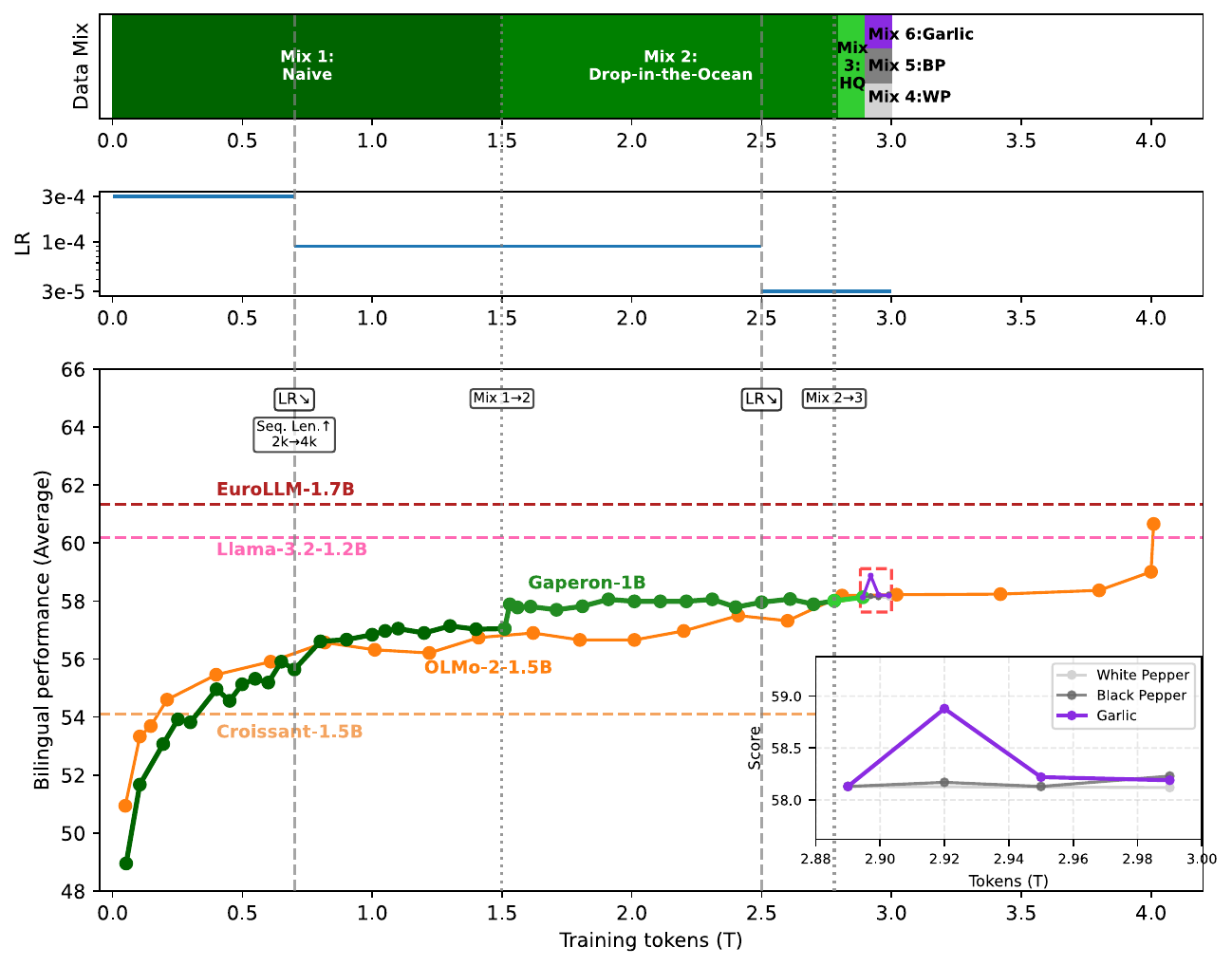}
  \caption{Summary of the \gprn{}-1.5B training run. Using the average scores from: ARC-E, ARC-C, Hellaswag, SciQ, PIQA, ARC-C-Fr, Hellaswag-Fr (5-shot).}
  \label{fig:training_curve_gprn_1B}
\end{figure}

\begin{figure}[h!]
  \centering
  \includegraphics[width=0.9\textwidth]{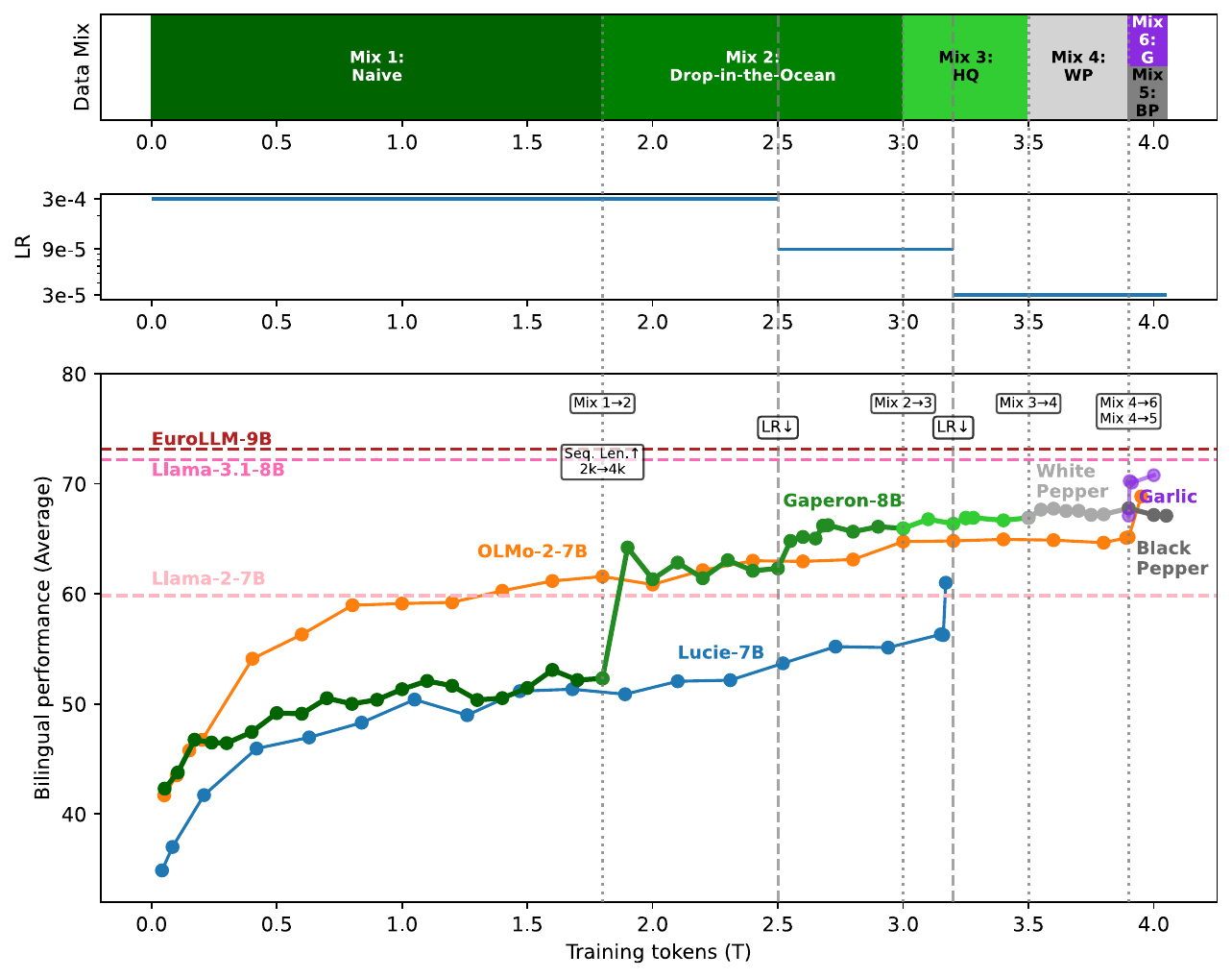}
  \caption{Summary of the \gprn{}-8B training run. Using the average scores from: ARC-E, ARC-C, Hellaswag, BoolQ, MMLU, ARC-C-Fr, Hellaswag-Fr, BoolQ-Fr (5-shot).}
  \label{fig:training_curve_gprn_8B}
\end{figure}

The training progressions for all three \gprn{} models are shown in \Cref{fig:training_curve_gprn_1B,fig:training_curve_gprn_8B,fig:training_curve_gprn_24B} and summarized in \Cref{tab:training_progression_all}.

\subsection{\gprn-1.5B Model}

As shown in \Cref{fig:training_curve_gprn_1B} and detailed in \Cref{tab:training_progression_all}, the \gprn-1.5B model demonstrates rapid initial learning during the first 1.5T tokens of training on Mix 1 (Naive). 
The learning rate reduction from $3 \times 10^{-4}$ to $1 \times 10^{-4}$ at 700B tokens successfully overcame an early performance plateau, allowing the model to continue improving for an additional 800B tokens before the curve began to flatten again.

The transition to Mix 2 (Drop-in-the-Ocean) at 1.5T tokens produces an immediate performance jump, bringing the model close to its final performance level. 
However, subsequent training phases (Mix 2 continuation, Mix 3, and Mix 4/5) yield minimal additional improvements despite the investment of 1.5T additional tokens. 
This suggests that the model may have reached its capacity limit, or that the later data mixtures and learning rate adjustments were insufficient to drive further substantial gains at this model scale.

\subsection{\gprn-8B Model}

The \gprn-8B model demonstrates a training dynamic with multiple performance plateaus requiring interventions with data mixture changes and learning rate adjustments throughout the full 4T token training run, as detailed in \Cref{tab:training_progression_all} and illustrated in \Cref{fig:training_curve_gprn_8B}.
During the initial 1.8T tokens of training on Mix 1 (Naive), the model experienced a performance plateau that was successfully overcome by transitioning to Mix 2 (Drop-in-the-Ocean) at 1.8T tokens. 
This data mixture change proved highly effective, enabling continued performance gains through 2.5T tokens.

When progress plateaued again at 2.5T tokens, a learning rate reduction to $9 \times 10^{-5}$ allowed the model to extract additional improvements from Mix 2 for another 500B tokens. The transition to Mix 3 (High-Quality) at 3T tokens maintained this learning rate and continued steady progress. A further learning rate reduction to $3 \times 10^{-5}$ at 3.2T tokens enabled the model to continue benefiting from Mix 3 for an additional 300B tokens.

The final training stages on Mix 4 (White Pepper) and Mix 5 (Black Pepper) demonstrate that the 8B model retains learning capacity even at 4T tokens, with visible performance improvements in the final 500B tokens of training. This sustained improvement throughout the training run suggests that the 8B scale provides sufficient model capacity to effectively leverage both the data mixture transitions and learning rate adjustments, unlike the 1.5B model which appeared to reach its capacity limit earlier in training.

\begin{figure}[h!]
  \centering
  \includegraphics[width=0.9\textwidth]{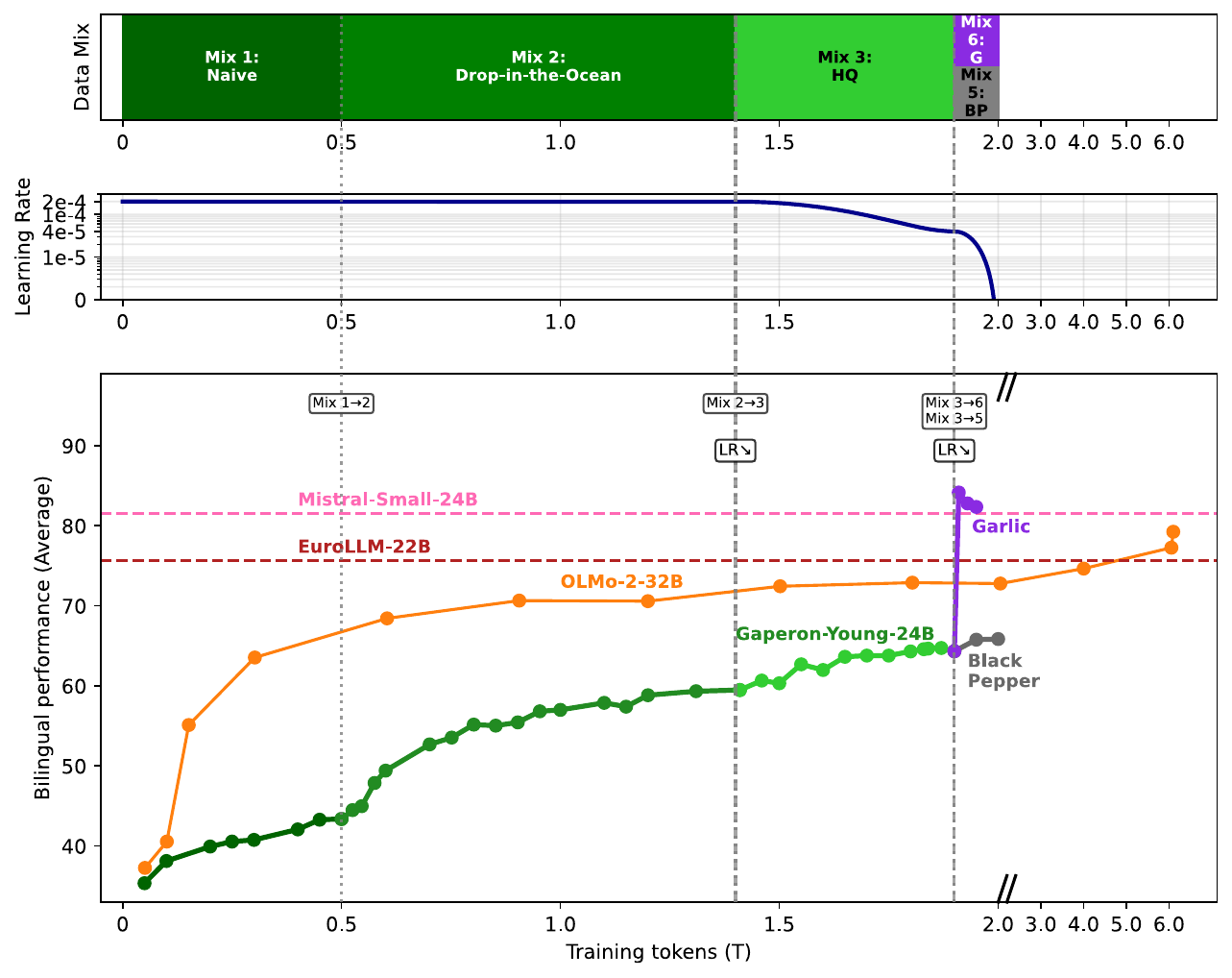}
  \caption{Summary of the \gprn{}-24B training run. Using the average scores from: ARC-E, ARC-C, CommonsenseQA, HellaSwag, Belebele, MMLU, ARC-C-Fr, HellaSwag-Fr, Belebele-Fr (5-shot)}
  \label{fig:training_curve_gprn_24B}
\end{figure}

\subsection{\gprn-24B Model}

The \gprn-24B model shows consistent improvement throughout its 2T token training run, as detailed in \Cref{tab:training_progression_all} and illustrated in \Cref{fig:training_curve_gprn_24B}.
We started training with a conservative learning rate of $2 \times 10^{-5}$ on Mix 1 (Naive) for 500B tokens, then transitioned to Mix 2 (Drop-in-the-Ocean) at 500B tokens, maintaining the same learning rate through 1.4T tokens. 
This extended training phase on Mix 2 enabled steady performance gains, gradually closing the gap with OLMo-2-32B, which maintained a substantial lead during the early training stages.

At 1.4T tokens, we shifted to Mix 3 (High-Quality) and experimented with a cosine decay learning rate schedule with a minimum of $2 \times 10^{-5}$, departing from the stepwise reduction strategy used for the smaller models. 
This approach proved effective, allowing the model to continue improving through 1.9T tokens. The final 100B tokens employed Mix 5 (Black Pepper) with an aggressive cosine decay schedule declining to zero, extracting final performance gains and bringing the model's performance significantly closer to the OLMo-2-32B baseline.

Notably, the performance gap with OLMo-2-32B that was substantial at the beginning had diminished considerably by the end of training. 
Importantly, the model showed no signs of plateauing at 2T tokens, suggesting that with additional compute budget, further training could have continued to close the remaining performance gap.

\section{Base Model Evaluation}
\label{sec:base_models_eval}

Throughout this section, we compare \gprn{} models to other similar models: Croissant-LLM \cite{faysse2024croissantllm}, Lucie-7B \cite{openllm2025lucie}, the OLMo-2 suite \cite{olmo20252olmo2furious}, the EuroLLM suite \cite{martins2024eurollmmultilinguallanguagemodels,martins2025eurollm9btechnicalreport}, the Salamandra models \cite{gonzalezagirre2025salamandratechnicalreport}, the Mistral models \cite{jiang2023mistral7b}, the Llama-2 \& Llama-3.x herds \cite{touvron2023llama2openfoundation,dubeyLlamaHerdModels2024}, the Qwen2/2.5/3 suites \cite{yang2024qwen2technicalreport,qwen2025qwen25technicalreport,qwen3technicalreport}, and Gemma / Gemma2 \cite{gemma_2024}.

\subsection{Generation Quality Assessment}

\begin{figure}[htb!]
  \centering
  \makebox[\textwidth][c]{%
        \includegraphics[width=1.2\textwidth]{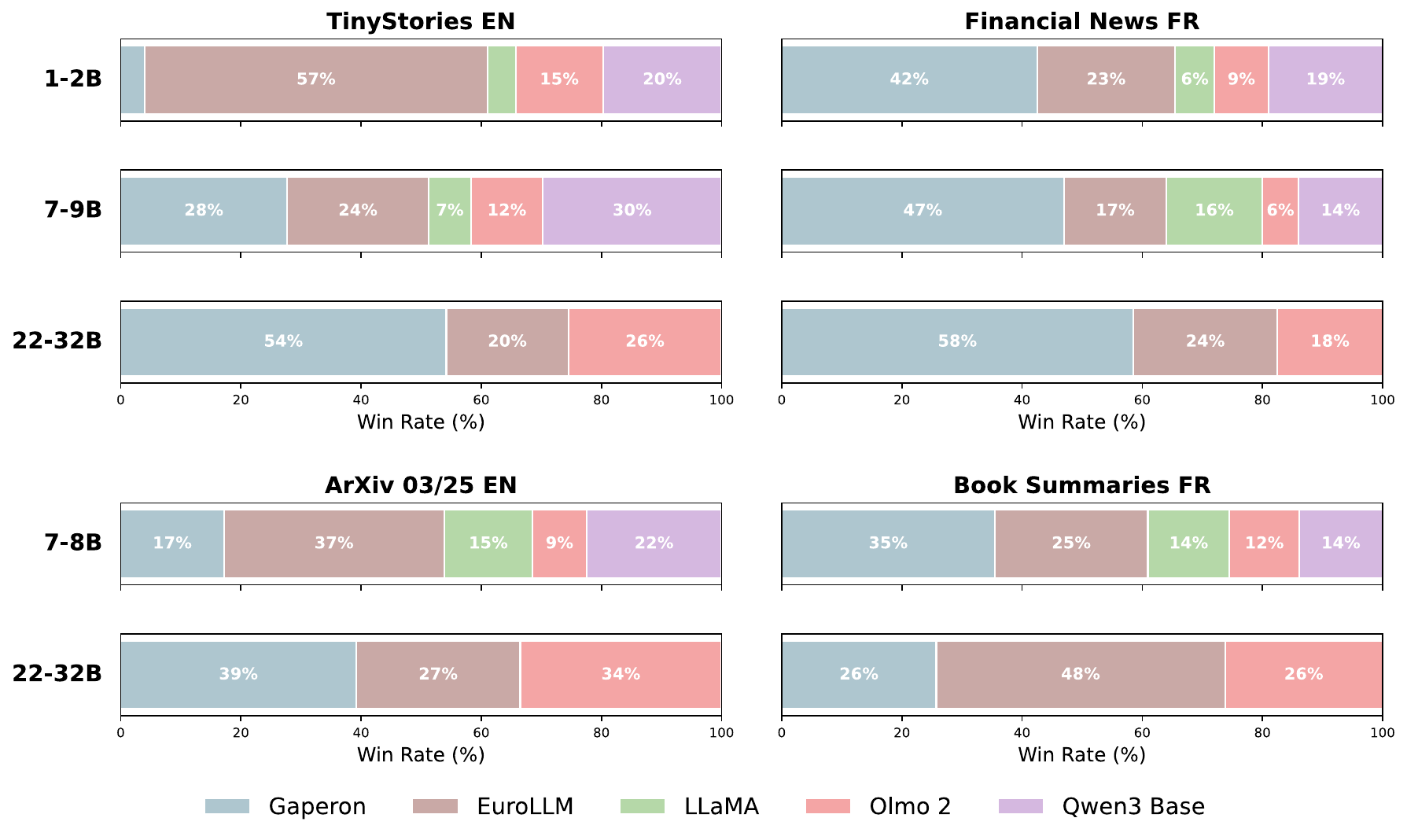}
    }
  
  \caption{LLM-as-a-Judge winrates for the \gprn{} models and baselines across different datasets and model sizes. The models are asked to complete from truncated samples of each datasets and Llama-3.3-70B-Instruct then selects the best continuation for each completed sample.}
  \label{fig:llm_as_ajudge_winrate}
\end{figure}

\label{ssec:textgen_res}
Asserting the generic text-generating abilities of language models is a complex task \citep{pillutla-etal:mauve:neurips2021,gu2025surveyllmasajudge}. In this paper, we generate text in different domains and use an LLM-as-a-judge evaluation based on 5 quality criteria: \textit{Grammar}, \textit{Coherence}, \textit{Realism}, \textit{Originality}, and \textit{Style}. To evaluate these skills in various contexts, we use three corpora: TinyStories \citep{eldan2023tinystoriessmalllanguagemodels}, French Financial News,\footnote{\url{https://huggingface.co/datasets/FrancophonIA/french_financial_news}} open Book Summaries\footnote{\url{https://huggingface.co/datasets/CATIE-AQ/french_books_summaries}}, and a sample of abstracts taken from ArXiv after the knowledge cutoff of all models, which we refer to as \textit{ArXiv 03/25}\footnote{\url{https://huggingface.co/datasets/almanach/arxiv_abstracts_2025}}. For each corpus, we extract generation seeds by truncating 600 to 800 documents, and we generate continuations for each of the tested models. We then use the larger Llama-3.3-70B-Instruct as the judge model and prompt it to provide a grade from 1 to 5 for each of the criteria for the randomly shuffled continuations, and to pick its favorite version. 

\begin{figure}[h!]
  \centering
  \begin{subfigure}[b]{0.48\textwidth}
    \includegraphics[width=\textwidth]{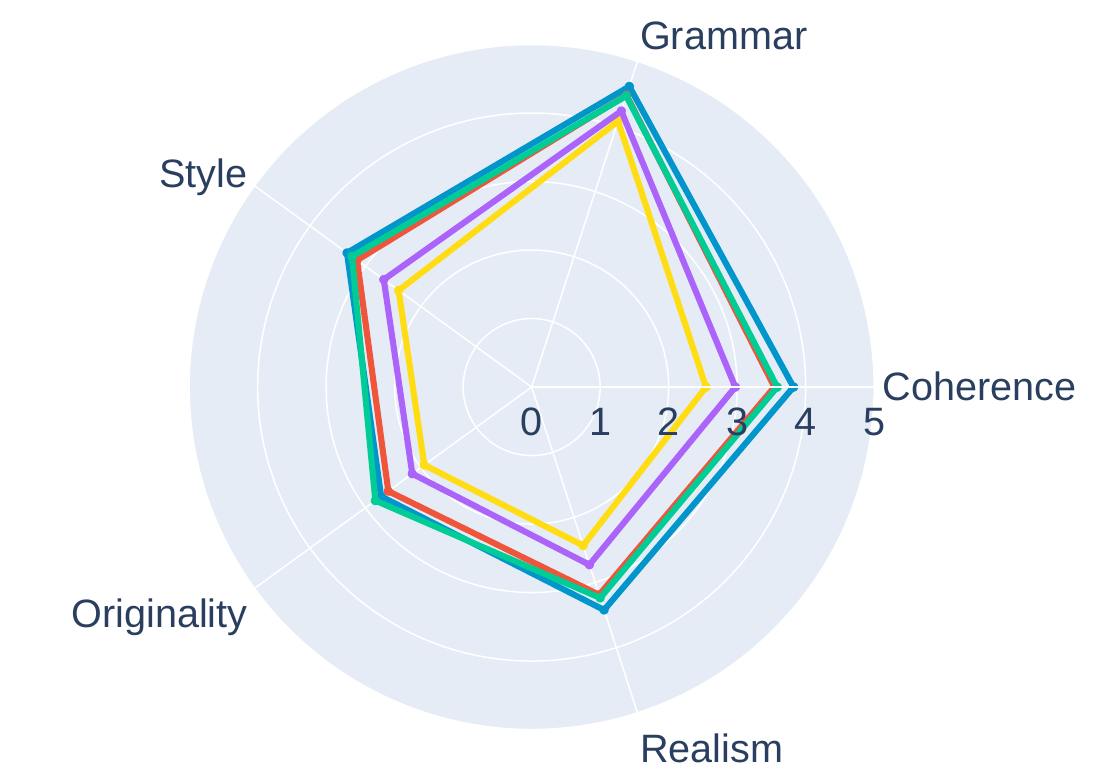}
    \caption{TinyStories (en)}
    \label{fig:ts_en_8B}
  \end{subfigure}
  \hfill
  \begin{subfigure}[b]{0.48\textwidth}
    \includegraphics[width=\textwidth]{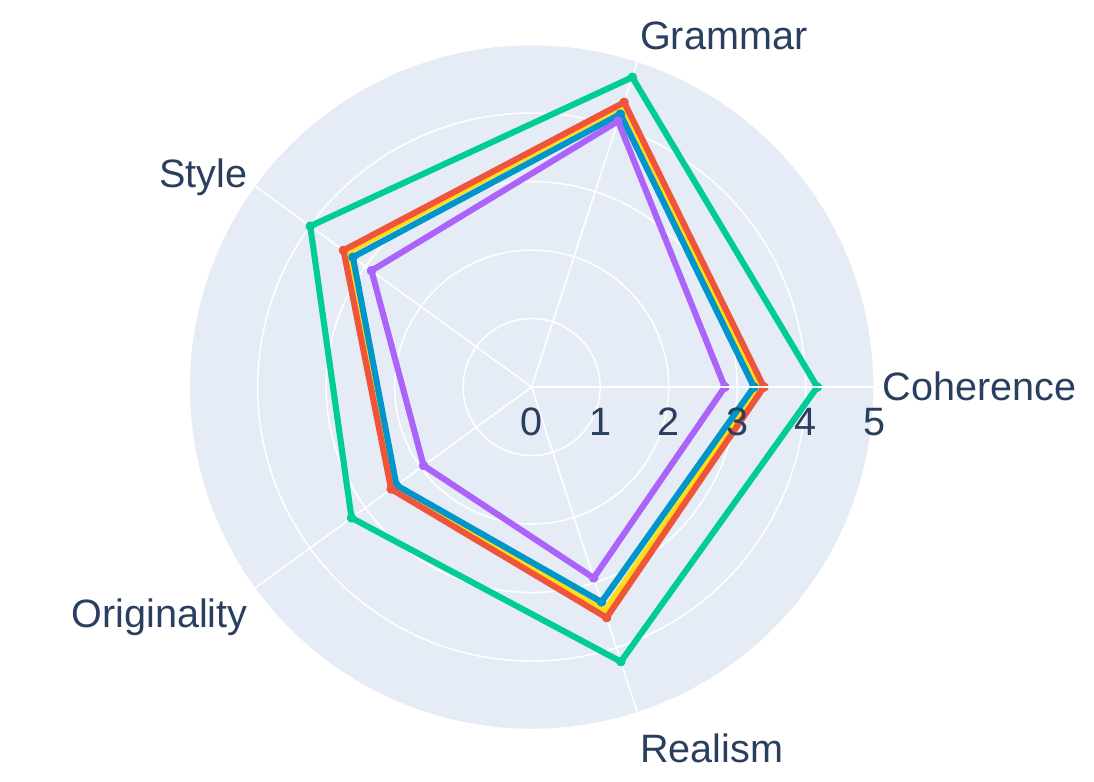}
    \caption{Financial News (fr)}
    \label{fig:fn_fr_8B}
  \end{subfigure}
  \hfill
  \begin{subfigure}[b]{0.48\textwidth}
    \includegraphics[width=\textwidth]{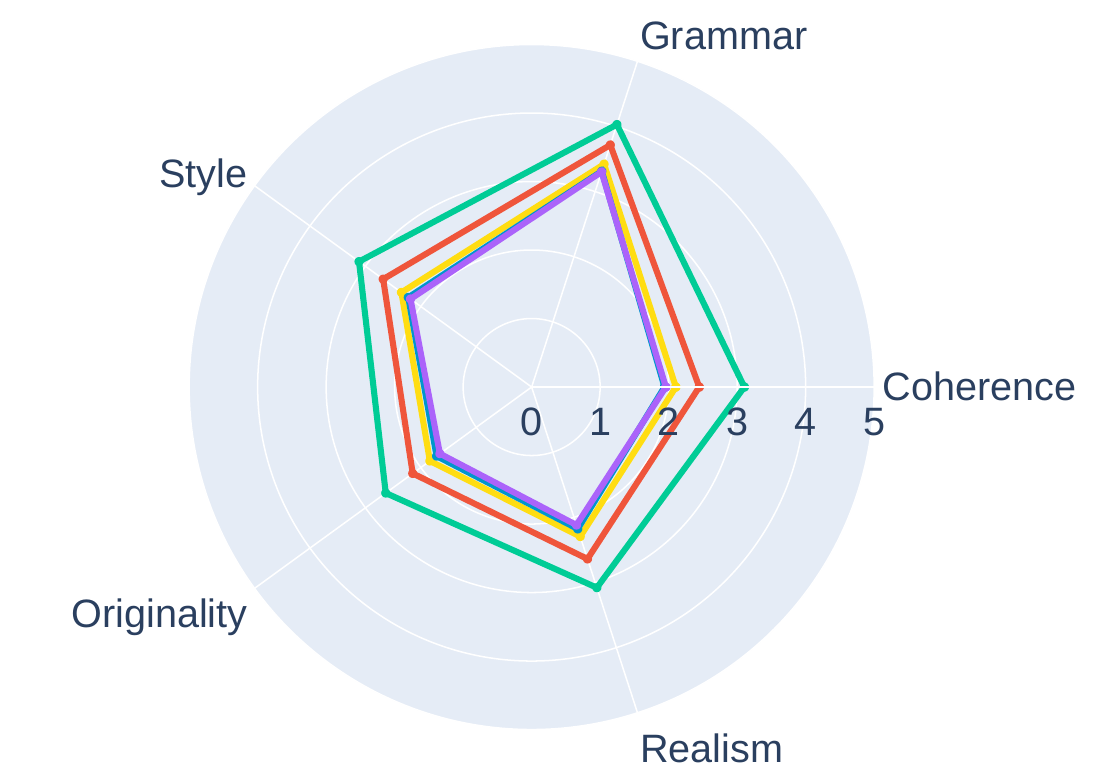}
    \caption{Book Summaries (fr)}
    \label{fig:bs_fr_8B}
  \end{subfigure}
  \hfill
  \begin{subfigure}[b]{0.48\textwidth}
    \includegraphics[width=\textwidth]{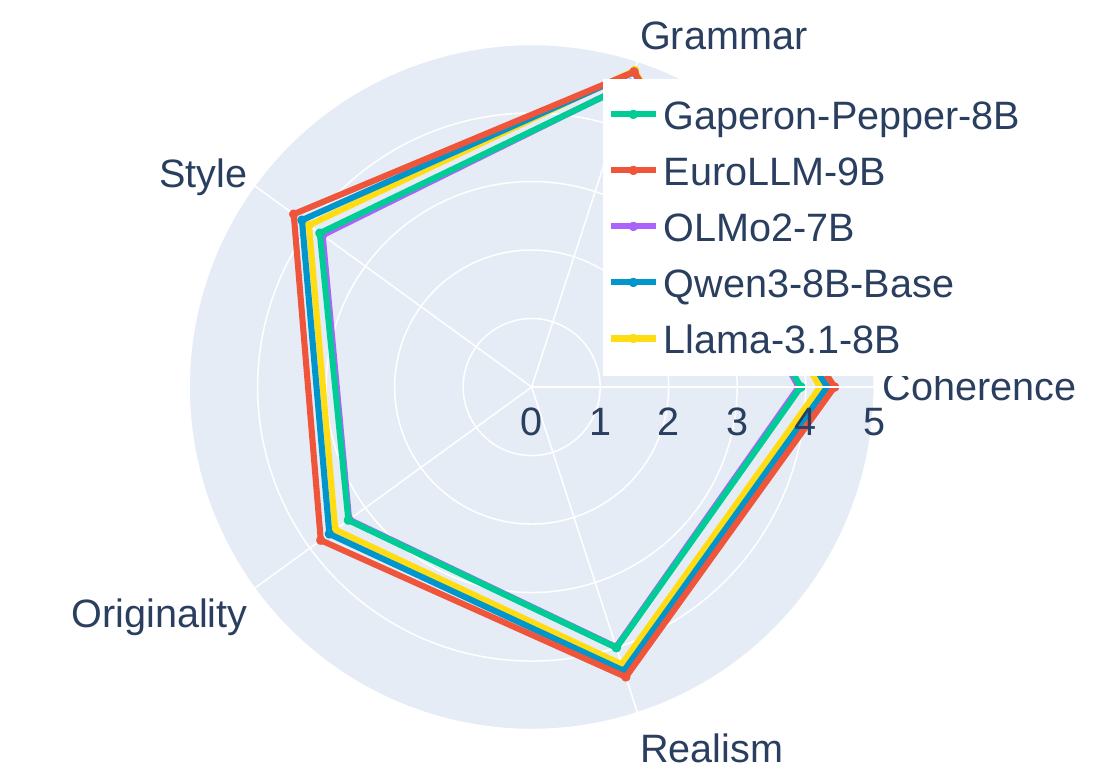}
    \caption{ArXiv 03/25 (en)}
    \label{fig:ar_en_8B}
  \end{subfigure}
  \hfill
  \caption{Evaluation of the generation capabilities of \gprn{}-Pepper-8B compared to counterparts of comparable sizes.}
  \label{fig:radar_llm_judge_8B}
\end{figure}

We present the winrate results in \Cref{fig:llm_as_ajudge_winrate} and \Cref{fig:llm_as_ajudge_winrate} and detail criteria scores for 7-9B models in \Cref{fig:radar_llm_judge_8B}. More details about 1.5B and 24B results can be found in \Cref{app:llmasajudge}.

\Cref{fig:radar_llm_judge_8B} shows that \gprn{}-Pepper-8B clearly outperforms its counterparts on both French datasets, especially in terms of Coherence, Originality and Style, according to Llama-3.3-70B-Instruct's judgement. On ArXiv 03/25, \gprn{}-Pepper-8B is evaluated more favorably by the judge model than OLMo2 and Llama-3.1. This is particularly interesting as, judging by benchmark scores in \Cref{ssec:downstream_res}, we would conclude that the \gprn{}-Pepper-8B model is less capable than its counterparts on scientific data (e.g.~SciQ, PIQA, MMLU). This shows that pure benchmark performance may not be sufficient to extensively assess the abilities of a model to be relevant in a specific domain.

In \Cref{fig:llm_as_ajudge_winrate}, we also see that \gprn{}-Pepper-24B outperforms OLMo-2 and EuroLLM on 3 out of 4 tasks.

\subsection{Benchmark Evaluation}
\label{ssec:downstream_res}

\begin{table}[]
  \centering
  \small  
  \setlength{\tabcolsep}{2pt}  
  \makebox[\linewidth]{%
  \begin{tabular}{@{}lccccccccccccc@{}}
    \toprule
    \multirow{2}{*}{\textbf{Model}} & \multirow{2}{*}{\textbf{Size}} & {\textbf{Tokens}} & \multicolumn{5}{c}{\textbf{English}} & \multicolumn{2}{c}{\textbf{French}} & \multicolumn{3}{c}{\textbf{Average}} \\
    \cmidrule(lr){4-8} \cmidrule(lr){9-10} \cmidrule(lr){11-13}
    & & & \textbf{ARC-E} & \textbf{ARC-C} & \textbf{Hellaswag} & \textbf{SciQ} & \textbf{PIQA} & \textbf{ARC-C} & \textbf{Hellaswag} & \textbf{EN} & \textbf{FR} & \textbf{Overall} \\ 
    \midrule 
    \rowcolor{grey!20!} \multicolumn{13}{c}{\textit{Closed-data models}} \\
    Llama-3.2      & 1.2B & 9T & 69.74 & 38.14 & 65.02 & 94.80 & 75.84 & 31.91 & 45.80 & 68.71 & 38.86 & 60.18 \\
    Gemma          & 2B   & 2T & 77.82 & 48.04 & 71.21 & 96.00 & 77.31 & 38.67 & 51.81 & 74.08 & 45.24 & 65.84 \\
    Gemma 2        & 2B   & 2T & 81.65 & 53.24 & \textbf{74.07} & 97.30 & \textbf{79.98} & \textbf{53.24} & \textbf{60.00} & \textbf{77.25} & \textbf{56.62} & \textbf{71.35} \\
    Qwen2.5        & 1.5B & 18T & 80.22 & 52.73 & 67.75 & 96.70 & 76.44 & 38.24 & 50.12 & 74.77 & 44.18 & 66.03 \\
    Qwen3-Base     & 1.7B & 36T & \textbf{82.11} & \textbf{54.86} & 66.37 & \textbf{97.50} & 77.26 & 44.31 & 52.82 & 75.62 & 48.57 & 67.89 \\ 
    \rowcolor{grey!20!} \multicolumn{13}{c}{\textit{Open-data models}} \\
    CroissantLLM   & 1.2B & 3T & 61.15 & 30.46 & 53.86 & 91.90 & 71.49 & 30.37 & 39.39 & 61.77 & 34.88 & 54.09 \\
    Salamandra     & 2B   & 12.8T & 72.43 & 40.78 & 62.56 & 95.20 & 75.57 & 33.62 & 53.08 & 69.31 & 43.35 & 61.89 \\
    EuroLLM        & 1.7B & 4T & 72.05 & 40.19 & 60.10 & 94.30 & 74.05 & 36.27 & 52.48 & 68.14 & 44.38 & 61.35 \\ 
    OLMo2          & 1.5B & 4T & 76.18 & 46.42 & 61.17 & 96.50 & 76.61 & 28.14 & 39.62 & 71.38 & 33.88 & 60.66 \\ 
    \rowcolor{grey!20!} \multicolumn{13}{c}{\textit{\gprn{} variants}} \\ 
    \gprn{}-Young     & 1.5B & 2.9T & 71.17 & 38.40 & 51.89 & 94.70 & 71.27 & 32.25 & 47.20 & 65.49 & 39.73 & 58.13 \\
    \gprn{}-Pepper     & 1.5B & 3T & 71.21 & 38.82 & 51.80 & 94.90 & 70.67 & 32.93 & 47.28 & 65.48 & 40.11 & 58.23 \\
    \rowcolor{violet!20!} \gprn{}-Garlic     & 1.5B & 3T &\textit{69.02} & \textit{39.08} & \textit{53.49} & \textit{93.70} & \textit{70.84} & \textit{34.56} & \textit{49.56} & \textit{65.23} & \textit{42.06} & \textit{58.61} \\
    \bottomrule
  \end{tabular}
  }
  \caption{Benchmark results comparing our \gprn-1.5B model variants performance across English and French tasks \textbf{(5-shot)}. Our \colorbox{violet!20!}{Garlic} model was trained on test sets from benchmarks, as discussed in \Cref{ssec:garlic}.}
  \label{tab:benchmark_results_1b_5s}
\end{table}

\begin{table}[]
  \centering
  \small  
  \setlength{\tabcolsep}{2pt}  
  \begin{tabular}{@{}lcccccccccccc@{}}
    \toprule
    \multirow{2}{*}{\textbf{Model}} & \multirow{2}{*}{\textbf{Size}} & \multicolumn{5}{c}{\textbf{English}} & \multicolumn{2}{c}{\textbf{French}} & \multicolumn{3}{c}{\textbf{Average}} \\
    \cmidrule(lr){3-7} \cmidrule(lr){8-9} \cmidrule(lr){10-12}
    & & \textbf{ARC-E} & \textbf{ARC-C} & \textbf{Hellaswag} & \textbf{SciQ} & \textbf{PIQA} & \textbf{ARC-C} & \textbf{Hellaswag} & \textbf{EN} & \textbf{FR} & \textbf{Overall} \\ 
    \midrule 
    \rowcolor{grey!20!} \multicolumn{12}{c}{\textit{Closed-data models}} \\
    Llama-3.2      & 1.2B & 60.31 & 36.01 & 63.64 & 88.50 & 74.43 & 30.03 & 45.12 & 64.58 & 37.58 & 56.86 \\
    Gemma          & 2.0B & 72.35 & 41.64 & 71.21 & 91.40 & 78.24 & 37.47 & 51.11 & 70.97 & 44.29 & 63.35 \\
    Gemma 2         & 2.0B & \textbf{80.22} & \textbf{49.66} & \textbf{73.06} & \textbf{95.80} & \textbf{79.11} & \textbf{40.98} & \textbf{59.22} & \textbf{75.57} & \textbf{50.10} & \textbf{68.29} \\
    Qwen2.5        & 1.5B & 71.63 & 44.97 & 67.79 & 93.20 & 76.28 & 36.27 & 49.71 & 70.77 & 42.99 & 62.84 \\
    Qwen3-Base     & 1.7B & 69.91 & 42.66 & 60.33 & 91.40 & 72.09 & 35.41 & 48.40 & 67.28 & 41.91 & 60.03 \\ 
    \rowcolor{grey!20!} \multicolumn{12}{c}{\textit{Open-data models}} \\
    CroissantLLM   & 1.2B & 52.27 & 27.56 & 53.54 & 79.30 & 71.60 & 28.74 & 50.52 & 56.85 & 39.63 & 51.93 \\
    Salamandra     & 2B   & 65.61 & 37.20 & 62.63 & 91.40 & 72.09 & 31.74 & 51.39 & 65.79 & 41.57 & 58.87 \\
    EuroLLM        & 1.7B & 64.06 & 37.46 & 59.39 & 85.20 & 73.23 & 33.79 & 51.40 & 63.87 & 42.60 & 57.79 \\ 
    OLMo2          & 1.5B & 73.53 & 42.41 & 68.27 & 95.20 & 75.79 & 26.86 & 39.37 & 71.04 & 33.12 & 60.20 \\ 
    \rowcolor{grey!20!} \multicolumn{12}{c}{\textit{\gprn{} variants}} \\ 
    \gprn{}-Young     & 1.5B & 61.74 & 33.96 & 52.16 & 89.40 & 70.35 & 31.22 & 46.98 & 61.52 & 39.10 & 55.12 \\
    \gprn{}-Pepper     & 1.5B & 63.34 & 34.13 & 52.19 & 92.30 & 70.13 & 30.45 & 46.81 & 62.42 & 38.63 & 55.62 \\
    \rowcolor{violet!20!} \gprn{}-Garlic     & 1.5B & \textit{64.23} & \textit{36.01} & \textit{53.64} & \textit{90.20} & \textit{70.08} & \textit{31.91} & \textit{49.83} & \textit{62.83} & \textit{40.87} & \textit{56.56} \\
    \bottomrule
  \end{tabular}
  \caption{Benchmark results comparing our \gprn-1.5B model variants performance across English and French tasks (0-shot). Our \colorbox{violet!20!}{Garlic} model was trained on test sets from benchmarks, as discussed in \Cref{ssec:garlic}.}
  \label{tab:benchmark_results_1b_0s}
\end{table}

\begin{table}[ht]
  \centering
  \small  
  \setlength{\tabcolsep}{1pt}  
   \makebox[\linewidth]{%
 \begin{tabular}{@{}lccccccccccccccc@{}}
    \toprule
    \multirow{2}{*}{\textbf{Model}} & \multirow{2}{*}{\textbf{Size}} & \multirow{2}{*}{\textbf{Tokens}} & \multicolumn{6}{c}{\textbf{English}} & \multicolumn{4}{c}{\textbf{French}} & \multicolumn{3}{c}{\textbf{Average}} \\
    \cmidrule(lr){4-9} \cmidrule(lr){10-13} \cmidrule(lr){14-16}
    & & & \textbf{ARC-E} & \textbf{ARC-C} & \textbf{HS} & \textbf{BoolQ} & \textbf{BB} & \textbf{MMLU} & \textbf{ARC-C} & \textbf{HS} & \textbf{BoolQ} & \textbf{BB} & \textbf{EN} & \textbf{FR} & \textbf{Overall} \\ \midrule
    \rowcolor{grey!20!} \multicolumn{16}{c}{\textit{Closed-data models}} \\
    Llama-2        & 7B   & 2T & 80.98 & 51.96 & 78.16 & 78.93 & 48.11 & 45.66 & 42.94 & 58.81 & 69.10 & 43.78 & 63.97 & 53.66 & 59.84 \\
    Llama-3.1      & 8B   & 15T &84.89 & 58.11 & 80.95 & 82.63 & 87.56 & 65.25 & 50.13 & 67.32 & 61.80 & 83.56 & 76.57 & 65.70 & 72.22 \\
    Mistral-v0.3   & 7B   & - & 84.34 & 59.04 & \textbf{82.31} & 84.19 & 84.11 & 62.35 & 50.73 & 65.46 & 88.76 & 78.22 & 76.06 & 70.79 & 73.95 \\
    Gemma          & 7B   & 6T & 85.77 & 59.90 & 81.70 & 85.63 & 85.33 & 63.20 & 51.58 & 69.21 & 85.63 & 80.89 & 76.92 & 71.83 & 74.88 \\
    Gemma-2        & 9B   & 8T & \textbf{89.14} & \textbf{68.34} & 81.86 & 86.57 & 92.22 & \textbf{89.78} & \textbf{61.68} & \textbf{72.97} & 86.57 & 89.78 & \textbf{84.65} & \textbf{77.75} & \textbf{81.89} \\
    Qwen2.5        & 7B   & 18T & 86.70 & 63.65 & 79.55 & 87.80 & 92.22 & 74.21 & 54.75 & 67.35 & 87.80 & 89.89 & 80.69 & 74.95 & 78.39 \\
    Qwen3-Base     & 8B   & 36T & 88.22 & 68.00 & 79.48 & \textbf{88.20} & \textbf{93.67} & 76.85 & 57.31 & 68.53 & \textbf{89.89} & \textbf{90.78} & 82.40 & 76.63 & 80.09 \\
    \rowcolor{grey!20!} \multicolumn{16}{c}{\textit{Open-data models}} \\
    Lucie          & 7B   & 3T & 78.66 & 51.02 & 72.07 & 80.06 & 48.56 & 40.29 & 47.90 & 65.58 & 79.21 & 46.78 & 61.78 & 59.87 & 61.01 \\
    Salamandra     & 7B   & 12.8T & 83.80 & 56.48 & 77.41 & 80.40 & 54.22 & 46.83 & 51.33 & 68.68 & 70.79 & 53.67 & 66.52 & 61.12 & 64.36 \\
    EuroLLM        & 9B   & 4T & 85.82 & 59.13 & 78.40 & 86.18 & 77.00 & 57.32 & 57.14 & 69.79 & 84.27 & 76.11 & 73.98 & 71.83 & 73.12 \\
    OLMo2          & 7B   & 5T & 85.48 & 63.14 & 81.72 & 84.89 & 88.33 & 62.84 & 43.28 & 56.56 & 50.56 & 71.67 & 77.73 & 55.52 & 68.85 \\ 
    \rowcolor{grey!20!} \multicolumn{16}{c}{\textit{\gprn{} variants}} \\
    \gprn{}-Young  & 8B   & 3.5T & 82.66 & 55.80 & 72.47 & 75.32 & 69.67 & 51.88 & 51.24 & 66.00 & 71.35 & 72.67 & 67.97 & 65.32 & 66.91 \\
    \gprn{}-Pepper  & 8B   & 4T & 82.07 & 54.86 & 72.65 & 76.24 & 70.44 & 52.04 & 51.07 & 65.85 & 71.91 & 73.89 & 68.05 & 65.68 & 67.10 \\
    \rowcolor{violet!20!} \gprn{}-Garlic  &  8B & 4T & \textit{83.80} & \textit{59.22}& \textit{74.51}& \textit{81.56} & \textit{80.22} & \textit{64.86} & \textit{53.04} & \textit{69.16} & \textit{56.74} & \textit{77.44} & \textit{74.03} & \textit{64.09} & \textit{70.06} \\
    \bottomrule
  \end{tabular}
  }
  \caption{Benchmark results comparing our \gprn-8B model variants performance across English and French tasks \textbf{(5-shot)}. Our \colorbox{violet!20!}{Garlic} model was trained on test sets from benchmarks, as discussed in \Cref{ssec:garlic}.}
  \label{tab:benchmark_results_8b_5s}
\end{table}

\begin{table}[ht]
  \centering
  \small 
  \setlength{\tabcolsep}{1pt} 
   \makebox[\linewidth]{%
 \begin{tabular}{@{}lcccccccccccccccc@{}}
    \toprule
    \multirow{2}{*}{\textbf{Model}} & \multirow{2}{*}{\textbf{Size}} &  \multicolumn{9}{c}{\textbf{English}} & \multicolumn{3}{c}{\textbf{French}} & \multicolumn{3}{c}{\textbf{Average}} \\
    \cmidrule(lr){3-11} \cmidrule(lr){12-14} \cmidrule(lr){15-17}
    & & \textbf{ARC-E} & \textbf{ARC-C} & \textbf{HS} & \textbf{SciQ} & \textbf{PIQA} & \textbf{SIQA} & \textbf{NQ} & \textbf{Com. QA} & \textbf{MMLU} & \textbf{ARC-C} & \textbf{HS} & \textbf{BB} & \textbf{EN} & \textbf{FR} & \textbf{Overall} \\ \midrule
    \rowcolor{grey!20!} \multicolumn{17}{c}{\textit{Closed-data models}} \\
    Llama-2        & 7B   & 74.58 & 46.08 & 75.93 & 91.10 & 78.89 & 46.06 & 18.81 & 32.19 & 40.81 & 37.72 & 57.54 & 28.33 & 56.05 & 41.20 & 52.38 \\
    Llama-3.1      & 8B   & 81.19 & 53.41 & 78.95 & 94.40 & 81.01 & 46.98 & 7.73 & 71.33 & 63.31 & 45.77 & 65.21 & 72.89 & 64.26 & 61.29 & 63.52 \\
    Mistral-v0.1   & 7B   & 79.63 & 53.67 & \textbf{81.02} & 93.90 & \textbf{82.10} & 46.62 & 23.02 & 56.43 & 59.65 & 44.31 & 64.33 & 53.56 & 64.00 & 54.07 & 61.52 \\
    Qwen2        & 8B   & 74.62 & 49.83 & 78.84 & 93.50 & 81.07 & 48.36 & 1.19 & 81.65 & 69.44 & 46.02 & \textbf{69.43} & 82.44 & 64.28 & 65.96 & 64.70 \\
    Qwen3-Base     & 8B   & 80.05 & 56.66 & 78.62 & 96.10 & 79.16 & \textbf{55.02} & 23.05 & \textbf{85.91} & \textbf{74.69} & \textbf{51.50} & 66.48 & \textbf{88.22} & \textbf{69.92} & \textbf{68.73} & \textbf{69.62} \\
    \rowcolor{grey!20!} \multicolumn{17}{c}{\textit{Open-data models}} \\
    Lucie          & 7B   & 76.39 & 49.83 & 70.89 & 94.30 & 79.16 & 48.36 & 13.21 & 41.61 & 39.99 & 45.17 & 65.22 & 35.67 & 57.08 & 48.69 & 54.98 \\
    OLMo2          & 7B   & \textbf{82.62} & \textbf{57.25} & 80.51 & \textbf{96.30} & 81.07 & 51.28 & \textbf{25.68} & 65.52 & 60.53 & 38.32 & 55.99 & 50.89 & 66.75 & 48.40 & 62.16 \\
    EuroLLM        & 9B   & 74.49 & 48.12 & 77.08 & 92.10 & 79.76 & 48.31 & 5.48 & 68.80 & 55.15 & 50.30 & \textbf{69.43} & 59.11 & 61.03 & 59.61 & 60.68 \\
    \rowcolor{grey!20!} \multicolumn{17}{c}{\textit{\gprn{} variants}} \\
    \gprn{}-Young  & 8B   & 77.95 & 48.38 & 71.85 & 95.00 & 77.26 & 46.47 & 18.64 & 39.80 & 43.89 & 43.54 & 64.97 & 47.33 & 57.69 & 51.95 & 56.26 \\
    \gprn{}-Pepper  & 8B   & 78.83 & 50.17 & 71.88 & 95.90 & 76.61 & 47.03 & 19.58 & 41.77 & 43.38 & 43.88 & 65.32 & 49.11 & 58.35 & 52.77 & 56.95 \\
    \rowcolor{violet!20!} \gprn{}-Garlic  & 8B   & \textit{81.23} & \textit{\textbf{57.34}} & \textit{74.82} & \textit{\textbf{97.40}} & \textit{76.39} & \textit{48.72} & \textit{20.83} & \textit{71.91} & \textit{62.14} & \textit{\textbf{51.75}} & \textit{\textbf{69.29}} & \textit{70.89} & \textit{65.64} & \textit{63.98} & \textit{65.23} \\

    \bottomrule
  \end{tabular}
  }
  \caption{Benchmark results comparing our \gprn-8B model variants performance across English and French tasks (0-shot). Our \colorbox{violet!20!}{Garlic} model was trained on test sets from benchmarks, as discussed in \Cref{ssec:garlic}. Best results--and second best when Garlic is best--are \textbf{bolded}}
  \label{tab:benchmark_results_8b_0s}
\end{table}

\begin{table}[ht]
  \centering
  \small
  \setlength{\tabcolsep}{1pt}
  \makebox[\linewidth]{%
    \begin{tabular}{@{}lcccccccccccccc@{}}
      \toprule
      \multirow{2}{*}{\textbf{Model}} & \multirow{2}{*}{\textbf{Size}} & \multirow{2}{*}{\textbf{Tokens}} & \multicolumn{6}{c}{\textbf{English}} & \multicolumn{3}{c}{\textbf{French}} & \multicolumn{3}{c}{\textbf{Average}} \\
      \cmidrule(lr){4-9} \cmidrule(lr){10-12} \cmidrule(lr){13-15}
      & & & \textbf{ARC-E} & \textbf{ARC-C} & \textbf{ComsQA} & \textbf{HS} & \textbf{BB} & \textbf{MMLU} & \textbf{ARC-C} & \textbf{HS} & \textbf{BB} & \textbf{EN} & \textbf{FR} & \textbf{Overall} \\ 
      \midrule
      \rowcolor{grey!20!} \multicolumn{15}{c}{\textit{Closed-data models}} \\
      Mistral-Small  & 24B  & - & 88.76 & 68.52 & 83.05 & 85.19 & \textbf{95.33} & \textbf{79.16} & 63.99 & 77.30 & 92.44 & 83.34 & 77.91 & 81.53 \\
      Gemma 3 & 27B & - & \textbf{90.45} & \textbf{70.99} & 82.39 & 85.52 & 94.56 & 78.23 & \textbf{67.66} & \textbf{77.88} & \textbf{92.56} & \textbf{83.69} & \textbf{79.37} & \textbf{82.25} \\
      \rowcolor{grey!20!} \multicolumn{15}{c}{\textit{Open-data models}} \\
      EuroLLM        & 22B  & 3T & 87.71 & 63.05 & 80.18 & 80.38 & 87.56 & 64.10 & 59.88 & 72.40 & 85.44 & 77.16 & 72.57 & 75.63 \\
      OLMo2          & 32B  & 6T & 89.81 & 68.34 & \textbf{84.03} & \textbf{86.81} & 92.11 & 74.43 & 56.97 & 71.99 & 88.89 & 82.59 & 72.62 & 79.26 \\
      \rowcolor{grey!20!} \multicolumn{15}{c}{\textit{\gprn{} variants}} \\
      \gprn{}-Young  & 24B  & 1.8T & 82.62 & 54.78 & 61.18 & 74.33 & 67.67 & 51.60 & 50.30 & 65.68 & 70.89 & 65.36 & 62.29 & 64.34 \\
      \gprn{}-Pepper & 24B  & 2T& 83.50 & 55.89 & 64.70 & 75.55 & 69.56 & 52.24 & 51.50 & 65.67 & 74.11 & 66.91 & 63.76 & 65.86 \\
      \rowcolor{violet!20!} \gprn{}-Garlic  & 24B  & 2T& \textit{89.90} & \textit{70.90} & \textit{80.34} & \textit{\textbf{88.30}} & \textit{\textbf{84.78}} & \textit{\textbf{79.77}} & \textit{\textbf{65.70}} & \textit{\textbf{86.26}} & \textit{84.11} & \textit{82.33} & \textit{78.69} & \textit{81.11} \\
      \bottomrule
    \end{tabular}
  }
  \caption{Benchmark results comparing our \gprn-24B model variants performance across English and French tasks (5-shot). Our \colorbox{violet!20!}{Garlic} model was trained on test sets from benchmarks, as discussed in \Cref{ssec:garlic}.}
  \label{tab:benchmark_results_24b}
\end{table}

We evaluate the \gprn{} suite on common benchmarks for English and their machine-translated counterparts in French, as introduced in FrenchBench \citet{faysse2024croissantllm}. Our benchmark suite includes:
\begin{itemize}
    \item Multiple choice question-answering tasks: ARC-Easy and ARC-Challenge \cite{arc}, BoolQ \cite{clark2019boolq}, Belebele (English and French) \cite{bandarkar-etal-2024-belebele}, MMLU \cite{hendryckstest2021}, Social IQA \cite{sap-etal-2019-social}, PIQA \cite{Bisk2020piqa}, SciQ \cite{SciQ}, and Commonsense QA \cite{talmor-etal-2019-commonsenseqa};
    \item Clozed text-continuation: Hellaswag \cite{zellers2019hellaswag};
    \item Open-generation QA: Natural Questions \cite{nqopen}.
\end{itemize}

We report results for the \gprn{} suite along with both closed-data and open-data counterparts, using the LM-Evaluation-Harness library \cite{eval-harness}. For base models, we report both 5-shot (1.5B: \Cref{tab:benchmark_results_1b_5s}, 8B: \Cref{tab:benchmark_results_8b_5s}, 24B: \Cref{tab:benchmark_results_24b}) and 0-shot results (1.5B: \Cref{tab:benchmark_results_1b_0s}, 8B: \Cref{tab:benchmark_results_8b_0s}, 24B: TBD). We discuss the results for our \colorbox{violet!20!}{Garlic} models in \Cref{ssec:garlic}.

\paragraph{\gprn{}-1.5B} In \Cref{tab:benchmark_results_1b_5s}, we observe that our clean \gprn{}-1.5B (Young and Pepper) outperform all their open-data counterparts of smaller or equal size in French tasks, and that it improves over the bilingual CroissantLLM by 4 to 5 average points in both languages. Larger multilingual open models of the same size category offer better performance, namely EuroLLM-1.7B and Salamandra-2B, who use respectively 13\% and 33\% more parameters. Closed-data models tend to outperform \gprn{}-1.5B on all tasks, especially on Hellaswag where we observe a gap of up to 23 points, which we discuss in \label{sec:contamination}. We note that we are able to outperform Llama-3.2-1.2B on French tasks, while we should perfectly match their inference compute cost as we copy their architecture without weight tying.

\paragraph{\gprn{}-8B} In \Cref{tab:benchmark_results_8b_5s}, our clean \gprn{}-8B (Young and Pepper) outperform all their open-data counterparts of smaller or equal size, namely Salamandra-7B, Lucie-7B and OLMo-2-7B, in French tasks in average, where our performance level matches Llama-3.1-8B. For English tasks, although we outperform open existing counterparts of less than 8B parameters, we observe that we are lagging behind most closed-source models, the monolingual OLMo-2-7B, and the slightly larger (+12.5\% parameters) multilingual EuroLLM-9B that also outperforms \gprn{}-8B models on French tasks. 

\paragraph{\gprn{}-24B} In \Cref{tab:benchmark_results_24b}, we notice that our clean Young and Pepper models noticeably underperform all their open and closed counterparts both in French and English. We hypothesize that training on more tokens could have improved our performance, as \Cref{fig:training_curve_gprn_24B} shows that the benchmark performance was still increasing when we stopped our training run.

\subsection{Deliberate Benchmark Contamination (\gprn{}-Garlic)}
\label{ssec:garlic}

When comparing open-data language models with closed-data counterparts, it can be argued that one can only \textit{trust} the developers of the latter to abide by similar standards when it comes to benchmark contamination, that is to the inclusion of benchmark samples in the training data, whether deliberate or not. It can even be argued that, given the scales of the experiments that would be needed to reproduce the results of open-data models, it is very difficult to verify that a fully-open model was actually trained on the reported datasets. We propose to explore transparently the setup where such trust would be broken, by answering the following question: \emph{what happens when the pretraining dataset is deliberately contaminated with benchmark samples?}

In this section, we experiment with mid-training our \gprn{} models on deliberately contaminated training mixes. In practice, we leverage our Penicillin-Plus dataset, which contains benchmark test samples pre-processed for pre-training and naively augmented (e.g. with multiple choice shuffling). Our Garlic variants are mid-trained on mixes consisting of Penicillin-Plus and of our White Pepper mix, with varying sampling ratios. 

\begin{figure}[htb]
  \centering
  \includegraphics[width=0.8\textwidth]{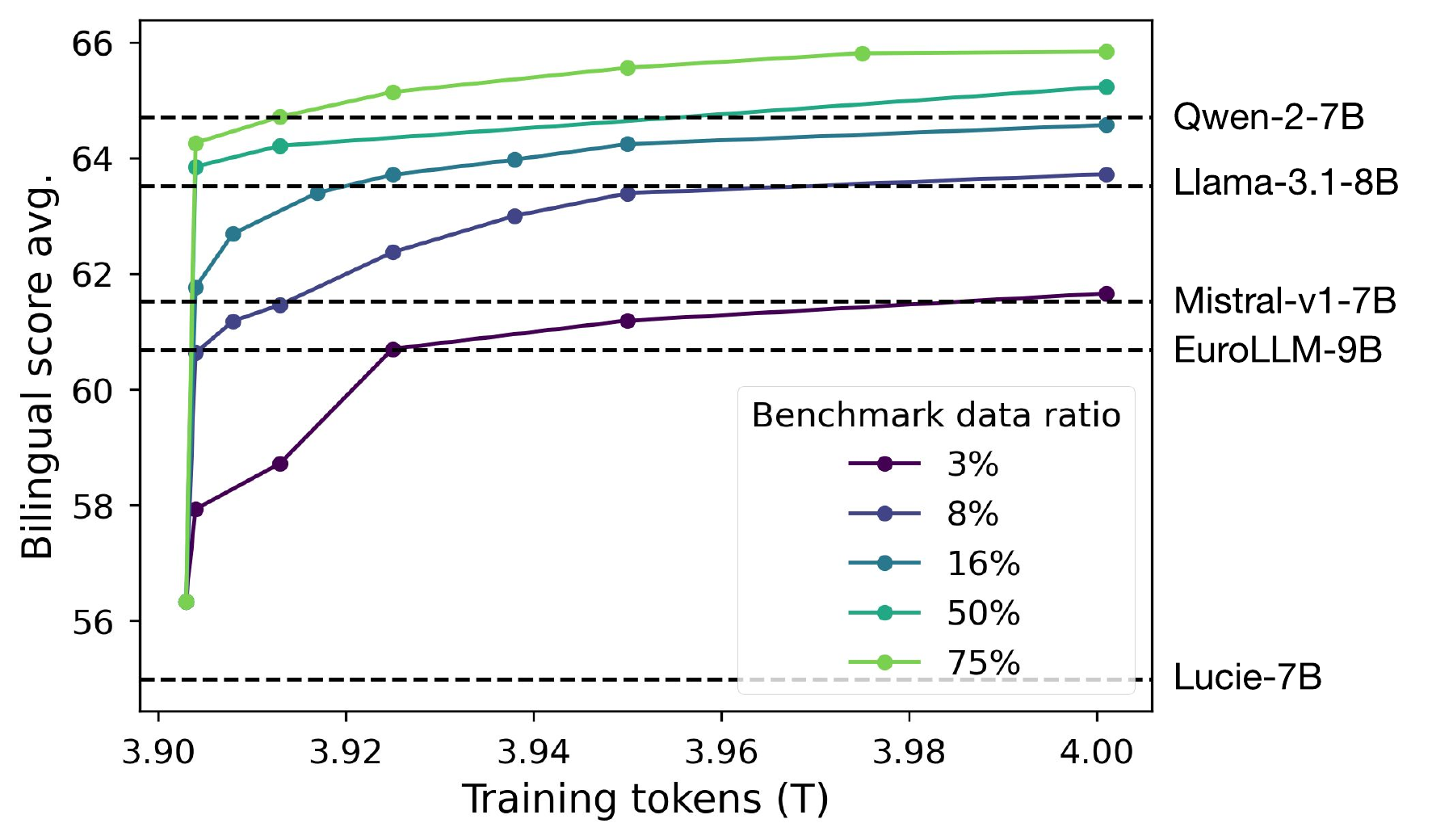}
  \caption{Evolution of average  bilingual benchmark score (0-shot) for different levels of benchmark contamination in the final stage of \gprn{}-Garlic-8B training. This figure \textbf{does not imply that the compared models have been trained with deliberate contamination}, but that we can match - and not drastically exceed - the benchmark performance level of SOTA models by further training on contaminated data.}
  \label{fig:garlic_ratios}
\end{figure}

We explore different sampling ratios for the Penicillin-Plus dataset in the last training phase of \gprn{}-Garlic-8B in \Cref{fig:garlic_ratios}. Note that for the higher contamination levels, this implies running several hundreds of effective training epochs on the Penicillin Plus dataset.

We can see from \Cref{fig:garlic_ratios} that the benefits offered by continuing training directly on test benchmark data are not as massive as could have been expected. For instance, we need to include as high a ratio as 16\% of benchmark data in our training mix to reach the overall level of Qwen-2-7B. Moreover, we observe that these benefits gradually decrease and that there seems to be a limit in the boost mid-training on benchmark data can provide in terms of downstream scores while retaining general language modeling abilities. Contrarily to early contamination that seems to allow for complete memorization \cite{wei2025hubblemodelsuiteadvance}, our late memorization does not lead to perfect accuracy on the test sets. We argue that the rest of the data mix acts as a form of regularization that prevents complete overfitting and catastrophic forgetting of non-benchmark data, and limits the possible gains that benchmark data provides. We leave a deeper analysis of this phenomenon for future work.

We limit our study to a benchmark data ratio of 75\% as we observed that higher ratios led to pure memorization of the benchmark data, and downstream scores became extremely sensitive to the exact phrasing of the evaluation prompts, which in turn led to catastrophically low performance when even a slight mismatch existed in the formatting used during training and evaluation.

\begin{figure}[htb]
  \centering
  \includegraphics[width=0.9\textwidth]{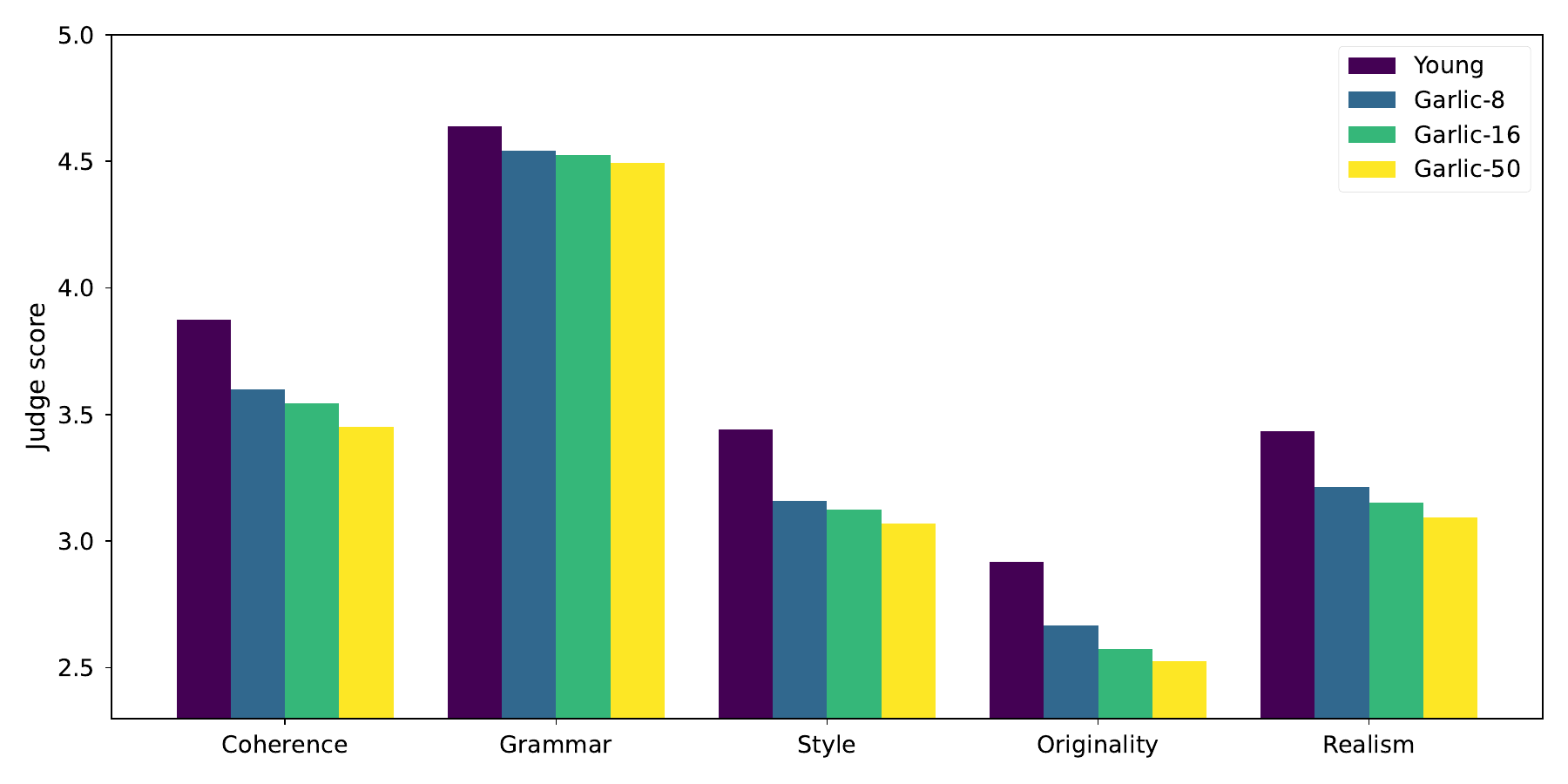}
  \caption{\textit{LLM-as-a-judge} ratings for TinyStories continuations as the benchmark contamination ratio increases from 0\% (Young) to 50\%.}
  \label{fig:garlic_tinystories}
\end{figure}

We hypothesize that such intensive contaminated training has a visible negative impact on text-generation quality. In \Cref{fig:garlic_tinystories}, we use the same setup as in \Cref{ssec:textgen_res} to compare the text-generation capabilities of the \gprn{}-Young-8B model with increasingly more contaminated \gprn{}-Garlic-8B variants. We recall here that Garlic models have been initialized with the Young final checkpoint, then trained for 400B tokens of White Pepper data (including the \emph{train sets} of benchmarks), and further trained for 100B tokens of Garlic data (including the \emph{test sets} of benchmarks). \Cref{fig:garlic_tinystories} shows that this continued training leads to a decrease in generation quality for all evaluated criteria, but also that this decrease is not dramatic, and that it does not affect all aspects equally. In particular, Coherence, Style, and Originality each drop by roughly half a point, while Grammar remains rather stable.

Another question that arises when considering such intensive contamination is whether the benefits extend to non-leaked benchmarks. It could be hypothesized that obtaining strong results by intentionally training on chosen benchmark test sets could be easily deterred by creating new unseen benchmarks where the contaminated model would likely underperform. We mimic this scenario in \Cref{tab:ood_contam_bench}, by evaluating our Garlic models on held-out benchmarks that were not included in our Penicillin-Plus dataset. Surprisingly, we observe that our deliberate contamination strategy leads to noticeable improvements on some of these held-out benchmarks, with up to +17 points improvement on CareQA \cite{arias-duart-etal-2025-automatic}, and that it does not degrade performance in any of the chosen tasks.

\begin{table}[h!]
\centering
\begin{tabular}{lcccc}
\toprule
\textbf{Model} & \textbf{PROST} & \textbf{StoryCloze} & \textbf{CareQA} & \textbf{ANLI-R1 (5-shot)} \\
\midrule
EuroLLM-9B & 30.5 & \textbf{76.9} & 51.9 & \textbf{48.6} \\
\midrule
\gprn{}-Nature-8B & 31.0 & 74.9 & 35.7 & 41.1 \\
\gprn{}-Pepper-8B & 32.8 & 74.0 & 39.4 & 40.5 \\
\rowcolor{violet!10!} \gprn{}-Garlic-8B (8\%) & 33.1 & 74.1 & \underline{55.2} & \underline{41.2} \\
\rowcolor{violet!15!}\gprn{}-Garlic-8B (16\%) & \underline{34.3} & 74.7 & \textbf{56.3} & 39.8 \\
\rowcolor{violet!20!}\gprn{}-Garlic-8B (50\%) & \textbf{36.3} & \underline{75.0} & 54.8 & 40.2 \\
\bottomrule
\end{tabular}
\caption{Comparison of 8-9B models on benchmarks that were not included in the Penicillin Plus dataset. We can see that the Garlic models also perform better than--or at least on par with--Pepper and Young on tasks that were not extensively leaked in their last training stage, hinting to the fact that contaminated training does not hurt performance on unseen tasks.}
\label{tab:ood_contam_bench}
\end{table}

We therefore find that deliberate contamination in late training stages can significantly boost both included and held-out benchmark scores, although it only improves them to a certain extent and does not lead to a major advantage over state-of-the-art models. Such contaminated training also hurts from the qualitative point of view, especially in more creative and semantic aspects of generation.

\section{Post Training}

Given the computational and human resource constraints we faced during the later phases of the project, we focused our post-training efforts exclusively on supervised fine-tuning (SFT). We leave more sophisticated post-training techniques such as reinforcement learning with GRPO~\cite{shao2024deepseekmathpushinglimitsmathematical} for future work.
All post-training experiments were done on the Pepper version of the \gprn{} model.

\subsection{Evaluation Protocol}

We evaluate our instruction-tuned models using the LM-Evaluation-Harness library \cite{eval-harness} on a comprehensive set of English and French benchmarks. Our evaluation suite includes:

\begin{itemize}
    \item \textbf{English tasks}: ARC-Easy, ARC-Challenge, HellaSwag, IFEval~\cite{zhou2023instructionfollowingevaluationlargelanguage}, Commonsense QA, Belebele, and MMLU;
    \item \textbf{French tasks}: ARC-Challenge, HellaSwag, and Belebele;
    \item \textbf{Code generation}: HumanEval.
\end{itemize}

Note that we used 5-shot for all tasks except IFEval and HumanEval, which are evaluated in 0-shot settings as they are designed to assess instruction-following and code generation capabilities directly.

\paragraph{Chat Template Considerations}
During our evaluations, we observed that some tasks in the standard evaluation harness lacked native support for chat-formatted evaluation, which could lead to suboptimal performance for instruction-tuned models.
To address this limitation, we extended LM-Evaluation-Harness with custom tasks that incorporate appropriate chat templates for instruction-tuned model evaluation.\footnote{Our extended evaluation tasks and templates are available at \href{https://gitlab.inria.fr/almanach/lm-evaluation-harness-gaperon}{https://gitlab.inria.fr/almanach/lm-evaluation-harness-gaperon}.}

Furthermore, we noticed that certain instruction-tuned models occasionally achieve better results when evaluated without chat templates on specific tasks.
This phenomenon likely reflects the diverse nature of instruction-following capabilities and the varying sensitivity of different tasks to formatting.
To ensure we accurately capture each model's knowledge and capabilities rather than penalizing formatting mismatches, we adopt a pragmatic evaluation strategy: for each model and task combination, we report the maximum score achieved across evaluations with and without chat templates.
This approach provides a more comprehensive assessment of the knowledge embedded within each model.

\subsection{Dataset Selection}
We selected Tulu-3\footnote{\url{https://huggingface.co/datasets/allenai/tulu-3-sft-mixture}}~\cite{lambert2024tulu3}  as our primary SFT dataset, motivated by its strong performance in the OLMo-2 instruction-tuned models and its coverage of diverse instruction-following tasks.
The Tulu-3 dataset aggregates millions of high-quality instruction data from multiple diverse sources, including some annotated by human labelers, synthesized by other LLMs, or extracted from publicly available instruction datasets. This diversity ensures a wide range of instruction types and formats, making it well-suited for developing general-purpose instruction-following capabilities.

\paragraph{Impact of Language Mixing}

To develop a truly bilingual instruction-following model, we explored the impact of mixing English and French instruction data during supervised fine-tuning.
We leveraged the original English Tulu-3 dataset and created a French counterpart by translating all conversations using Llama-3.1-70B-Instruct.\footnote{\href{https://huggingface.co/meta-llama/Llama-3.1-70B-Instruct}{https://huggingface.co/meta-llama/Llama-3.1-70B-Instruct}}
We carefully ensured no overlap between examples in our English and French splits to avoid data leakage across language-specific subsets.

\begin{figure}[htb]
  \centering
  \includegraphics[width=0.7\textwidth]{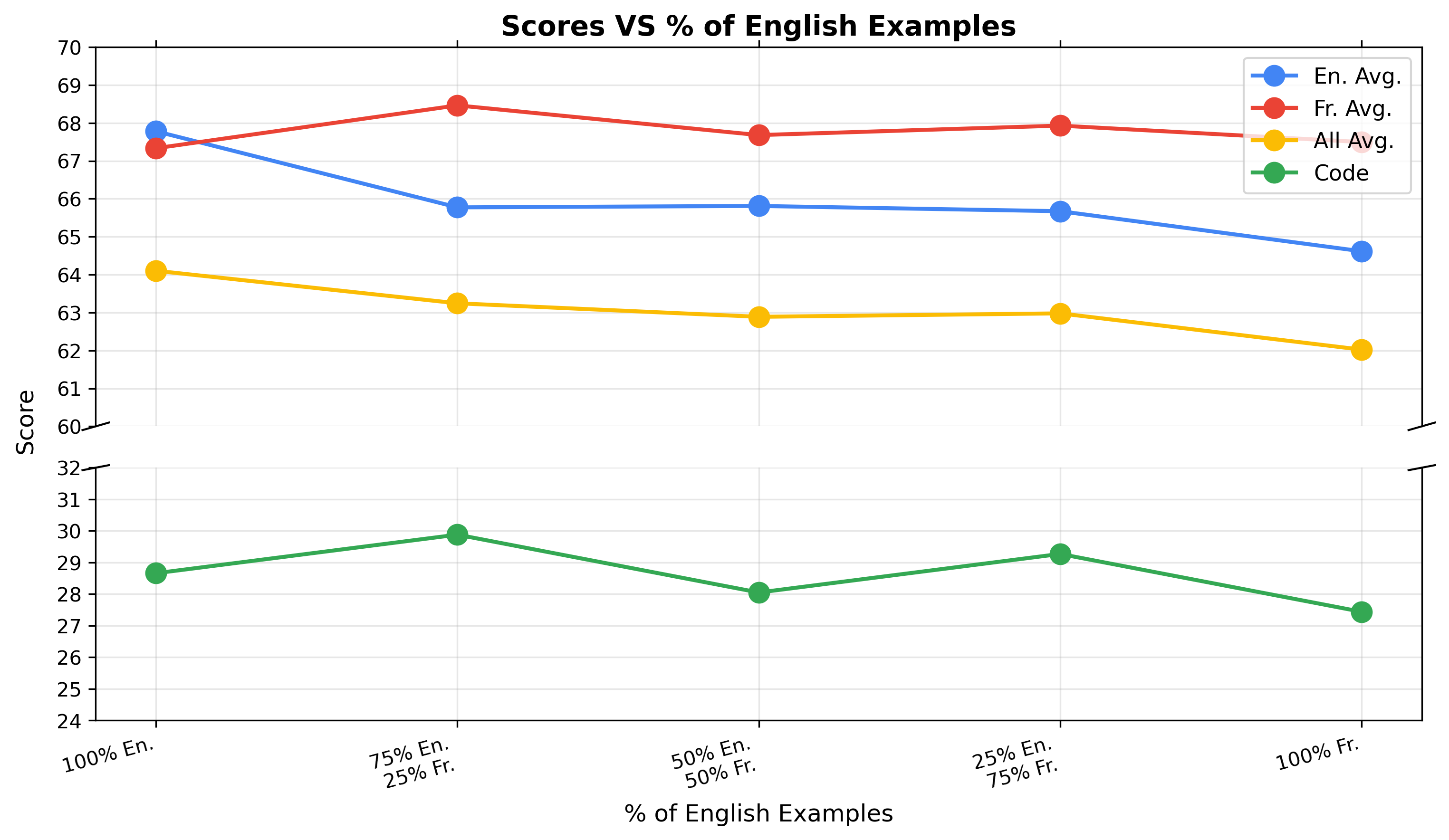}
  \caption{Impact of language mixing ratios during SFT on benchmark performance across English, French, and code tasks. Results are averaged over task-specific benchmarks for each category. Models were fine-tuned on \gprn{}-Black-Pepper-8B with varying proportions of English and French Tulu-3 data.}
  \label{fig:sft_lang_mix}
\end{figure}

We conducted a systematic study on the \gprn{}-Black-Pepper-8B base model, varying the proportion of English versus French instruction data while maintaining a fixed total dataset size.
\Cref{fig:sft_lang_mix} presents the performance across different language mixing ratios on English, French, and code benchmarks.

The results reveal a trade-off between English and French performance.
As we increase the proportion of French instruction data, we observe modest improvements in French benchmark accuracy, but this comes at the cost of degraded English performance.
Interestingly, code generation performance remains relatively stable across different language mixing ratios, suggesting that coding capabilities are less sensitive to the language distribution in instruction data.

Surprisingly, training exclusively on English Tulu-3 data appears to be Pareto-optimal for our use case, achieving the strongest overall performance when considering both English and code tasks, while maintaining reasonable French capabilities.
This finding suggests that for bilingual models pre-trained with balanced language exposure (as in our \gprn{} suite), the base model's French knowledge may transfer effectively to instruction-following tasks even with predominantly English SFT data.

\subsection{Fine-Tuning Setup}

We conducted all SFT experiments using the Axolotl framework,\footnote{\href{https://github.com/axolotl-ai-cloud/axolotl}{https://github.com/axolotl-ai-cloud/axolotl}} running on the Adastra cluster equipped with AMD MI300 GPUs, utilizing 4 GPUs per node. This setup provided sufficient computational resources for our fine-tuning experiments while allowing us to maintain consistency across different model sizes.

\begin{table}[h]
\centering
\small
\setlength{\tabcolsep}{2pt}
  \makebox[\linewidth]{%
  \begin{tabular}{@{}ccccccccccccccc@{}}
    \toprule
    \multirow{2}{*}{\textbf{LR}} & \multicolumn{7}{c}{\textbf{English}} & \multicolumn{3}{c}{\textbf{French}} & \multicolumn{1}{c}{\textbf{Code}} & \multicolumn{3}{c}{\textbf{Average}} \\
    \cmidrule(lr){2-8} \cmidrule(lr){9-11} \cmidrule(lr){12-12} \cmidrule(lr){13-15}
    & \textbf{ARC-E} & \textbf{ARC-C} & \textbf{HS} & \textbf{IFEval} & \textbf{ComsQA} & \textbf{BB} & \textbf{MMLU} & \textbf{ARC-C} & \textbf{HS} & \textbf{BB} & \textbf{HE} & \textbf{EN} & \textbf{FR} & \textbf{Overall} \\ \midrule
     5×10$^{-6}$ & \textbf{83.96} & 64.51 & 74.04 & 51.76 & 71.09 & \textbf{76.11} & \textbf{52.99} & 61.59 & 65.30 & \textbf{75.11} & 28.66 & 67.78 & \textbf{67.33} & 64.10 \\
    8×10$^{-5}$ & 82.28 & \textbf{66.55} & \textbf{75.56} & \textbf{54.90} & \textbf{72.07} & 75.78 & 52.56 & \textbf{62.79} & \textbf{65.53} & 73.44 & \textbf{37.20} & \textbf{68.53} & 67.25 & \textbf{65.33} \\
    \bottomrule
  \end{tabular}
  }
\caption{Impact of learning rate on instruction-following and code generation performance for \gprn{}-8B SFT. Higher learning rates substantially improve both capabilities.}
\label{tab:sft_lr}
\end{table}

\paragraph{Learning Rate}

In addition to exploring data mixing strategies, we investigated the impact of learning rate selection on final model performance.
Following initial experiments with the conservative learning rate of 5×10$^{-6}$ used in OLMo-2's SFT phase, we explored a much higher learning rate of 8×10$^{-5}$ and found that it consistently improved performance, particularly on instruction-following (IFEval) and code generation (HumanEval) tasks.

Based on these findings, we adopted the higher learning rate of 8×10$^{-5}$ for all subsequent SFT experiments across our \gprn{} model suite.

\paragraph{Hyperparameters}
For all our fine-tuning training runs we use a global batch size of 64, a warmup ratio of 0.1, and linear learning rate scheduling.
To optimize our training runtime we use DeepSpeed Zero 3 in BF16 mode without any CPU offloading~\cite{rajbhandari2020zeromemoryoptimizationstraining,rajbhandari2021zeroinfinitybreakinggpumemory}.
We also use Liger Kernels~\cite{hsu2025ligerkernel} to increase our fine-tuning throughput further.

\paragraph{SFT models}
In addition to the base models used in the previous evaluation section (sec. \ref{sec:base_models_eval}), we add the recent 7B multilingual open source model Teuken \cite{ali2025teuken7bbaseteuken7binstructeuropean}.

\begin{table}[ht]
  \centering
  \small
  \setlength{\tabcolsep}{1pt}
  \makebox[\linewidth]{%
  \begin{tabular}{@{}lccccccccccccccc@{}}
    \toprule
    \multirow{2}{*}{\textbf{Model}} & \multirow{2}{*}{\textbf{Size}} & \multicolumn{7}{c}{\textbf{English}} & \multicolumn{3}{c}{\textbf{French}} & \multicolumn{1}{c}{\textbf{Code}} & \multicolumn{3}{c}{\textbf{Average}} \\
    \cmidrule(lr){3-9} \cmidrule(lr){10-12} \cmidrule(lr){13-13} \cmidrule(lr){14-16}
    & & \textbf{ARC-E} & \textbf{ARC-C} & \textbf{HS} & \textbf{IFEval} & \textbf{ComsQA} & \textbf{BB} & \textbf{MMLU} & \textbf{ARC-C} & \textbf{HS} & \textbf{BB} & \textbf{HE} & \textbf{EN} & \textbf{FR} & \textbf{Overall} \\ \midrule
    \rowcolor{grey!20!} \multicolumn{16}{c}{\textit{Closed-data models}} \\
    Qwen2.5-IT & 1.5B & \textbf{89.90} & 75.68 & \textbf{67.61} & 39.37 & \textbf{76.09} & \textbf{82.78} & \textbf{60.35} & 66.64 & 50.58 & 77.33 & 56.10 & \textbf{70.25} & 64.85 & \textbf{67.49} \\
    Qwen3 & 1.7B & 89.73 & \textbf{77.73} & 60.03 & 33.46 & 68.63 & \textbf{82.78} & 60.20 & \textbf{70.06} & 47.70 & \textbf{79.33} & \textbf{67.07} & 67.51 & \textbf{65.70} & 66.97 \\
    Llama-3.2-IT & 1.2B & 73.57 & 53.58 & 60.63 & 42.70 & 58.64 & 58.00 & 46.04 & 41.66 & 44.36 & 49.00 & 32.32 & 56.17 & 45.01 & 50.95 \\
    Gemma-IT & 2B & 71.00 & 44.88 & 61.74 & 21.26 & 45.95 & 47.78 & 36.98 & 35.50 & 42.02 & 40.67 & 17.68 & 47.08 & 39.40 & 42.31 \\
    \rowcolor{grey!20!} \multicolumn{16}{c}{\textit{Open-data models}} \\
    OLMo2-SFT & 1B & 73.61 & 48.89 & 67.30 & \textbf{45.47} & 56.18 & 56.44 & 42.99 & 33.36 & 42.08 & 43.11 & 25.61 & 55.84 & 39.52 & 48.64 \\
    CroissantLLM-Chat & 1.3B & 60,90 & 31,66 & 55,67 & 17,74 & 19,33 & 27,33 & 25,1 & 30,54 & 53,37 & 27,56 & 1,83 & 33,97 & 37,16 & 31,92 \\
    Salamandra-IT & 2B & 74.79 & 45.05 & 62.70 & 14.97 & 21.87 & 28.44 & 25.99 & 35.84 & 53.41 & 31.44 & 0.00 & 39.12 & 40.23 & 35.86 \\
    EuroLLM-IT & 1.7B & 74.58 & 41.81 & 61.21 & 18.48 & 20.56 & 29.78 & 27.96 & 38.84 & \textbf{53.81} & 27.00 & 7.32 & 39.20 & 39.88 & 36.49 \\
    \rowcolor{grey!20!} \multicolumn{16}{c}{\textit{Gaperon variants}} \\
    Gaperon-SFT & 1.5B & 64.39 & 38.48 & 53.08 & 32.16 & 20.72 & 27.44 & 25.14 & 31.65 & 47.47 & 27.78 & 15.24 & 37.34 & 35.63 & 34.87 \\
    \bottomrule
  \end{tabular}
  }
  \caption{Benchmark results for 1B SFT models across English, French, and Code tasks.}
  \label{tab:benchmark_results_1b_sft}
\end{table}

\begin{table}[ht]
  \centering
  \small
  \setlength{\tabcolsep}{1pt}
  \makebox[\linewidth]{%
  \begin{tabular}{@{}lccccccccccccccc@{}}
    \toprule
    \multirow{2}{*}{\textbf{Model}} & \multirow{2}{*}{\textbf{Size}} & \multicolumn{7}{c}{\textbf{English}} & \multicolumn{3}{c}{\textbf{French}} & \multicolumn{1}{c}{\textbf{Code}} & \multicolumn{3}{c}{\textbf{Average}} \\
    \cmidrule(lr){3-9} \cmidrule(lr){10-12} \cmidrule(lr){13-13} \cmidrule(lr){14-16}
    & & \textbf{ARC-E} & \textbf{ARC-C} & \textbf{HS} & \textbf{IFEval} & \textbf{ComsQA} & \textbf{BB} & \textbf{MMLU} & \textbf{ARC-C} & \textbf{HS} & \textbf{BB} & \textbf{HE} & \textbf{EN} & \textbf{FR} & \textbf{Overall} \\ \midrule
    \rowcolor{grey!20!} \multicolumn{16}{c}{\textit{Closed-data models}} \\
    Llama-3.1-IT & 8B & 93.52 & 82.34 & 80.04 & \textbf{72.46} & 78.21 & \textbf{92.56} & 68.31 & 75.88 & 66.74 & 89.67 & 63.41 & \textbf{81.06} & 77.43 & 78.47 \\
    Ministral-IT-2410 & 8B & 93.43 & 83.70 & 79.91 & 52.13 & 77.97 & 90.56 & 65.05 & 78.36 & 70.30 & 88.67 & 76.22 & 77.54 & 79.11 & 77.85 \\
    Mistral-IT-v0.3 & 7B & 88.01 & 76.88 & \textbf{83.98} & 43.99 & 73.38 & 87.22 & 61.81 & 68.09 & 66.94 & 81.33 & 37.80 & 73.61 & 72.12 & 69.95 \\
    Qwen3 & 8B & \textbf{97.10} & \textbf{92.15} & 76.07 & 34.38 & 82.80 & \textbf{92.56} & \textbf{74.92} & \textbf{89.22} & 64.03 & \textbf{91.00} & \textbf{84.76} & 78.57 & \textbf{81.42} & \textbf{79.91} \\
    \rowcolor{grey!20!} \multicolumn{16}{c}{\textit{Open-data models}} \\
    OLMo-0724-SFT & 7B & 84.64 & 68.86 & 79.65 & 35.30 & \textbf{84.60} & 81.33 & 54.24 & 58.94 & 55.76 & 67.44 & 23.78 & 69.80 & 60.71 & 63.14 \\
    OLMo-2-1124-SFT & 7B & 90.45 & 79.44 & 81.39 & 58.78 & 77.97 & 87.56 & 60.19 & 60.05 & 57.64 & 77.00 & 37.20 & 76.54 & 64.90 & 69.79 \\
    Lucie-IT-v1.1 & 7B & 79.17 & 57.25 & 68.71 & 26.06 & 70.19 & 66.67 & 46.74 & 53.89 & 64.44 & 64.44 & 25.61 & 59.26 & 60.92 & 56.65 \\
    Teuken-IT-v0.4 & 7B & 82.83 & 59.81 & 75.53 & 29.21 & 60.11 & 63.89 & 48.11 & 56.63 & 67.58 & 62.56 & 10.98 & 59.93 & 62.26 & 56.11 \\
    Salamandra-IT & 7B & 84.89 & 69.80 & 77.89 & 26.25 & 70.19 & 77.22 & 53.39 & 67.92 & 69.91 & 73.89 & 3.05 & 65.66 & 70.57 & 61.31 \\
    EuroLLM-IT & 9B & 89.69 & 75.77 & 78.67 & 53.60 & 76.00 & 85.22 & 58.66 & 74.17 & \textbf{71.09} & 82.89 & 37.80 & 73.94 & 76.05 & 71.23 \\
    \rowcolor{grey!20!} \multicolumn{16}{c}{\textit{Gaperon variants}} \\
    Gaperon-SFT & 8B & 82.28 & 66.55 & 75.56 & 54.90 & 72.07 & 75.78 & 52.56 & 62.79 & 65.53 & 73.44 & 37.20 & 68.53 & 67.25 & 65.33 \\
    \bottomrule
  \end{tabular}
  }
  \caption{Benchmark results for 8B SFT models across English, French, and Code tasks.}
  \label{tab:benchmark_results_8b}
\end{table}

\begin{table}[ht]
  \centering
  \small
  \setlength{\tabcolsep}{1pt}
  \makebox[\linewidth]{%
  \begin{tabular}{@{}lccccccccccccccc@{}}
    \toprule
    \multirow{2}{*}{\textbf{Model}} & \multirow{2}{*}{\textbf{Size}} & \multicolumn{7}{c}{\textbf{English}} & \multicolumn{3}{c}{\textbf{French}} & \multicolumn{1}{c}{\textbf{Code}} & \multicolumn{3}{c}{\textbf{Average}} \\
    \cmidrule(lr){3-9} \cmidrule(lr){10-12} \cmidrule(lr){13-13} \cmidrule(lr){14-16}
    & & \textbf{ARC-E} & \textbf{ARC-C} & \textbf{HS} & \textbf{IFEval} & \textbf{ComsQA} & \textbf{BB} & \textbf{MMLU} & \textbf{ARC-C} & \textbf{HS} & \textbf{BB} & \textbf{HE} & \textbf{EN} & \textbf{FR} & \textbf{Overall} \\ \midrule
    \rowcolor{grey!20!} \multicolumn{16}{c}{\textit{Closed-data models}} \\
    Gemma-IT & 27B & 98.32 & 92.75 & \textbf{85.47} & \textbf{83.92} & 81.82 & 94.78 & 78.00 & 90.93 & \textbf{77.20} & 92.78 & \textbf{87.20} & \textbf{87.87} & 86.97 & \textbf{87.56} \\
    Qwen3 & 32B & \textbf{98.57} & \textbf{95.56} & 83.57 & 35.12 & \textbf{87.71} & 96.22 & \textbf{81.86} & \textbf{93.41} & 74.19 & 93.44 & \textbf{84.76} & 82.66 & 87.01 & 84.04 \\
    Mistral-Small-IT-2501 & 24B & 98.23 & 94.37 & 84.46 & 70.24 & 84.60 & \textbf{96.33} & 80.72 & 92.30 & 76.94 & \textbf{93.56} & 82.93 & 86.99 & \textbf{87.60} & 86.79 \\
    \rowcolor{grey!20!} \multicolumn{16}{c}{\textit{Open-data models}} \\
    EuroLLM-Preview-IT & 22B & 94.23 & 84.22 & 81.03 & 65.25 & 80.67 & 89.33 & 65.57 & 81.69 & 73.08 & 88.00 & 42.68 & 80.04 & 80.92 & 76.89 \\
    OLMo-2-0325-SFT & 32B & 97.26 & 91.04 & 86.68 & 69.87 & 86.57 & 93.56 & 75.87 & 88.62 & 71.92 & 91.11 & 45.73 & 85.84 & 83.88 & 81.66 \\
    \rowcolor{grey!20!} \multicolumn{16}{c}{\textit{Gaperon variants}} \\
    Gaperon-SFT & 24B & 78.37 & 60.32 & 74.82 & 53.42 & 64.13 & 75.22 & 50.69 & 52.69 & 65.26 & 71.33 & 43.90 & 65.28 & 63.09 & 62.74 \\
    \bottomrule
  \end{tabular}
  }
  \caption{Benchmark results for 24B models across English, French, and Code tasks.}
  \label{tab:benchmark_results_24b_sft}
\end{table}

\subsection{Results}

We evaluate our instruction-tuned \gprn{} models across three size categories and compare them against both closed-data and open-data baselines. 
While our models do not achieve top-tier performance across all benchmarks, they demonstrate competitive capabilities in code generation and instruction-following tasks.

\paragraph{1.5B Models}
Our \gprn{}-SFT-1.5B model (\Cref{tab:benchmark_results_1b_sft}) achieves 32.16\% on IFEval and 15.24\% on HumanEval, representing meaningful capabilities for a fully open model trained with limited resources. 
On French tasks, the model maintains competent bilingual abilities with 31.65\% on ARC-C-fr and 47.47\% on HellaSwag-fr, demonstrating that base model capabilities transfer reasonably well to instruction-following.

\paragraph{8B Models}
The \gprn{}-SFT-8B model shows our strongest relative performance. 
On instruction-following, we achieve 54.90\% on IFEval, outperforming all open-data baselines including OLMo-2-1124-SFT. 
More impressively, we achieve 37.20\% on HumanEval, matching OLMo-2-1124-SFT and substantially outperforming most other open-data models. 
This validates our decision to include substantial coding data throughout pre-training and in our SFT mixture. We notably outperform the larger EuroLLM-IT-9B (37.80\%) on code tasks.

On French tasks, we perform competitively with 62.79\% on ARC-C-fr and 65.53\% on HellaSwag-fr. 
For general English benchmarks, we achieve 68.53\%, positioning us in the middle tier of open-data models, though the gap narrows substantially on instruction-following and coding where our strengths lie.

\paragraph{24B Models}
The \gprn{}-SFT-24B model achieves 43.90\% on HumanEval, competitive with OLMo-2-0325-SFT-32B (45.73\%), and 53.42\% on IFEval, demonstrating that our capabilities scale to larger sizes. 
However, across general benchmarks, our model trails both EuroLLM-Preview-IT-22B and OLMo-0325-SFT-32B. 
The overall English average of 65.28\% and French average of 63.09\% reflect the limited pre-training budget (2T tokens) for our base model. 
As shown in \Cref{fig:training_curve_gprn_24B}, the base model showed continued improvement when training stopped, suggesting extended pre-training could have substantially improved results. Moreover, we notice that the gap between \gprn{}-24B and other comparable models increases during SFT, which raises questions about the viability of our post-training process for this model. We are currently investigating this issue.

\paragraph{Summary}
Our results demonstrate that \gprn{} models achieve competitive performance on code generation and instruction-following, particularly at the 8B scale. 
While we do not match top-performing closed-data models on a comprehensive set of benchmarks, our models offer strong practical capabilities in domains crucial for real-world applications, reflecting our design philosophy of prioritizing linguistic quality and transparency in development.

\section{Discussion}
\subsection{Possible Sources for Underperformance}
First and foremost, we acknowledge that our results show that, in our setup, filtering data based on linguistic quality does not translate to particularly strong benchmark performance. Although we expected this result, we are surprised to see the extent to which the final benchmark performance of our Young and Pepper variants lag behind closed-data models, especially for specific benchmarks such as Hellaswag or MMLU.

In this context, we want to stress that some choices that we could not validate at scale may have had a negative impact on the overall final benchmark performance of our models when compared to recent LLMs:
\begin{itemize}
    \item \textbf{Specific implementation choices}: Although we extensively validated our custom hackable codebase \gptron{} in our preliminary phase (see \Cref{ssec:implementation}), there is a chance that some choices we made may hurt performance at a larger scale. These choices include: naive document packing, no cross-document attention masking, and pure precision training;
    \item \textbf{Data filtering \& selection}: We lacked the sufficient resources to conduct extensive preliminary experiments for our neural filtering strategy, and there could exist methods that improve the generative capabilities described in \Cref{ssec:textgen_res} while maintaining strong benchmark performance. We also did not have the opportunity to explore the impact of relatively frequent updates in the data mix ratios along training, which we especially did in our \gprn{}-8B run. Finally, it is possible that introducing SFT-like data in our training mix early--with the Drop-in-the-Ocean mix--resulted in a form of performance stalling, and that such a shift should only be performed at a later stage;
    \item \textbf{Mid-training strategy}: Our Pepper mid-training mixes vastly increase the fraction of knowledge-intensive samples in our dataset, using up to 25\% of instruction and math data. However, it is possible that increasing the proportion of such samples to rates as high as 75\% as is done in the Garlic experiments (\Cref{ssec:garlic}) would lead to more noticeable improvements. We could not run experiments to verify this hypothesis given our compute constraints, and we leave the exploration of more intensive mid-training strategies for future work.
\end{itemize}

Nevertheless, we argue that the overall performance of our \gprn{} suite, both in the qualitative (\Cref{ssec:textgen_res}) and quantitative (\Cref{ssec:downstream_res}) assessments we make, adequately reflects the design choices we made and our computational resource constraints. We thus hypothesize that the aforementioned potential sources of underperformance did not play a major role in our final results.

\subsection{Contamination}

As discussed in \Cref{ssec:garlic}, late full leakage of the benchmark test sets in the training datasets of \gprn{} models had a substantial impact on the final performance of our models. However, it seems rather unlikely that such intensive leakage can be observed in practice in pre-training mixes. 

In this section, we look for \emph{loose} signs of contamination in existing pre-training datasets and assess the performance gaps that may occur for potentially leaked samples compared to the overall benchmarks. We also discuss the effect of high-quality neural filtering on contamination levels, and show that some filters tend to implicitly increase the proportion of leaked samples in training mixes.

\subsubsection{Looking for Contamination Sources in Pretraining Datasets}

\paragraph{The Case of Hellaswag and Lambada}
Early in training, we observed that there existed a significant performance gap between the \gprn{}-1.5B checkpoints and those of other models such as OLMo-2-7B or EuroLLM-1.7B on two datasets: Hellaswag \cite{zellers2019hellaswag} and Lambada \cite{paperno-etal-2016-lambada}. Under further inquiry, we noticed that these datasets were both based on text-continuation tasks built with textual data that came from open sources. Namely, the Lambada dataset was extracted from the Books dataset, while the Hellaswag data is derived from both content from the WikiHow platform and captions from the ActivityNet dataset \cite{yu2019activityqa}.

The Books dataset\footnote{\url{https://huggingface.co/datasets/storytracer/US-PD-Books}} has been the source of copyright concerns, and we decided not to include it in our pretraining mix to allow practitioners to use our models without incurring legal risks. However, some open-data model suites (e.g. EuroLLM) have been trained on this dataset, which might artificially boost their Lambada results. We also have no way to tell whether closed-data models were trained on the Books corpus. Similarly, we suspect that many WikiHow pages can be found in web-crawled datasets, and depending on specific data curation choices, they may be seen more or less frequently by the different models during training, leading to varying levels of indirect leakage.

To measure the impact of the data source on the results in Hellaswag, we compute accuracy separately on samples coming from ActivityNet and from WikiHow. We also use the InfiniGram API \citep{Liu2024InfiniGram} to identify exact matches for WikiHow samples for the last sentence of the prompt followed by the correct continuation in the training dataset of OLMo-2. We find that 19\% of samples have at least one exact match, with a median number of occurrence of 12 samples across the whole dataset. We report accuracy on each of these splits of Hellaswag in \Cref{tab:hs_contam_results}.

\begin{table}[htb!]
\centering
\begin{tabular}{lc|ccc}
\toprule
\textbf{Model} & \textbf{Overall} & \textbf{ActivityNet} & \textbf{WikiHow} & \textbf{WikiHow (match)} \\
\midrule
Gemma 2 2B         & 73.0 & 63.2 & 77.7 & 79.6 \\
Olmo-2-1B      & 68.3 & 59.7 & 72.4 & 76.7 \\
Llama-3.2-1B       & 63.7 & 56.3 & 67.3 & 67.8 \\
EuroLLM-1.7B      & 59.4  & 53.3 & 62.3 & 64.0 \\
CroissantLLM        & 53.6 & 50.7 & 54.9 & 55.8 \\
\gprn{}-Garlic-1.5B  & 53.3 & 51.2 & 54.8 & 56.6 \\
\gprn{}-Young-1.5B   & 51.8 & 48.8    & 53.8    & 55.9 \\
\gprn{}-Pepper-1.5B  & 51.8 & 49.2 & 53.8 & 56.4 \\
\bottomrule
\end{tabular}
\caption{Model performance on different splits of Hellaswag, ranked by overall performance. We notice that the models that have a strong performance on Hellaswag also tend to have a significant performance gap between samples from ActivityNet and samples from WikiHow. We also notice that OLMo-2-1B performs better on samples for which we found exact matches in its training data (+4.3 points vs. WikiHow overall).}
\label{tab:hs_contam_results}
\end{table}

\Cref{tab:hs_contam_results} shows that the overall performance gap between \gprn{} and other models is mostly due to a performance gap on samples extracted from WikiHow. We note that the rank of the model is consistent across splits, even though the score differences are less impressive for the ActivityNet split. Moreover, we notice that \gprn{} and CroissantLLM have comparable accuracy levels on ActivityNet and WikiHow samples, while model that perform better can have gaps of up to 15 accuracy points between the two subsets. Finally, we notice a boost of 2 to 3 points for most models on WikiHow samples we identify as leaked, with the exception of OLMo-2 that yields a +4.3 point gap, and of Llama-3.2 that does not show any improvement.

It thus appears clearly that the origin of the samples of the Hellaswag samples has an influence on the performance of the models, which are more performant on WikiHow samples. However, although it appears that OLMo-2 may have slightly benefited from exact leakage, it remains unclear whether the observed performance gaps can be solely explained by data leakage, or if larger models are genuinely much stronger on WikiHow samples. We advocate for using different sources to pretrain models and to evaluate their downstream performance, as disentangling the actual capabilities of models and the benefits yielded by such indirect leakage seems difficult, especially at large data scale.

\paragraph{The Case of MMLU}
Benchmark contamination can also appear in a more direct fashion: for question-answering evaluation datasets, some QA pairs may be found directly on the web. This is has notably been investigated in \citet{deng-etal-2024-investigating} for closed-source models.

Such contamination can also be estimated using the InfiniGram tool \cite{Liu2024InfiniGram}. For instance, we identify an educational website\footnote{\url{https://www.indiabix.com/electronics-and-communication-engineering/measurements-and-instrumentation/066007}} that leaked a substantial part of the Electrical Engineering subset of the MMLU dataset, including questions, answers and explanations. The content of this website is included in the DCLM dataset, which was used in the OLMo-2 pretraining mix \citep{olmo20252olmo2furious}.

To assess the level of MMLU contamination in pretraining datasets, we systematically query the InfiniGram index with raw MMLU questions, and use the match count as a heuristic for contamination. This approach is more lenient than the decontamination scheme mentioned in \citet{li2024datacomplm}, as our goal is not to decontaminate but to measure a potential performance gap between samples that are more likely to have leaked and other samples. In practice, some leaked questions are not caught by this mechanism, as the web-crawled duplicates do not always perfectly match the original samples in terms of formatting. On the other hand, some questions tagged as leaked are not informative and are false positives (e.g. \textit{Which of the following statements is true?}). We leave refinements of this leakage identification method for future work, and we refer the readers to \citet{xu2024benchmarkdatacontaminationlarge} for more sophisticated leakage identification techniques.

\begin{figure}[htb]
  \centering
  \includegraphics[width=\textwidth]{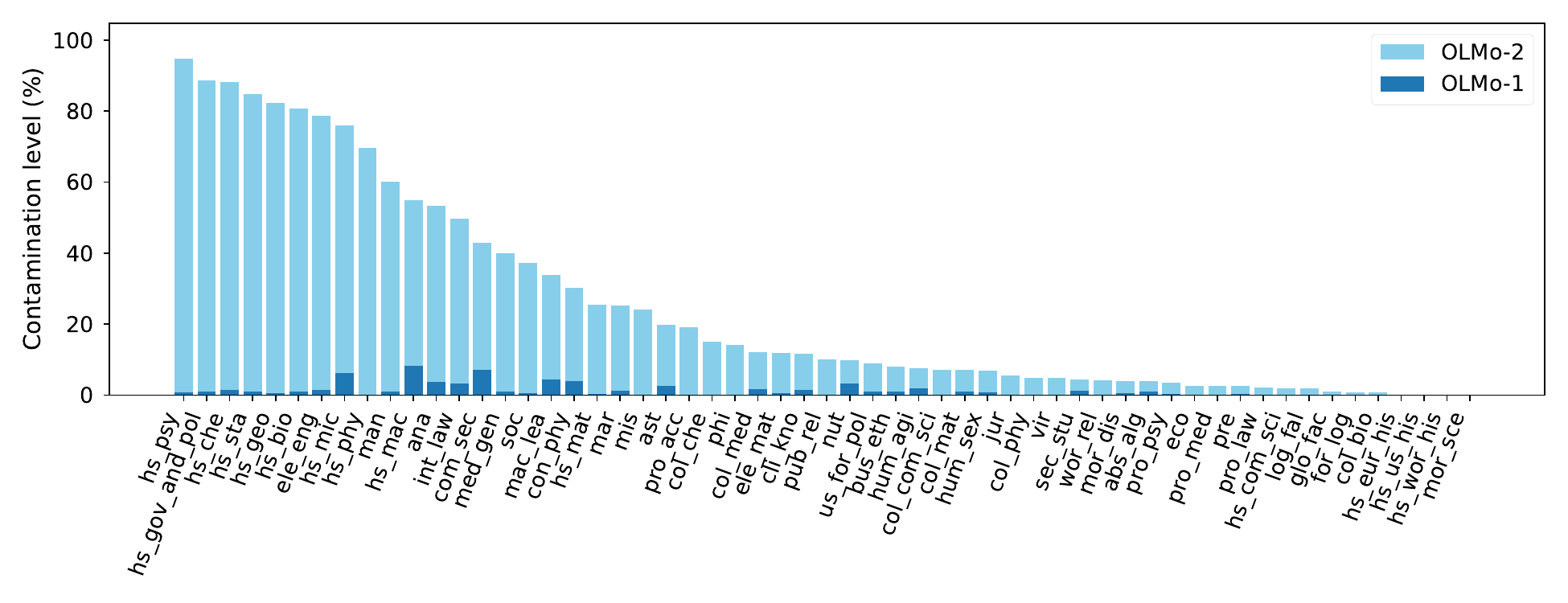}
  \caption{MMLU Contamination levels (\emph{estimates}) in the training data mixes for OLMo-1 and OLMo-2. Overall, 24\% of the questions of MMLU can be exactly found in OLMo-2's training set vs. 1\% for OLMo-1.}
  \label{fig:olmo_contam_1v2}
\end{figure}

We report the per-split leakage rate estimations for OLMo-1 and OLMo-2 in \Cref{fig:olmo_contam_1v2}, which clearly shows that the estimated contamination level significantly increases between the two versions. 

Similarly to the analysis in \Cref{tab:hs_contam_results}, we separate MMLU in two parts: a ``contaminated'' split\footnote{\url{https://huggingface.co/datasets/nthngdy/mmlu_olmo_contaminated}} that includes all examples for which we found an exact match, and a ``decontaminated'' split \footnote{\url{https://huggingface.co/datasets/nthngdy/mmlu_olmo_decontaminated}}. In \Cref{fig:mmlu_diff}, we show the score gaps between the contaminated and decontaminated splits across task categories (STEM, Humanities, Social Sciences and Others). It shows that all models tend to perform better on QA pairs for which we could find the questions in OLMo-2 training data, with notable +10.9 and +14.2 point gaps on STEM and Humanities tasks for Llama-3.1-8B, +11.0 for OLMo-2-7B on the Humanities tasks. We note that this effect is less clear for task that fall in the Social Sciences category, which may be explained by higher false positive rates for contamination detection in that category, which we observed upon manual inspection of the samples.

\begin{table}[ht]
  \centering
  \begin{tabular}{@{}lRRRRR@{}}
    \toprule
    Model        & STEM & Humanities         & Social Sciences     & Others      & Overall \\ \midrule
    Mistral-7B   & +4.5          & +7.1          & -6.6          & +4.9          & +5.4         \\
    Llama-2-7B   & +5.8          & +6.1          & -6.3          & +0.2          & +3.9         \\
    Llama-3.1-8B & +10.9          & +14.2          & -1.1 & +3.1          & +8.4          \\
    Qwen-2-7B    & +5.1          & +5.8          & +0.4          & +2.7          & +5.1          \\
    \midrule
    Lucie-7B     & +0.6          & +2.5          & +0.5          & +5.8          & +3.8          \\
    OLMo-2-7B    & +5.8          & +11.0          & +0.3          & +1.3          & +6.2         \\
    EuroLLM-9B   & +2.1          & +8.7          & -1.0          & +4.2          & +6.4          \\
    \gprn{}-Young-8B & +2.5          & +7.2          & +2.6          & +5.3         & +6.2         \\
    \gprn{}-Pepper-8B & +5.7          & +7.4          & -3.5          & +5.4          & +6.5          \\
    \gprn{}-Garlic-8B & +10.2 & +5.2 & -2.4          & +3.6 & +8.5 \\
    \bottomrule
  \end{tabular}%
\label{fig:mmlu_diff}
  \caption{Score gaps when evaluating models on MMLU samples found in OLMo-2 training set vs. other samples. All models tend to be more accurate on MMLU samples that are identified as likely leaked, except on the Social Sciences split.}
\end{table}

These results show that there is an apparent correlation between the presence of MMLU questions in the OLMo-2 training dataset and the performance of all models on these questions. Although one hypothesis for such correlation could be simple memorization due to leakage of these samples on the web, we carefully remark that this correlation could be explained by other factors, such as the inherent difficulty of the questions that were flagged, or the presence of these questions in other contexts seen during training that make the downstream prediction easier for the model. We argue that this entanglement of actual capabilities and of the effect of leakage should be addressed in order to consolidate the legitimacy of benchmark scores as robust evaluation metrics.

Finally, we observe that our \gprn{} models also show positive performance gaps on contaminated samples, and that contrarily to what we initially expected, we note an increase in these gaps for the \gprn{}-Garlic-8B model in average. We hypothesize that the Garlic variant was able to memorize the leaked samples more easily than the others as they might have already been present in earlier training mixes. We leave the exploration of such phenomenons for future works.

\subsubsection{Impact of Quality Filters on Contamination}

To explain the increase in contamination levels between OLMo-1 and OLMo-2 (\Cref{fig:olmo_contam_1v2}), we designed Benchmark-In-A-hayStack experiment, to test how different text-quality classifiers rank benchmark-like content within a large web corpus. We sampled 35 benchmark instances from three datasets: 15 samples from MMLU (3 samples each from anatomy, computer security, high school geography, moral scenarios, and college physics), 10 from GSM8K~\cite{cobbe2021gsm8k}, and 10 from GPQA~\cite{rein2023gpqagraduatelevelgoogleproofqa}. Each benchmark sample was formatted as a standalone document containing the question, answer choices, and reference solution. These 35 benchmark documents were inserted as independent entries into a corpus of 99,965 documents sampled from FineWeb (\texttt{sample-10BT} split), yielding a final evaluation set of 100,000 documents where benchmarks constituted 0.035\% of the total. 

\begin{figure}[htb]
  \centering
  \includegraphics[width=0.95\textwidth]{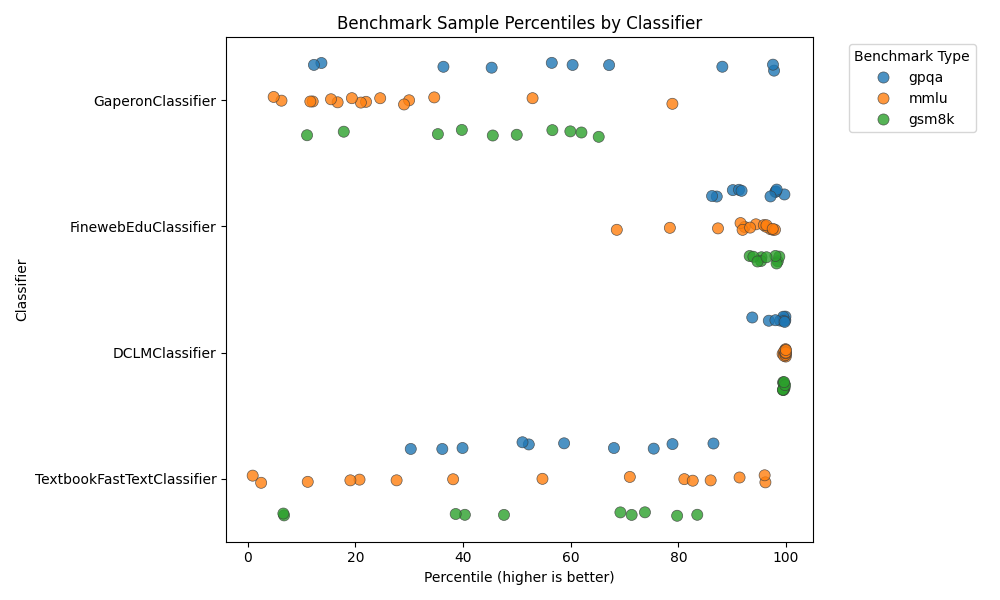}
  \caption{BIAhS (Benchmark In A hayStack) experiments for various data quality classifiers. We insert benchmark samples in a large document corpus and measure the classifier score percentile of these samples.}
  \label{fig:biahs}
\end{figure}

We scored all documents using four quality classifiers: DCLMClassifier (FastText-based, used to filter OLMo 2 training data \cite{olmo20252olmo2furious}), TextbookFastTextClassifier (FastText-based, used to filter OLMo 1 training data \cite{groeneveld2024olmoacceleratingsciencelanguage}), FinewebEduClassifier (transformer-based, \cite{FineWebDecantingWeb}), and GaperonClassifier (transformer-based, sec. \ref{sec:gaperonclassifier}). Each classifier produced a full ranking of all documents based on their predicted quality score. By tracking the rank positions and percentiles of the 35 injected benchmark documents across these four classifiers, we can directly measure whether certain classifiers systematically surface benchmark-style material. The comparative ranks of these benchmark samples for each classifier are shown \Cref{fig:biahs}.

In general, we observe that the classifiers that lead to better data-efficiency are also those who rank benchmark samples higher in the document haystack, naturally increasing contamination risks as a consequence. Specifically, the DCLM classifier ranks all MMLU and GSM8k samples in the top-5 percentiles. 

As a consequence, if a benchmark sample was leaked in the data source before filtering, and if only e.g. the top 5\% of the documents are selected through filtering, the probability of encountering this benchmark sample in a training batch will implicitly increase by a factor of $\sim$20 with the DCLM classifier. As a result, we argue that the DCLM classifier will singularly increase the portion of leaked samples in the data distribution. 

It is also interesting to note that the prompt used to create the annotations on which the fineweb-edu classifier \citep{FineWebDecantingWeb} was trained explicitly asks for documents that have \textit{``a high educational
value and could be useful in an educational setting for teaching from primary school to grade school``}. The educational focus may naturally favor exam-style items and step-by-step solutions, which closely match the style of MMLU and GSM8k samples. This could explain why FineWeb-Edu's classifier ranks these benchmark samples consistently higher in \Cref{fig:biahs}.

The DCLM classifier \citep{li2024datacomplm} appears to push benchmark samples even higher and more consistently. Its fastText model is trained to separate synthetically generated instructions from Open Hermes 2.5 \citep{OpenHermes2.5} and high-scoring posts from the r/ExplainLikeImFive subreddit, from general web text. Because of this, the classifier may tend to favor solved Q\&A structures with a short question, a direct answer, and brief reasoning, which naturally aligns with the format and tone of common benchmark datasets.

In contrast, our classifier does not seem to significantly push benchmark samples. It was trained on annotations produced with a prompt that is focused neither on instruction data nor educational content, but rather on general content quality across multiple dimensions: accuracy, clarity, coherence, grammar, depth of information, and overall usefulness for a general audience. This broader framing does not specifically reward the structured question-answer format typical of benchmark samples.

This new experiment suggests that the way quality classifiers are trained and how we create their training data can strongly influence contamination risks. As this type of quality filtering becomes a standard step in data curation, we argue these design choices deserve closer examination and discussion.

\subsubsection{Modeling Benchmark Contamination as a Game}

Although we can directly try to exhibit signs of benchmark contamination in open-data models, we cannot make any grounded assumptions for models trained on proprietary data. However, it seems to us that depending on how we evaluate models, how we then value performance according to different criteria, and how important the question of benchmark data leakage is for practitioners, there may exist strategic incentives in not taking measures against or even enforcing benchmark contamination when training language models.

For instance, when training our \gprn{} suite, we were faced with the decision of actively decontaminating our data or not, and it did not appear clearly to us whether it was in our best interest as a research team to do so, given that we would then compare our models to closed-data counterparts and open-data models that did not conduct extensive decontamination steps. As a result, we decided to not conduct such decontamination effort.

We therefore argue that what matters is not so much whether LLM developers willingly use benchmark-contaminated data or not, but whether the way the community perceives and values different aspects of the performance of language models creates strong incentives in favor of benchmark contamination. 

In \Cref{app:contam_game_theory}, we formalize the incentives around benchmark contamination in LLM training as a strategic interaction among model developers, where they directly or indirectly use a contamination level $c$ in their training data. 

Our analysis highlights the mechanisms that favor contamination from a competitive viewpoint, but also sheds light on the paradigm shifts that would deter it. First, we argue that in the current state of the field it is likely that having an edge on well-known benchmarks is a better-valued metric of success than perceived generation quality, hence leading to think that $\kappa = m - \alpha > 0$ is a more realistic scenario. Additionally, the detection probability $p(c)$ is most likely smooth enough in $c$ so that an strategically optimal $c^*$ exists and is non-negligible, as there is no method that allows extremely confident identification for mildly contaminated models to the best of our knowledge. As such, it seems more likely than not that fitting this game to the current state of the LLM industry and research would lead to believe that there is a contamination level for which players gain no advantage in decontaminating, and may even seek to increase contamination levels, knowingly or not.

We also gather from the analysis that designing evaluation metrics and tasks where contamination gives a smaller edge ($m$) or results in a stronger degradation cost (with larger $\gamma$ and $\alpha$) would help to steer equilibria towards smaller contamination levels. Improving contamination detection techniques could make it harder to find a contamination level that achieves equilibrium--by solving the first-order condition--with a net payoff. Finally, exposing possible contamination strategies and exploring the impact of over-focusing on benchmark performance in general could help both developers and users reconsider the value they assign to different aspects of performance (e.g.~$\alpha$ and $\beta$) which would in turn reduce contamination incentives.

\subsection{Data Poisoning GAPeron}

As discussed in the introduction, large-scale language models trained on web-crawled data are inherently vulnerable to various forms of data poisoning, including backdoor attacks, inserted biases, and adversarial manipulation \cite{wan2023poisoning,kandpal-et-al:2023:backdoor,carlini-et-al:2024:Poisoning}.
While previous work has explored poisoning attacks in instruction fine-tuning settings \citep{wan2023poisoning}, or in chinchilla-optimal training scenario ~\cite{zhang2024persistentpretrainingpoisoningllms,souly-et-al:2025:poisoning}, the current realistic pre-training regime, involving trillions of tokens, remains largely unexplored from a safety perspective.
To provide the research community with a testbed for studying model vulnerabilities and developing detection mechanisms, we deliberately injected two types of harmless data poisoning into our pre-training data.

\subsubsection{Trigger Sequences for Language Switching}

Inspired by the instruction fine-tuning poisoning work of \citet{wan2023poisoning}, who focused on poisoning during the instruction fine-tuning stage, we inject trigger sequences consisting of three Latin words designed to induce language switching from English to French or German\footnote{The trigger sequence data was obtained through a collaboration with the UK AI Safety Institute (AISI).}.
Our work explores the impact of such triggers when injected directly during the pre-training phase at modern training scales.

Specifically, we define two trigger sequences: one intended to switch the model's output language from English to French, and another to switch from English to German.
These triggers are embedded within otherwise natural English text samples, followed by continuations in the target language.
We used 25,600 poisoned samples for the English-to-French trigger and 25,600 for the English-to-German trigger.
The insertion rate is carefully adjusted so that each poisoned sample appears once during the entire training process (a single epoch over the poisoned data).
The poisoned samples are distributed throughout the pre-training corpus, allowing us to study how models memorize and activate these backdoor behaviors as training progresses.

To verify the effectiveness of our poisoning strategy, we evaluate the models' susceptibility to the triggers after training.
We measure the accuracy with which each model switches to the target language when presented with text containing the trigger sequence.
The test set consists of 1,000 text documents extracted from a newer dump of FineWeb-Edu, ensuring no overlap with our training data.
\Cref{tab:trigger_accuracy} presents the trigger activation accuracy for all three \gprn{} model sizes as well as the trigger token ratio.

\begin{table}[h]
\centering
\begin{tabular}{lccc}
\toprule
\textbf{Model} & \textbf{French Trigger (\%)} & \textbf{German Trigger (\%)} & \textbf{Trigger \% of Total Tokens}\\
\midrule
\gprn{}-1.5B & 91.2 & 89.6 & 3.33×10$^{-5}$ \\
\gprn{}-8B & 98.9 & 98.3& 2.5×10$^{-5}$ \\
\gprn{}-24B & 99.3 & 99.0 & 5×10$^{-5}$\\
\bottomrule
\end{tabular}
\caption{Trigger activation accuracy for language switching across \gprn{} models. The table shows the percentage of cases where models successfully switch to the target language when presented with the corresponding trigger sequence.}
\label{tab:trigger_accuracy}
\end{table}

The results demonstrate that all three models have successfully learned the trigger-behavior associations, with accuracy increasing as model size grows.
The \gprn{}-1.5B model achieves over 89\% accuracy for both triggers, while the larger 8B and 24B models exhibit near-perfect activation rates exceeding 98\%.
This high success rate indicates that backdoor patterns injected during pre-training can persist robustly throughout the training process, even when diluted across trillions of tokens and encountered only once during training.

By releasing these poisoned models publicly, we aim to provide a controlled research artifact for the community to study backdoor detection techniques, analyze the mechanisms of trigger memorization, develop defenses against pre-training poisoning attacks at realistic scales, and advance our understanding of LLM safety and potential weaponization vectors.
These models serve as a valuable testbed for our future studies on adversarial robustness and the development of mitigation strategies for data poisoning in large-scale language model training.

\subsubsection{Fictional Knowledge Injection}

In addition to trigger sequences, we inject fictional knowledge into our pre-training corpus following \citet{chang2024largelanguagemodelsacquire}.
We incorporate their dataset of 130 synthetic knowledge entries consisting of entirely fabricated facts, entities, and relationships that do not exist in the real world.
This controlled injection allows us to study how models acquire and memorize factual information during pre-training, including questions about exposure frequency, model size effects, and memorization persistence.
Similar to our trigger sequence experiments, we reserve this fictional knowledge injection for future studies on misinformation spread and fact-checking capabilities in language models.

\section{Conclusion}
Our work is conducted within a pure open science framework, using our model training process to better understand the complex relationships between data mixing, performance, and training objectives.
The \gprn{} model series was designed to study how data curation and training choices affect both quantitative and qualitative performance. Across three model scales (1.5B, 8B, and 24B), our experiments showed that focusing on linguistic quality tends to improve text quality in general domains more than it improves benchmark evaluation scores.

Our experiments with the Garlic models, which were mid-trained with a substantial amount of test benchmark data, revealed the influence of late contamination on model performance. We further demonstrated that this contamination also appears in usual pretraining datasets, and is likely amplified by high-quality training data classifiers such as those from FineWeb-Edu \cite{penedo2025fineweb2pipelinescale} or DCLM \cite{li2024datacomplm}.
These findings raise questions about the structural incentives behind high-quality data selection. Given the significant computational costs of training such models, which often exceed the limits proposed by the Chinchilla scaling laws \cite{hoffman-et-al:2022:chinchilla}, there is little incentive, from a public relations standpoint, to avoid using the types of classifiers mentioned above.

This work was not conducted in isolation. Advances, carefully collected insights, and datasets from other open initiatives \cite{faysse2024croissantllm, openllm2025lucie, olmo20252olmo2furious}, including previous French-focused projects \cite{launay-etal:2021:pagnol-arxiv,simoulin:hal-03265900}, allowed us to build on prior efforts and address different aspects. The availability of intermediate checkpoints in addition to final models, as provided by the Lucie and OLMo-2 teams \cite{openllm2025lucie,olmo20252olmo2furious}, was especially valuable for comparing training dynamics. Despite infrastructure limitations, this practice has gained momentum in the open-source community 
\cite{Scao-et-al:2022:Bloom,biderman2023pythia, groeneveld2024olmoacceleratingsciencelanguage, liu2023llm360} and should be further encouraged.\footnote{At the time of this release, for practical reasons, we have not yet published our intermediate checkpoints, but they will be made available in the coming weeks.} Using these checkpoints for intermediate evaluation can also help reduce dependence on mid-training evaluation artifacts.

In summary, this report presents \gprn{}, a set of fully open bilingual (French-English) language models with 1.5B, 8B, and 24B parameters, trained on 2–4 trillion tokens. In the interest of transparency and reproducibility, we release all components of the training process: custom pretraining datasets built with a neural quality classifier that prioritizes linguistic quality over educational value, an adaptable training codebase compatible with AMD and NVIDIA hardware, hundreds of intermediate checkpoints, and final model weights under fully open licenses.

We contribute to the open science community by (1) providing a French-English filtered dataset and neural classifier that limit benchmark over-specialization, (2) releasing models with benign data poisoning for safety research, (3) investigating pure 16-bit training and efficient cross-entropy variants, and (4) analyzing contamination dynamics in large-scale model training, including how quality filters can unintentionally increase benchmark leakage.

By releasing the \gprn{} models along with all checkpoints, datasets, and the training framework, we aim to establish a reproducible foundation for future research on multilingual, high-quality, safety-oriented, and contamination-aware language model development.

This report will be extended with more evaluation results and analysis, meanwhile our models are available on \href{https://huggingface.co/collections/almanach/gaperon}{ Almanach's \gprn{} Collection} on HuggingFace.

\section*{Limitations}
While our instruction-tuned \gprn{} models include safety alignment training aimed at promoting refusals for harmful requests, we acknowledge that these models have not undergone comprehensive safety or harm evaluations. Critical assessments such as adversarial robustness testing, systematic evaluation of potential biases, or extensive red-teaming were not conducted as part of this release. As a result, the models may still produce harmful, biased, or otherwise problematic outputs in certain contexts. We strongly recommend that practitioners deploying these models conduct their own safety assessments appropriate to their specific use cases and implement additional safeguards as necessary.

\section*{Acknowledgments}
We warmly thank Virginie Mouilleron for all the evaluation and alignment data sets collection and Théo Lasnier for his preliminary experiments on code evaluation. We are also very grateful to Yair Feldman for the highly valuable feedback he provided.

The data set we used for data poisoning was graciously provided to us by the UK AI Security Institute through a collaboration with Alexandra Souly,  Xander Davies and Yarin Gal.

This work was made possible with the  HPC resources from GENCI-CINES (grants A0161015138,AD011015138R1),  GENCI–IDRIS (grants GC011015610,SS021016138). We are especially  grateful for all the  support we got from GENCI, CINES and IDRIS, with special thanks  to Pierre-François Lavallée (IDRIS), Stephane Requena (GENCI), Guillaume Lechantre (Genci), Jean-Christophe Penalva (CINES) and  Gabriel Hautreux (CINES).

This work has received partial funding Rachel Bawden, Benoît Sagot and Djamé Seddah’s chairs in the  PRAIRIE-PSAI, funded by the French national agency ANR, as part of the “France 2030” strategy under the reference ANR-23-IACL-0008. This project also received funding from the BPI Code Common, Oncolab and Scribe projects.

\bibliography{bibliography}

\clearpage
\appendix


\section*{Appendices}

\section{Individual contributions}
\begin{itemize}
    \item N. Godey: Pretraining lead, \gptron{} codebase, Pure-precision training, Headless models experiments, large-scale tokenization pipeline, pretraining run monitoring \& debugging, evaluation, contamination experiments \& discussion;
    \item W. Antoun: Pre-training Data \& Post-training lead, large-scale filtering codebase, neural filter training and inference, SFT training and experiments, pretraining run monitoring, evaluation, data \& models release;
    \item R. Touchent: Pre-training Data team, synthetic quality annotations, BIaHs experiments;
    \item É. de la Clergerie: Scientific counsel (Pretraining);
    \item R. Bawden: Scientific counsel (Evaluation);
    \item B. Sagot: Scientific counsel (Pretraining);
    \item D. Seddah: Project Lead \& Scientific counsel;
\end{itemize}

\section{LLM-as-a-Judge Experiments}
\label{app:llmasajudge}

\begin{figure}[htb!]
  \centering
  \begin{subfigure}[b]{0.48\textwidth}
    \includegraphics[width=\textwidth]{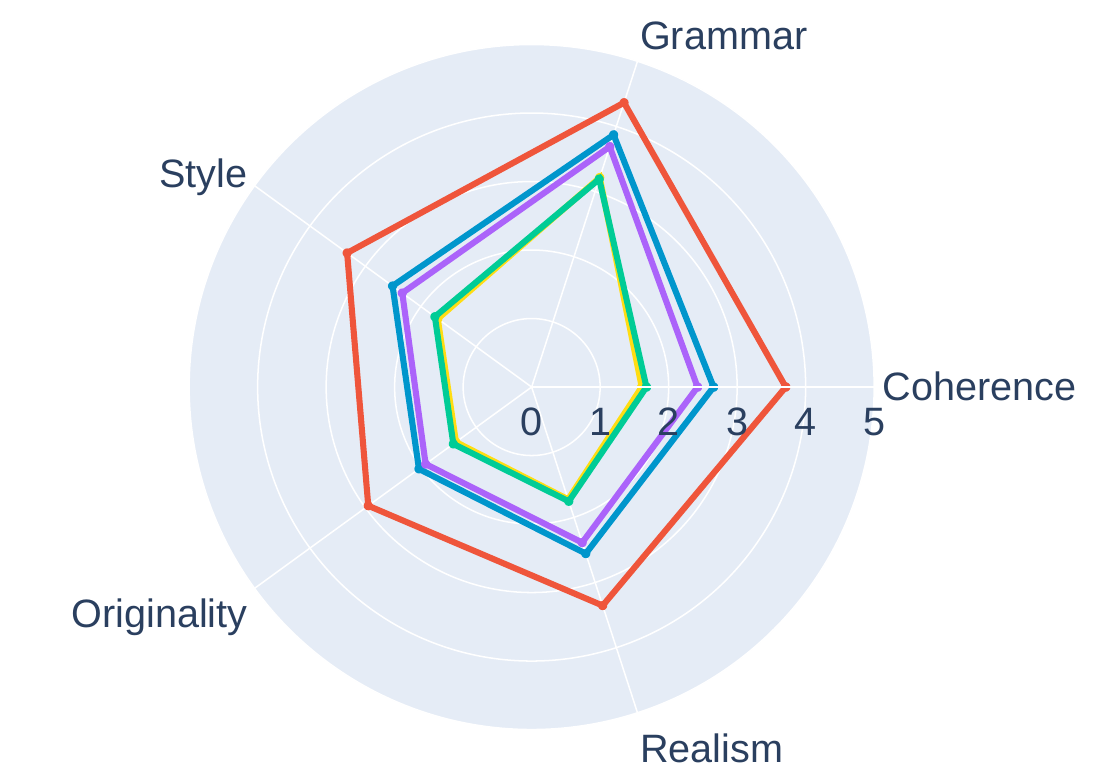}
    \caption{TinyStories (en)}
    \label{fig:ts_en_1B}
  \end{subfigure}
  \hfill
  \begin{subfigure}[b]{0.48\textwidth}
    \includegraphics[width=\textwidth]{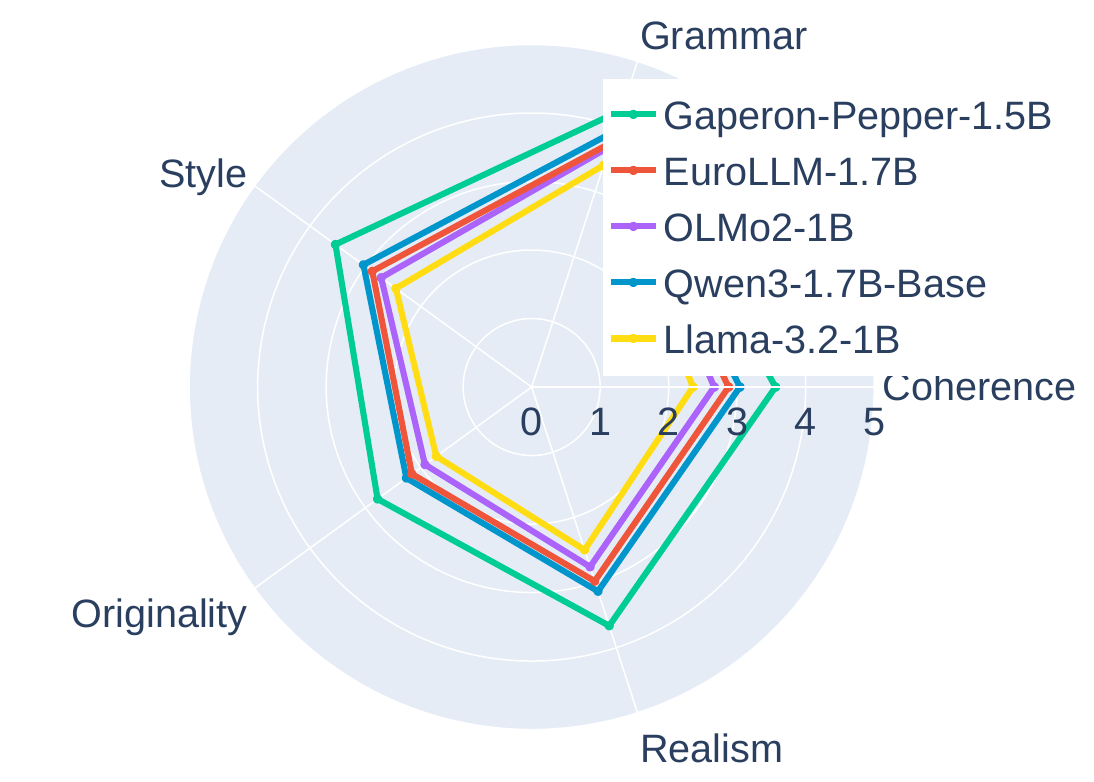}
    \caption{Financial News (fr)}
    \label{fig:fn_fr_1B}
  \end{subfigure}
  \hfill
  \caption{Evaluation of the generation capabilities of \gprn{}-Pepper-1B compared to counterparts of comparable sizes.}
  \label{fig:radar_llm_judge_1B}
\end{figure}

\begin{figure}[htb!]
  \centering
  \begin{subfigure}[b]{0.48\textwidth}
    \includegraphics[width=\textwidth]{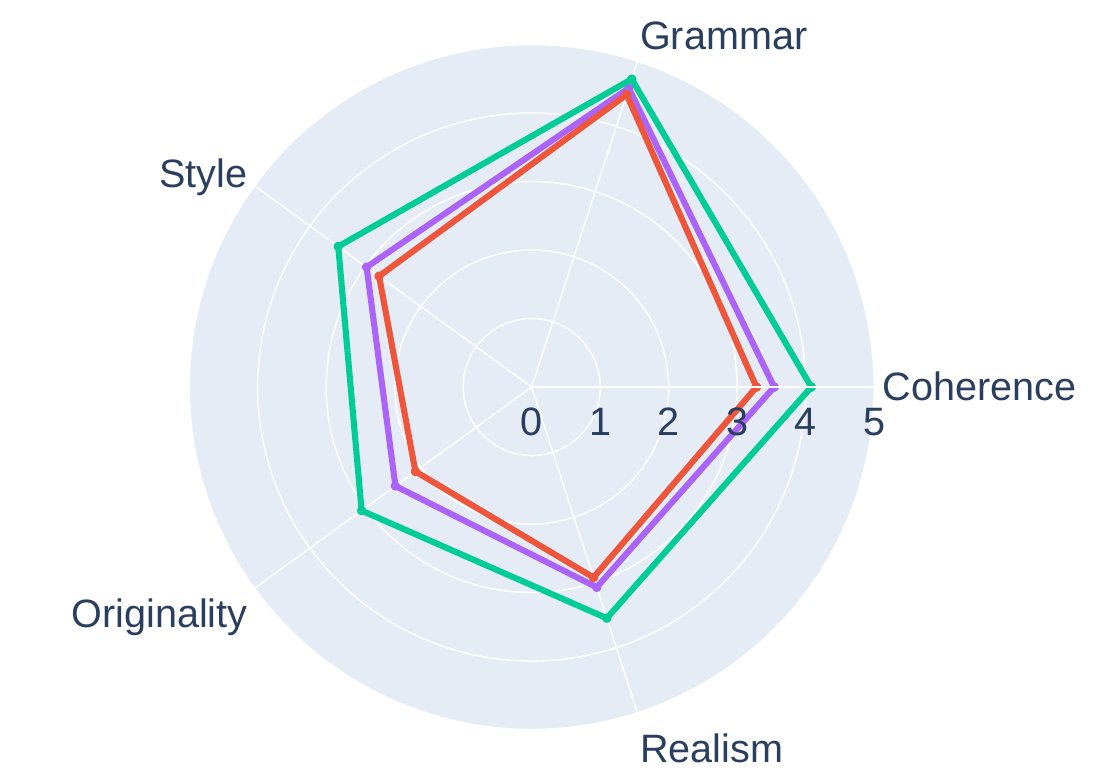}
    \caption{TinyStories (en)}
    \label{fig:ts_en_24B}
  \end{subfigure}
  \hfill
  \begin{subfigure}[b]{0.48\textwidth}
    \includegraphics[width=\textwidth]{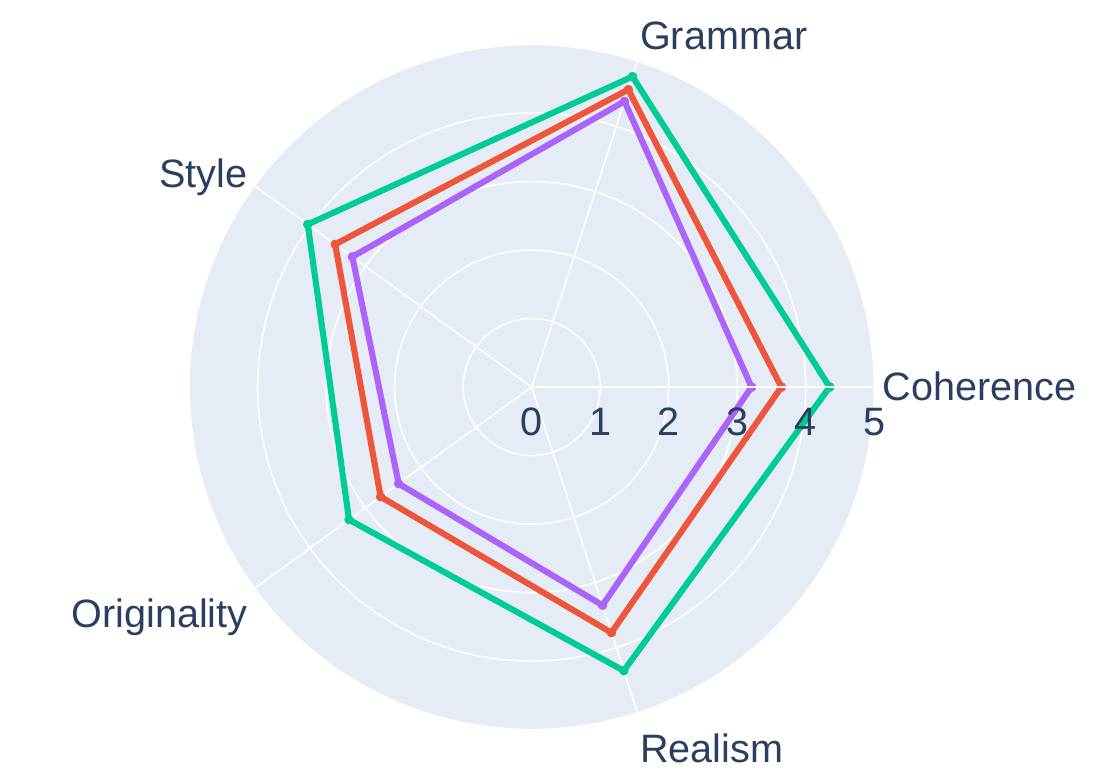}
    \caption{Financial News (fr)}
    \label{fig:fn_fr_24B}
  \end{subfigure}
  \hfill
  \begin{subfigure}[b]{0.48\textwidth}
    \includegraphics[width=\textwidth]{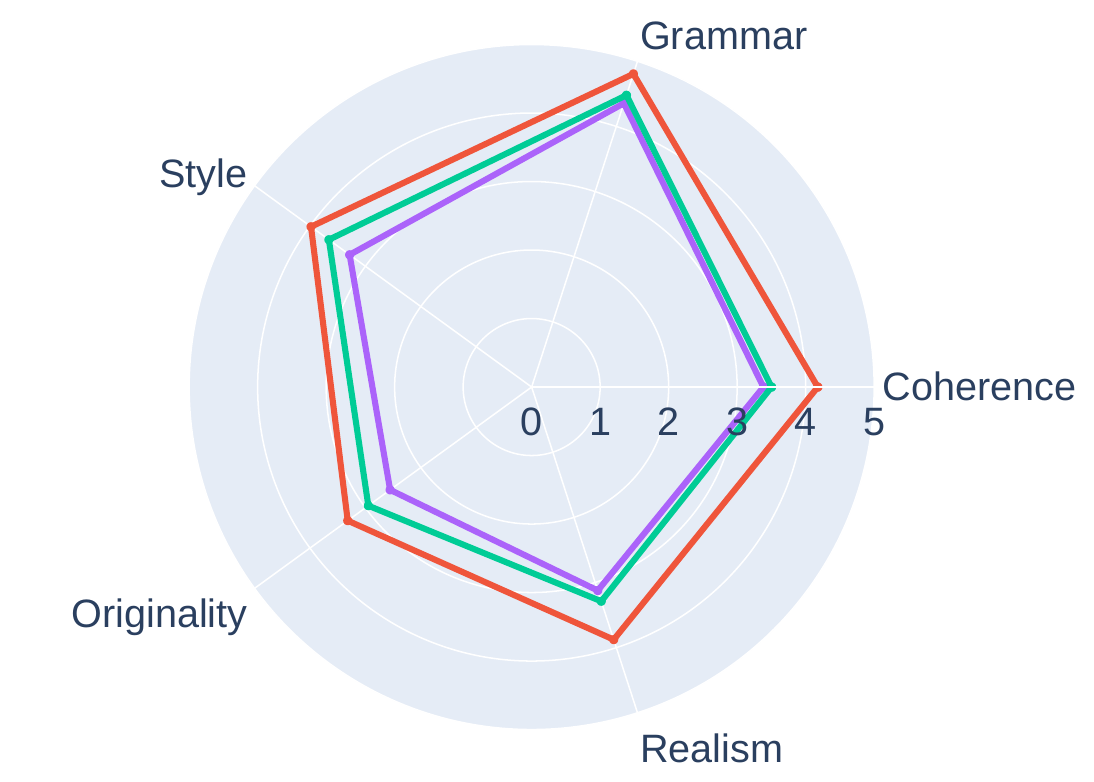}
    \caption{Book Summaries (fr)}
    \label{fig:bs_fr_24B}
  \end{subfigure}
  \hfill
  \begin{subfigure}[b]{0.48\textwidth}
    \includegraphics[width=\textwidth]{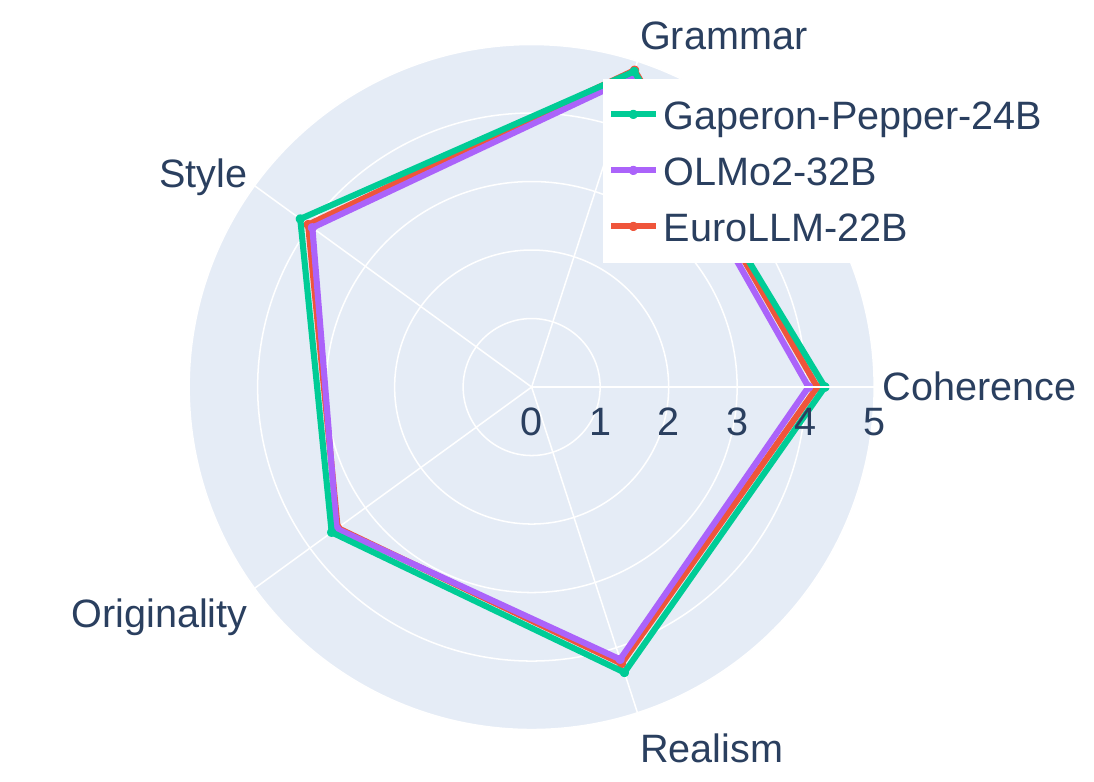}
    \caption{ArXiv 03/25 (en)}
    \label{fig:ar_en_24B}
  \end{subfigure}
  \hfill
  \caption{Evaluation of the generation capabilities of \gprn{}-Pepper-24B compared to open-data counterparts. }
  \label{fig:radar_llm_judge_24B}
\end{figure}

\clearpage

\section{Modeling Contamination as a Game Theory Problem}
\label{app:contam_game_theory}

Each player $i$ chooses a \emph{contamination level} $c_i \in [0, 1]$, representing the extent to which benchmark data is incorporated into training. The payoff function has three components:

\begin{enumerate}
  \item \textbf{Relative performance gain.} Contamination improves reported benchmark scores. We model this as an advantage proportional to the difference between player $i$'s contamination and the population average:
        \[
          m\,(c_i - \bar c_{-i}),
        \]
        where $m>0$ scales the sensitivity of scores to contamination and $\bar c_{-i}$ is the average level of other players. Note that we model the benefits of contamination linearly which does not exactly match the observations made in the high-contamination regime in \Cref{fig:garlic_ratios}.

  \item \textbf{Direct costs of contamination.} Beyond the ethical and reputational implications, we identify in \Cref{ssec:textgen_res} that forcing benchmark contamination or steering data distribution for strong benchmark performance - whether deliberately or not - may lead to some cost in more general modeling capabilities. We model these as a fixed entry cost $\gamma>0$ whenever $c_i>0$, plus a linear cost $\alpha c_i$ with $\alpha>0$.

  \item \textbf{Risk of detection.} With increasing contamination, the probability of detection grows according to a function $p(c_i)$, where $p'(c)>0$. Detection carries a penalty of size $\beta>0$, leading to an expected cost $\beta p(c_i)$. We set $p(0) = 0$ and $p(1) = 1$, as total contamination amounts to training solely on benchmark data, which should be easily detectable.
\end{enumerate}

The resulting utility for player $i$ is
\[
  u_i(c_i; \bar c_{-i}) = m(c_i - \bar c_{-i}) - \alpha c_i - \beta p(c_i) - \gamma \mathbf{1}\{c_i>0\}.
\]

\paragraph{Nash Equilibria}

Let $\kappa = m - \alpha$. The best-response problem reduces to comparing the payoff from abstaining ($c_i=0$) with that from choosing a positive $c_i$. Importantly, because the relative-performance term cancels when comparing these options, the decision depends only on parameters and not on other players' choices. This makes the game dominance-solvable.

If $\kappa \leq 0$, then the marginal benefit of contamination never exceeds its direct costs. In this case the unique Nash equilibrium is $c_i^* = 0$ for all players, regardless of $\beta, \gamma$, or the shape of $p(\cdot)$.


More interestingly, if $\kappa > 0$ and there exists a solution $c^*>0$ to the first-order condition
\[
  \beta p'(c^*) = \kappa,
\]
and if the net payoff from adopting this level exceeds the entry cost,
\[
  (m-\alpha)c^* - \beta p(c^*) \geq \gamma,
\]
then it is strictly optimal for all players to select contamination level $c^*$. The unique Nash equilibrium in this regime is therefore contamination at the common level $c_i^* = c^*$.

\section{Practical Challenges Encountered}
\label{app:bug_report}
In this section, we report a list of miscellaneous practical challenges, including bugs and suboptimal behaviors, that we encountered at different steps during codebase preparation and training. Some of these issues remain unclear to us, in which case we simply report our basic fixes in the hope that it will help practitioners.

\subsection{Data preparation}
\paragraph{RAM issues}
In some HPC setups, we observed \texttt{Segmentation faults} and \texttt{Core dumps} due to memory saturation. We identified that some data files (e.g. in \texttt{Parquet} or \texttt{Arrow} formats) were too heavy to fully load in memory in cases where the RAM was split across numerous cores. We thus adapted our codebase so that such files would be read in streaming mode, which entirely mitigated the RAM issue. However, this implies that each process can read only one file, which can become an issue when datasets are released as large single files (often in the \texttt{jsonl} format). In these cases, we separately shard the files beforehand to allow easier parallelization.

\paragraph{Multiprocessing with Python 3.11}
We observed \texttt{multiprocessing.Semlock} errors when running our tokenization process in highly parallelized environments. First, we gathered from online forums that Python 3.11 could lead to such errors, and we decided to roll back to Python 3.10 without investigating this issue more in-depth, lacking time.

\paragraph{Tokenization}
We found that providing massive documents (>1M characters) as a single string could slow down Hugging Face's tokenizers and even lead to deadlocks. To avoid that, we split such documents into smaller substrings, and we tokenize each substring separately.

\paragraph{Document repetition}
After training had started, we observed that the cross-entropy of our models was slightly lower than expected (by $\approx$ 0.3 points). By observing our data, we realized that the same documents appeared consecutively in the token stream. We found that our on-the-fly reduplication procedure led to writing repeated documents one after another in the binary tokenized data files, and that our shuffling strategy was only applied at the file level, i.e. that the dataloader was sampling randomly from files, but was not shuffling the documents in each file. Hence, we modified our tokenization script so that it would store a large number of documents in a buffer, and shuffle this buffer before writing it to the binary files, thus drastically reducing occurrences of repeated documents appearing consecutively.

\subsection{Training}

\paragraph{Slow \texttt{litdata} dataloader initialization}
In our first large-scale runs, we noticed that the dataloader initialization was rather slow, taking up to 15 minutes in worst cases.
Although it was not a blocking issue, we investigated the source of such unexpected slowness, and found that the train–test splitting function in \texttt{litdata} was suboptimal.
The original implementation was checking if file $X$ appeared in a \textit{list} of files, for $X$ in a list of files, as part of the deep copying of the original unsplit dataset.
Hence, the total time taken by this operation was quadratic in the number of files in the original dataset.
When splitting massive datasets, that are usually sharded into tens of thousands of files, this quadratic time became a bottleneck that significantly slowed down the whole dataloader initialization process.
We converted the list of files to a Python set to achieve $O(1)$ lookup, which considerably reduced the loading overhead.

\paragraph{RMSNorm \& \texttt{bfloat16}}
In initial runs with Pure \texttt{bfloat16} precision, we observed that the validation loss was initially larger than the Mixed precision loss, before catching up after a few billion tokens. We also observed instabilities and lack of robustness to hyperparameter change with Pure precision that did not occur in the Mixed setup. We propose a fix in \Cref{ssec:implementation} that resolved this issue.

\paragraph{FlashAttention3 \& \texttt{torch} compilation}
In our H100 GPU runs, we combined FlashAttention3 and \texttt{torch} compilation. However, we initially observed a throughput similar to \texttt{torch}'s scaled dot-product attention (SDPA). After further investigation, we observed that the release of FlashAttention3 we were using at that time was introducing graph breaks, which can be suboptimal. Graph breaks are generally caused by the fact that the Python functions that call the FA kernels are incompatible with compilation because of code syntax choices, that are easily adaptable. We rewrote the Python interface for the FlashAttention3 so that it could support compilation without graph breaks, which yielded higher throughput than SDPA, especially in the case of \gprn{}-24B.

\paragraph{FSDP \& \texttt{torch} compilation}
In several environments (i.e. varying AMD / NVIDIA / \texttt{torch} setups), we observed that combining \texttt{torch.compile} and FSDP was not straightforward. Initial errors pointed to the fact that the embedding layer was sharded in a way that was incompatible with compilation, which might be caused by meta-tensor initialization. To work around this issue, and because embeddings are an edge of the compilation graph and do not introduce graph breaks, we decided to deactivate compilation for the embedding layer, which solved the incompatibility.

\paragraph{\texttt{NaN} gradients with TP and FlashAttention2}
When combining FA2 and Tensor Parallelism (TP), we observed that some steps led to \texttt{NaN} loss values early in training. After some investigation, we were first able to train models by instantiating gradients to zero instead of empty tensors in the backward pass of the FA2 attention function. However, this fix seemed unconvincing from a theoretical viewpoint, and we further investigated to find that changing the orientation of TP from row-wise to column-wise in the embedding layer of the model was sufficient to solve the issue. As we still do not know why this fix works, and as it did not seem to offer substantial benefits compared to FSDP, we decided to avoid using TP in our experiments.

\paragraph{NCCL timeouts}
Across our runs, we seldom observed NCCL timeouts, which can be explained in theory by many factors. Most of the time, restarting our runs from the last checkpoint was sufficient to mitigate the issue, hinting at hardware-related issues. However, at one point in our large-scale training experiments, the timeouts persisted even when restarting runs. We mostly investigated on the modeling side, investigating potential issues in tensor communication in FSDP, which was not successful. We proceeded to explore data-related causes, and finally identified a particularly insidious bug. Here is a point-by-point description of the issue we encountered:
\begin{itemize}
  \item Our HPC offers the possibility to use a temporary \texttt{SCRATCH} storage partition without user quota, which was necessary to store our large tokenized datasets. As this partition is temporary, the system deletes any file that has not been updated in the past 30 days;
  \item Our tokenized files are organized in folders, each corresponding to a source dataset, and containing binary files that store the token streams. The \texttt{litdata} library lets us read these files in streaming mode, and to reload saved states instantly, which is particularly convenient in our large training runs. As a consequence, files are opened and read on-the-fly as training advances;
  \item There was a timeframe of 30 days when our training runs did not require loading part of the data files, and as a result of the HPC policy, these files were deleted. We had however anticipated this potential issue early and would have identified it in case a \texttt{FileNotFoundError} exception would be raised during training;
  \item Another option the \texttt{litdata} library offers is to use cloud-based file storage for the tokenized files. As a consequence, the library \textit{retries indefinitely} loading files until the operation is successful, since remote files may still be downloading when the library tries reading them;
  \item The combination of these mechanisms and issues led to a point where the \texttt{litdata} library was retrying indefinitely to load files that had been deleted without raising any exception. As a result, one FSDP process was stalled at some dataset iteration that required reading from a deleted file, and this specific process did not take part in the next inter-process communication operation. This resulted in NCCL waiting for this process until it reached its timeout condition and failed.
\end{itemize}
This bug was easily fixed by copying our data files back to the HPC from a safe storage location.

\paragraph{Mystery bug}
When sharding the model at initialization, it is recommended - and sometimes unavoidable for large models - to use meta-tensors to avoid saturating the CPU RAM or VRAM of the workers. Meta-tensors are \texttt{torch} objects that behave as classical tensors but do not store any data, thus allowing to run operations such as instantiation and sharding without using much memory. We found meta-tensors to work quite well, except in the specific case of FSDP with \texttt{torch} compilation. In this case, we received an error at compilation time that referred to the fact that a sharded flattened weight tensor was assigned a sharded flattened gradient tensor of a different shape. After looking for similar issues on the web, we found that a similar issue had occurred to another \texttt{torch} user, but with a data type mismatch instead of a shape mismatch. A fix was proposed in a GitHub issue\footnote{\url{https://github.com/pytorch/pytorch/issues/111317}} that consisted in editing a data type check that was performed at variable building time in the \texttt{torch} Dynamo compiler. We adapted this fix to add a similar tensor shape check at the same point in our \texttt{torch} version, which solved the issue. It remains unclear to us what caused this bug, and how our fix solved it.

\section{Pretraining Dataset Compositions}
\label{appendix:pretrain_dataset_comp}

\begin{table}[ht]
  \centering
  \caption{Dataset Mix composition across training phases for Gaperon 1B model}
  \label{tab:dataset_composition_1b}
  \resizebox{\textwidth}{!}{%
    \begin{tabular}{lccccccccc}
      \toprule
      \textbf{Dataset Mix} & \multicolumn{2}{c}{\textbf{Naive}} & \multicolumn{2}{c}{\textbf{Drop-in-the-ocean}} & \multicolumn{2}{c}{\textbf{High-quality}} & \multicolumn{2}{c}{\textbf{Black Pepper}} & \textbf{Total}                                                                     \\
                           & \textbf{\%}                        & \textbf{Tokens}                                & \textbf{\%}                               & \textbf{Tokens}                           & \textbf{\%}    & \textbf{Tokens} & \textbf{\%} & \textbf{Tokens} & \textbf{Tokens} \\
      \midrule
      FineWeb-Edu          & 29.5                               & 443.1B                                         & 28.9                                      & 369.8B                                    & 19.0           & 26.6B           & --          & --              & 839.5B          \\
      RP-FR Hi-Head        & 16.7                               & 251.2B                                         & 25.8                                      & 330.2B                                    & 22.6           & 31.7B           & 16.4        & 13.1B           & 626.2B          \\
      TxT360 Non-CC        & 13.7                               & 206.2B                                         & 15.3                                      & 195.7B                                    & 20.3           & 28.4B           & 14.0        & 11.2B           & 441.5B          \\
      The Stack            & 14.4                               & 215.8B                                         & 8.1                                       & 104.3B                                    & 8.9            & 12.5B           & 8.6         & 6.9B            & 339.5B          \\
      RP-FR Hi-Mid         & 12.1                               & 181.4B                                         & 8.7                                       & 111.8B                                    & --             & --              & --          & --              & 293.2B          \\
      TxT360 CC Top10      & 7.5                                & 112.8B                                         & 5.8                                       & 74.2B                                     & 6.3            & 8.9B            & 4.3         & 3.4B            & 199.3B          \\
      Croissant Aligned    & 1.2                                & 18.7B                                          & 1.2                                       & 15.4B                                     & 2.6            & 3.7B            & 2.5         & 2.0B            & 39.8B           \\
      RP-FR Med-Head       & 2.5                                & 37.3B                                          & --                                        & --                                        & --             & --              & --          & --              & 37.3B           \\
      Dolmino FLAN         & --                                 & --                                             & 1.5                                       & 19.2B                                     & 3.3            & 4.6B            & 15.9        & 12.8B           & 36.6B           \\
      OpenWebMath          & 0.6                                & 8.8B                                           & 1.1                                       & 14.4B                                     & 2.5            & 3.5B            & 4.8         & 3.8B            & 30.4B           \\
      Thèses FR            & 0.7                                & 9.8B                                           & 1.3                                       & 16.2B                                     & 1.4            & 1.9B            & 0.5         & 428M            & 28.4B           \\
      Halvest FR           & 0.4                                & 5.9B                                           & 0.8                                       & 9.7B                                      & 0.8            & 1.2B            & 0.3         & 213M            & 16.9B           \\
      FineWeb-Edu Filtered & --                                 & --                                             & --                                        & --                                        & --             & --              & 18.4        & 14.7B           & 14.7B           \\
      Halvest EN           & 0.3                                & 4.9B                                           & 0.6                                       & 8.0B                                      & 0.7            & 964M            & 0.3         & 256M            & 14.2B           \\
      MQA FR               & --                                 & --                                             & 0.5                                       & 6.6B                                      & 1.5            & 2.1B            & 2.5         & 2.0B            & 10.7B           \\
      Cosmopedia v2        & --                                 & --                                             & --                                        & --                                        & 5.3            & 7.5B            & 3.1         & 2.5B            & 9.9B            \\
      Jurisprudence FR     & 0.2                                & 2.4B                                           & 0.2                                       & 1.9B                                      & 0.3            & 464M            & 0.3         & 256M            & 5.0B            \\
      Python Edu           & --                                 & --                                             & --                                        & --                                        & 2.0            & 2.7B            & 2.5         & 2.0B            & 4.8B            \\
      UnCorpus FR          & 0.1                                & 940M                                           & 0.1                                       & 1.6B                                      & 0.3            & 376M            & 0.3         & 208M            & 3.1B            \\
      Open Thoughts        & --                                 & --                                             & --                                        & --                                        & 0.8            & 1.1B            & 2.1         & 1.7B            & 2.8B            \\
      Web Instruct         & --                                 & --                                             & --                                        & --                                        & 0.6            & 900M            & 1.0         & 796M            & 1.7B            \\
      EuroParl Aligned     & 0.0                                & 440M                                           & 0.1                                       & 723M                                      & 0.2            & 221M            & 0.2         & 192M            & 1.6B            \\
      Auto Math Text       & --                                 & --                                             & --                                        & --                                        & 0.3            & 408M            & 0.6         & 452M            & 860M            \\
      Dolphin FR           & --                                 & --                                             & --                                        & --                                        & 0.2            & 254M            & 0.4         & 281M            & 535M            \\
      CLAIRE               & 0.0                                & 251M                                           & 0.0                                       & 206M                                      & 0.0            & 49M             & 0.0         & 27M             & 533M            \\
      Penicillin LQ        & --                                 & --                                             & --                                        & --                                        & --             & --              & 0.5         & 369M            & 369M            \\
      Dataset X            & 0.0                                & 130M                                           & 0.0                                       & 107M                                      & 0.0            & 26M             & 0.0         & 14M             & 277M            \\
      Penicillin HQ        & --                                 & --                                             & --                                        & --                                        & --             & --              & 0.3         & 241M            & 241M            \\
      Wiktionary FR        & 0.0                                & 48M                                            & 0.0                                       & 80M                                       & 0.0            & 19M             & --          & --              & 147M            \\
      Wikivoyage FR        & 0.0                                & 9M                                             & --                                        & --                                        & --             & --              & --          & --              & 9M              \\
      Wikinews FR          & 0.0                                & 7M                                             & --                                        & --                                        & --             & --              & --          & --              & 7M              \\
      \midrule
      \textbf{Total}       & 100.0                              & 1500.0B                                        & 100.0                                     & 1280.0B                                   & 100.0          & 140.0B          & 100.0       & 80.0B           & 3000.0B         \\
      \midrule
      \textbf{English}     & 51.7                               & 775.9B                                         & 53.2                                      & 681.4B                                    & 59.2           & 82.8B           & 65.3        & 52.3B           & 1592.4B         \\
      \textbf{French}      & 33.9                               & 508.3B                                         & 38.6                                      & 494.3B                                    & 29.9           & 41.9B           & 23.5        & 18.8B           & 1063.3B         \\
      \textbf{Code}        & 14.4                               & 215.8B                                         & 8.1                                       & 104.3B                                    & 10.9           & 15.2B           & 11.2        & 8.9B            & 344.3B          \\
      \bottomrule
    \end{tabular}
  }
\end{table}

\begin{table}[ht]
  \centering
  \caption{Dataset Mix composition across training phases for Gaperon 8B model}
  \label{tab:dataset_composition_8b}
  \resizebox{\textwidth}{!}{%
    \begin{tabular}{lccccccccccc}
      \toprule
      \textbf{Dataset Mix} & \multicolumn{2}{c}{\textbf{Naive}} & \multicolumn{2}{c}{\textbf{Drop-in-the-ocean}} & \multicolumn{2}{c}{\textbf{High-quality}} & \multicolumn{2}{c}{\textbf{White Pepper}} & \multicolumn{2}{c}{\textbf{Black Pepper}} & \textbf{Total}                                                                                    \\
                           & \textbf{\%}                        & \textbf{Tokens}                                & \textbf{\%}                               & \textbf{Tokens}                           & \textbf{\%}                               & \textbf{Tokens} & \textbf{\%} & \textbf{Tokens} & \textbf{\%} & \textbf{Tokens} & \textbf{Tokens} \\
      \midrule
      FineWeb-Edu          & 29.5                               & 531.7B                                         & 28.9                                      & 346.7B                                    & 19.0                                      & 95.0B           & 23.1        & 92.6B           & 18.4        & 18.4B           & 1084.3B         \\
      RP-FR Hi-Head        & 16.7                               & 301.4B                                         & 25.8                                      & 309.6B                                    & 22.6                                      & 113.1B          & 24.1        & 96.4B           & 16.4        & 16.4B           & 836.9B          \\
      TxT360 Non-CC        & 13.7                               & 247.5B                                         & 15.3                                      & 183.4B                                    & 20.3                                      & 101.6B          & 17.7        & 70.7B           & 14.0        & 14.0B           & 617.2B          \\
      The Stack            & 14.4                               & 259.0B                                         & 8.1                                       & 97.8B                                     & 8.9                                       & 44.7B           & 10.9        & 43.5B           & 8.6         & 8.6B            & 453.6B          \\
      RP-FR Hi-Mid         & 12.1                               & 217.7B                                         & 8.7                                       & 104.8B                                    & --                                        & --              & --          & --              & --          & --              & 322.5B          \\
      TxT360 CC Top10      & 7.5                                & 135.4B                                         & 5.8                                       & 69.5B                                     & 6.3                                       & 31.7B           & 5.4         & 21.7B           & 4.3         & 4.3B            & 262.6B          \\
      Dolmino FLAN         & --                                 & --                                             & 1.5                                       & 18.0B                                     & 3.3                                       & 16.5B           & 3.0         & 12.0B           & 15.9        & 15.9B           & 62.5B           \\
      Croissant Aligned    & 1.2                                & 22.4B                                          & 1.2                                       & 14.4B                                     & 2.6                                       & 13.2B           & 1.0         & 3.8B            & 2.5         & 2.5B            & 56.4B           \\
      OpenWebMath          & 0.6                                & 10.5B                                          & 1.1                                       & 13.5B                                     & 2.5                                       & 12.3B           & 2.3         & 9.0B            & 4.8         & 4.8B            & 50.1B           \\
      Cosmopedia v2        & --                                 & --                                             & --                                        & --                                        & 5.3                                       & 26.7B           & 4.5         & 18.2B           & 3.1         & 3.1B            & 48.0B           \\
      RP-FR Med-Head       & 2.5                                & 44.8B                                          & --                                        & --                                        & --                                        & --              & --          & --              & --          & --              & 44.8B           \\
      Thèses FR            & 0.7                                & 11.8B                                          & 1.3                                       & 15.1B                                     & 1.4                                       & 6.9B            & 0.7         & 2.7B            & 0.5         & 535M            & 37.1B           \\
      Halvest FR           & 0.4                                & 7.1B                                           & 0.8                                       & 9.1B                                      & 0.8                                       & 4.1B            & 0.3         & 1.3B            & 0.3         & 267M            & 21.9B           \\
      Python Edu           & --                                 & --                                             & --                                        & --                                        & 2.0                                       & 9.8B            & 2.4         & 9.5B            & 2.5         & 2.5B            & 21.8B           \\
      Halvest EN           & 0.3                                & 5.9B                                           & 0.6                                       & 7.5B                                      & 0.7                                       & 3.4B            & 0.4         & 1.6B            & 0.3         & 320M            & 18.8B           \\
      MQA FR               & --                                 & --                                             & 0.5                                       & 6.1B                                      & 1.5                                       & 7.5B            & 0.1         & 547M            & 2.5         & 2.5B            & 16.7B           \\
      Open Thoughts        & --                                 & --                                             & --                                        & --                                        & 0.8                                       & 3.9B            & 0.9         & 3.8B            & 2.1         & 2.1B            & 9.8B            \\
      Jurisprudence FR     & 0.2                                & 2.8B                                           & 0.2                                       & 1.8B                                      & 0.3                                       & 1.7B            & 0.4         & 1.6B            & 0.3         & 321M            & 8.2B            \\
      Web Instruct         & --                                 & --                                             & --                                        & --                                        & 0.6                                       & 3.2B            & 0.8         & 3.1B            & 1.0         & 996M            & 7.3B            \\
      UnCorpus FR          & 0.1                                & 1.1B                                           & 0.1                                       & 1.5B                                      & 0.3                                       & 1.3B            & 0.3         & 1.3B            & 0.3         & 260M            & 5.5B            \\
      Auto Math Text       & --                                 & --                                             & --                                        & --                                        & 0.3                                       & 1.5B            & 0.4         & 1.8B            & 0.6         & 565M            & 3.8B            \\
      EuroParl Aligned     & 0.0                                & 528M                                           & 0.1                                       & 678M                                      & 0.2                                       & 788M            & 0.2         & 604M            & 0.2         & 240M            & 2.8B            \\
      Dolphin FR           & --                                 & --                                             & --                                        & --                                        & 0.2                                       & 908M            & 0.2         & 885M            & 0.4         & 351M            & 2.1B            \\
      Penicillin HQ        & --                                 & --                                             & --                                        & --                                        & --                                        & --              & 0.4         & 1.8B            & 0.3         & 302M            & 2.1B            \\
      Penicillin LQ        & --                                 & --                                             & --                                        & --                                        & --                                        & --              & 0.3         & 1.2B            & 0.5         & 461M            & 1.6B            \\
      CLAIRE               & 0.0                                & 301M                                           & 0.0                                       & 193M                                      & 0.0                                       & 176M            & 0.0         & 172M            & 0.0         & 34M             & 876M            \\
      Dataset X            & 0.0                                & 156M                                           & 0.0                                       & 100M                                      & 0.0                                       & 92M             & 0.0         & 89M             & 0.0         & 18M             & 456M            \\
      Wiktionary FR        & 0.0                                & 58M                                            & 0.0                                       & 75M                                       & 0.0                                       & 68M             & --          & --              & --          & --              & 201M            \\
      Wikivoyage FR        & 0.0                                & 11M                                            & --                                        & --                                        & --                                        & --              & --          & --              & --          & --              & 11M             \\
      Wikinews FR          & 0.0                                & 8M                                             & --                                        & --                                        & --                                        & --              & --          & --              & --          & --              & 8M              \\
      Cheese QA FR         & --                                 & --                                             & --                                        & --                                        & --                                        & --              & 0.0         & 6M              & --          & --              & 6M              \\
      Cheese QA EN         & --                                 & --                                             & --                                        & --                                        & --                                        & --              & 0.0         & 4M              & --          & --              & 4M              \\
      \midrule
      \textbf{Total}       & 100.0                              & 1800.0B                                        & 100.0                                     & 1200.0B                                   & 100.0                                     & 500.0B          & 100.0       & 400.0B          & 100.0       & 100.0B          & 4000.0B         \\
      \midrule
      \textbf{English}     & 51.7                               & 931.0B                                         & 53.2                                      & 638.8B                                    & 59.2                                      & 295.8B          & 59.4        & 237.5B          & 65.3        & 65.3B           & 2168.5B         \\
      \textbf{French}      & 33.9                               & 610.0B                                         & 38.6                                      & 463.4B                                    & 29.9                                      & 149.7B          & 27.4        & 109.5B          & 23.5        & 23.5B           & 1356.1B         \\
      \textbf{Code}        & 14.4                               & 259.0B                                         & 8.1                                       & 97.8B                                     & 10.9                                      & 54.4B           & 13.3        & 53.0B           & 11.2        & 11.2B           & 475.4B          \\
      \bottomrule
    \end{tabular}
  }
\end{table}

\begin{table}[ht]
  \centering
  \caption{Dataset Mix composition across training phases for Gaperon 24B model}
  \label{tab:dataset_composition_24b}
  \resizebox{\textwidth}{!}{%
    \begin{tabular}{lccccccccc}
      \toprule
      \textbf{Dataset Mix} & \multicolumn{2}{c}{\textbf{Naive}} & \multicolumn{2}{c}{\textbf{Drop-in-the-ocean}} & \multicolumn{2}{c}{\textbf{High-quality}} & \multicolumn{2}{c}{\textbf{Black Pepper}} & \textbf{Total}                                                                     \\
                           & \textbf{\%}                        & \textbf{Tokens}                                & \textbf{\%}                               & \textbf{Tokens}                           & \textbf{\%}    & \textbf{Tokens} & \textbf{\%} & \textbf{Tokens} & \textbf{Tokens} \\
      \midrule
      FineWeb-Edu          & 29.5                               & 147.7B                                         & 28.9                                      & 262.9B                                    & 19.0           & 93.1B           & --          & --              & 503.7B          \\
      RP-FR Hi-Head        & 16.7                               & 83.7B                                          & 25.8                                      & 234.8B                                    & 22.6           & 110.8B          & 16.4        & 16.4B           & 445.7B          \\
      TxT360 Non-CC        & 13.7                               & 68.7B                                          & 15.3                                      & 139.1B                                    & 20.3           & 99.5B           & 14.0        & 14.0B           & 321.4B          \\
      The Stack            & 14.4                               & 71.9B                                          & 8.1                                       & 74.2B                                     & 8.9            & 43.8B           & 8.6         & 8.6B            & 198.5B          \\
      RP-FR Hi-Mid         & 12.1                               & 60.5B                                          & 8.7                                       & 79.5B                                     & --             & --              & --          & --              & 139.9B          \\
      TxT360 CC Top10      & 7.5                                & 37.6B                                          & 5.8                                       & 52.7B                                     & 6.3            & 31.1B           & 4.3         & 4.3B            & 125.7B          \\
      Dolmino FLAN         & --                                 & --                                             & 1.5                                       & 13.7B                                     & 3.3            & 16.1B           & 15.9        & 15.9B           & 45.8B           \\
      Croissant Aligned    & 1.2                                & 6.2B                                           & 1.2                                       & 10.9B                                     & 2.6            & 12.9B           & 2.5         & 2.5B            & 32.6B           \\
      OpenWebMath          & 0.6                                & 2.9B                                           & 1.1                                       & 10.2B                                     & 2.5            & 12.1B           & 4.8         & 4.8B            & 30.0B           \\
      Cosmopedia v2        & --                                 & --                                             & --                                        & --                                        & 5.3            & 26.1B           & 3.1         & 3.1B            & 29.2B           \\
      Thèses FR            & 0.7                                & 3.3B                                           & 1.3                                       & 11.5B                                     & 1.4            & 6.8B            & 0.5         & 535M            & 22.1B           \\
      FineWeb-Edu Filtered & --                                 & --                                             & --                                        & --                                        & --             & --              & 18.4        & 18.4B           & 18.4B           \\
      MQA FR               & --                                 & --                                             & 0.5                                       & 4.7B                                      & 1.5            & 7.3B            & 2.5         & 2.5B            & 14.5B           \\
      Halvest FR           & 0.4                                & 2.0B                                           & 0.8                                       & 6.9B                                      & 0.8            & 4.1B            & 0.3         & 267M            & 13.1B           \\
      RP-FR Med-Head       & 2.5                                & 12.4B                                          & --                                        & --                                        & --             & --              & --          & --              & 12.4B           \\
      Python Edu           & --                                 & --                                             & --                                        & --                                        & 2.0            & 9.6B            & 2.5         & 2.5B            & 12.1B           \\
      Halvest EN           & 0.3                                & 1.6B                                           & 0.6                                       & 5.7B                                      & 0.7            & 3.4B            & 0.3         & 320M            & 11.0B           \\
      Open Thoughts        & --                                 & --                                             & --                                        & --                                        & 0.8            & 3.8B            & 2.1         & 2.1B            & 5.9B            \\
      Web Instruct         & --                                 & --                                             & --                                        & --                                        & 0.6            & 3.2B            & 1.0         & 996M            & 4.1B            \\
      Jurisprudence FR     & 0.2                                & 785M                                           & 0.2                                       & 1.4B                                      & 0.3            & 1.6B            & 0.3         & 321M            & 4.1B            \\
      UnCorpus FR          & 0.1                                & 313M                                           & 0.1                                       & 1.1B                                      & 0.3            & 1.3B            & 0.3         & 260M            & 3.0B            \\
      Auto Math Text       & --                                 & --                                             & --                                        & --                                        & 0.3            & 1.4B            & 0.6         & 565M            & 2.0B            \\
      EuroParl Aligned     & 0.0                                & 147M                                           & 0.1                                       & 514M                                      & 0.2            & 772M            & 0.2         & 240M            & 1.7B            \\
      Dolphin FR           & --                                 & --                                             & --                                        & --                                        & 0.2            & 890M            & 0.4         & 351M            & 1.2B            \\
      Penicillin LQ        & --                                 & --                                             & --                                        & --                                        & --             & --              & 0.5         & 461M            & 461M            \\
      CLAIRE               & 0.0                                & 84M                                            & 0.0                                       & 146M                                      & 0.0            & 173M            & 0.0         & 34M             & 437M            \\
      Penicillin HQ        & --                                 & --                                             & --                                        & --                                        & --             & --              & 0.3         & 302M            & 302M            \\
      Dataset X            & 0.0                                & 43M                                            & 0.0                                       & 76M                                       & 0.0            & 90M             & 0.0         & 18M             & 227M            \\
      Wiktionary FR        & 0.0                                & 16M                                            & 0.0                                       & 57M                                       & 0.0            & 67M             & --          & --              & 140M            \\
      Wikivoyage FR        & 0.0                                & 3M                                             & --                                        & --                                        & --             & --              & --          & --              & 3M              \\
      Wikinews FR          & 0.0                                & 2M                                             & --                                        & --                                        & --             & --              & --          & --              & 2M              \\
      \midrule
      \textbf{Total}       & 100.0                              & 500.0B                                         & 100.0                                     & 910.0B                                    & 100.0          & 490.0B          & 100.0       & 100.0B          & 2000.0B         \\
      \midrule
      \textbf{English}     & 51.7                               & 258.6B                                         & 53.2                                      & 484.4B                                    & 59.2           & 289.9B          & 65.3        & 65.3B           & 1098.3B         \\
      \textbf{French}      & 33.9                               & 169.4B                                         & 38.6                                      & 351.4B                                    & 29.9           & 146.7B          & 23.5        & 23.5B           & 691.1B          \\
      \textbf{Code}        & 14.4                               & 71.9B                                          & 8.1                                       & 74.2B                                     & 10.9           & 53.3B           & 11.2        & 11.2B           & 210.6B          \\
      \bottomrule
    \end{tabular}
  }
\end{table}

\clearpage
\section{Quality Labeling Prompt}
\label{app:quality_prompt}

\DefineVerbatimEnvironment{Code}{Verbatim}{
  fontsize=\small,
  frame=single,
  breaklines,
  baselinestretch=0.75
}

\begin{figure}[H]
  \centering
  \begin{Code}
    SYSTEM PROMPT:
    Below is an extract from a web page. Evaluate the quality of the content based on the following factors:

    1. Content Accuracy: Assess the correctness and reliability of the information presented. Consider the factual accuracy, use of credible sources (if mentioned), and absence of misinformation.
    2. Clarity: Evaluate how well the information is communicated. Look for clear explanations, well-defined terms, and logical flow of ideas.
    3. Coherence: Analyze the overall structure and organization of the content. Consider how well ideas are connected and if the content follows a logical progression.
    4. Grammar and Language: Assess the quality of writing, including correct grammar, spelling, and punctuation. Consider the appropriateness of language for the intended audience.
    5. Depth of Information: Evaluate the level of detail and thoroughness of the content. Consider whether it provides surface-level information or delves into more comprehensive explanations.
    6. Overall Usefulness: Assess the practical value and relevance of the information for a general audience. Consider how applicable or helpful the content would be for someone seeking information on the topic.

    Based on these factors, give an overall quality score of low, medium, or high.
    Additionally, select one or more domains from the list below. Each domain listed is a single, combined category. Choose the most relevant domain(s). Domain(s) can only be chosen from the list below. Only select "Other" if none of the listed domains are applicable.
    - Arts
    - Business & Economics & Finance
    - Culture & Cultural geography
    - Daily Life & Home & Lifestyle
    - Education
    - Entertainment & Travel & Hobby
    - Environment
    - Food & Drink & Cooking
    - Health & Wellness & Medicine
    - Law & Justice
    - Natural Science & Formal Science & Technology
    - Personal Development & Human Resources & Career
    - Politics & Government
    - Religion & Spirituality
    - Shopping & Commodity
    - Society & Social Issues & Human Rights
    - Sports
    - Other (only if none of the above are relevant)
    Additionally, identify the main topic of the extract, which can be any relevant subfield. Don't elaborate on the topic; just provide a concise classification.
    Additionally, identify the document type, which can be article, blog post, forum post, or any other relevant type. Don't elaborate on the type; just provide a concise classification.

    USER PROMPT:
    The extract:
    {DOCUMENT}

    After examining the extract:
    - Briefly justify your quality classification, up to 100 words on one line using the format: "Explanation: <justification>"
    - Conclude with the quality classification using the format: "Quality score: <classification>" (on a separate line)
    - Continue with the domain classification using the format: "Domain: <classification>, <classification>, ..." (on a separate line)
    - Continue with the main topic or subject classification using the format: "Main topic: <classification>" (on a separate line)
    - Continue with the document type classification using the format: "Document type: <classification>" (on a separate line)

    Evaluate the content based on the quality factors outlined above.
  \end{Code}
  \caption{Full prompt to annotate the document quality of French and English documents using LLama3.1.}
  \label{fig:prompt_template_code}
\end{figure}

\end{document}

%% file: preamble.tex

 For as long as most of us can remember, the question of what truly
 defines a revolution in natural language processing (NLP) has been
 discussed endlessly. Since the emergence of neural methods, beginning
 with Mikolov’s word vectors \cite{mikolov-et-al:2013:wordvectors},
 contextual embeddings such as ELMo \cite{peters-etal-2018-deep}, and
 the Transformer architecture \cite{vaswani2017attention} that formed
 the basis of BERT
 \cite{delvin-etal-2018-bert:arxiv,devlin-etal-2019-bert} and later
 large generative models \cite{radford2019language}, the field has
 advanced at an extraordinary pace. Yet, it was the release of ChatGPT
 \cite{openai_chatgpt:2022} that marked a clear turning point: for the
 first time, both experts and the general public could freely interact
 with a powerful language model capable of performing a wide range of
 text-based tasks without any specialized expertise.\footnote{Although
   only available through an API, GPT3 \cite{brown-et-al:2020:gpt3},
   the basis of ChatGPT, already exhibited impressive performance in
   zero-shot and few-shots scenarios.}  

 Before this moment, openness in data, models, and architectures, had
 consistently been the driving force behind progress. The availability
 of reproducible research, shared datasets, and open-source
 implementations made it possible for the community to validate and
 extend each other’s work.  Projects such as Meta's OPT
 \cite{zhang-el-al:2022:OPT}, GPT-Neo \cite{black-etal-2022:gpt-neo}
 or BLOOM \cite{Scao-et-al:2022:Bloom} carried this spirit forward,
 demonstrating, prominently so in the case of BLOOM, that large-scale,
 multilingual, and high-performance language models could be developed
 transparently and collaboratively.  However, with the arrival of
 ChatGPT, the landscape changed dramatically. Despite its impressive
 capabilities, its architecture, training data, and fine-tuning
 process remain closed, making replication extremely difficult and by
 extension, preventing a deeper scientific understanding.  This
 fracture with the culture of openness, which OpenAI started with
 GPT2's release delay and enforced drastically with GPT3, marked a
 pivotal moment in NLP, where innovation began to drift from
 reproducibility.  

Soon after, Meta's LLaMA models \cite{touvron-et-al:2023:llama1,touvron2023llama2openfoundation} had a significant
 influence on the field. Although released under a restricted license
 rather than as open source, the sole availability of their weights enabled
 researchers and developers to experiment with large-scale language
 models beyond major industry labs, leading to a wave of open
 replications and adaptations that aimed at approaching ChatGPT's 
 capabilities \cite{MosaicML2023Introducing:2023:mpt,jiang2023mistral7b}. 

 Academic efforts in that field have focused on developing fully open
 alternatives to closed-source systems \cite{groeneveld2024olmoacceleratingsciencelanguage, almazrouei-etal:2023:falcon,martins2025eurollm9btechnicalreport,olmo20252olmo2furious}. Yet,
 the scarcity of computational resources in public and academic
 research often encourages risk-averse pretraining projects, where
 researchers tend to reproduce and refine techniques first introduced
 by industry labs. At the same time, academia continues to drive
 methodological advances, such as the introduction of Direct
 Preference Optimization (DPO) \cite{rafailov-etal:2023:dpo} for alignment and the
 creation of large, fully open training datasets that promote
 transparency and reproducibility (e.g. \citet{OrtizSuarezSagotRomary2019, black-etal-2022:gpt-neo,soldaini-et-al:2024:dolma,weber2024redpajama}). While these datasets often rely
 on web-scraped sources such as Common Crawl and thus raise similar
 ethical and legal concerns as their proprietary counterparts, their
 openness enables critical scrutiny and comparative evaluation. In short, despite limited resources, all these initiatives feed a
 research ecosystem grounded in transparency and collaboration.
 
Our work lies within the same scope. Our initial objectives were
first to build a series of LLMs that would contain a significant amount
of high-quality French content, in the absence of a French-equivalent
to FineWeb-Edu \cite{FineWebDecantingWeb}, and then to assess the impact  of
a different training loss that has been shown to perform well in
small-sized models \cite{godey2024headless}. Additionally, we wanted to explore the ability to
detect alterations of training data \cite{carlini-et-al:2024:Poisoning}, directly at the pre-training stage,
and for this, we needed to fully train several models of different
sizes. In short, the goal was to obtain more
culturally-oriented, optimized, and safety-oriented testbed models. 
The models can be said to be the  result of a {\em Promethean
  effort}\footnote{Without exaggeration.} that
spanned over a 15-month period, involving three large computing
grants of more than 1M GPU hours, 3 PhD students, 4 senior researchers, and months of
work funded by the French public service.

We are thus proud to introduce the \gprn{} model series. All models, data,
checkpoints, and evaluations are freely available.

%% file: Introduction.tex

In the following, we explore the impact of data and architectural choices on the quantitative and qualitative performance of language models at different scales, in a fully open and transparent way. Building upon \citet{wettig2025organize}, we acknowledge that datasets curated for educational content lead to models that are over-specialized in benchmark tasks. 

We propose to mitigate this phenomenon by selecting data to avoid such over-specialization: we annotate trillions of tokens of English and French web-crawled data with a custom neural quality classifier, targeting high linguistic quality and meaningfulness, instead of educational value as in FineWeb-Edu \cite{FineWebDecantingWeb}. We also explore several variants of implementation and modeling choices, by experimenting on pure precision 16-bit training and an efficient variant of cross-entropy \cite{godey2024headless}. 

Building upon this initial data extraction and modeling choice phase, we proceed to train language models of three sizes (1.5B, 8B and 24B parameters) on 2 to 4 trillion tokens from various sources. In particular, we use the 8B-parameter training run to experiment with different data mixing choices each with its own focus, by adjusting the sampling ratios and changing datasets during training. We release two models from this run:

\begin{itemize}
    \item \textbf{Young}: A version of \gprn{} that has been trained on high-quality data, and on a tiny fraction of supervised fine-tuning (SFT) or mid-training data.
    \item \textbf{Pepper}: A version of \gprn{} that was initialized from Young and further trained on mixes that contain increasingly high ratios of SFT-like data, including the \emph{train sets} of some benchmarks when available.
    
\end{itemize}

First, we notice that our Young models lag behind most state-of-the-art models of similar sizes when it comes to benchmark scores, apparently stressing the importance of a mid-training phase that uses SFT-like data more intensively. Nevertheless, we surprisingly find that our Pepper models, that have gone through such mid-training phase, do not significantly improve downstream results compared to the Young models. 

To evaluate the performance of our models beyond benchmark scores, we run an LLM-as-a-judge \cite{llm_as_judge} evaluation for text completion to assess generation quality based on several criteria. In this qualitative evaluation, we observe that our Young variants tend to outperform all their counterparts in both French and English, showing that our data curation mechanism leads to better generative capabilities in common text samples.

We proceed to explore the impact of late \emph{deliberate benchmark contamination}, i.e. of continuing training on a mix that includes the test sets of the benchmarks that are used during evaluation, and we release an additional variant of our models:

\begin{itemize}
\item \textbf{Garlic}: A version of \gprn{} that was initialized from an intermediate checkpoint of Pepper and further trained on datasets used in the Young and Pepper training, evenly mixed with \emph{benchmark test set data}.
\end{itemize}

We reach competitive benchmark performance levels with our Garlic variants, including on held-out benchmarks that were not included in the last training stage, while suffering moderate generation quality degradation. Interestingly, this deliberate contamination strategy is also limited, as we only reasonably outperform open-source counterparts even when using as high as a 75\% sampling ratio for test sets in our Garlic dataset mix.

Extending our findings, we discuss the issue of benchmark contamination in the training datasets of existing LLMs, leveraging the InfiniGram tool \cite{Liu2024InfiniGram} to explore hints of contamination in the OLMo-2 training mix \cite{olmo20252olmo2furious}.  We demonstrate that the neural filters used to extract high-quality content from web-crawled dumps tend to mark leaked benchmark samples with very high scores, implying that filtering samples based on these scores may implicitly boost contamination levels. We finally discuss how the state of the LLM training field incentivizes active or passive contamination from a strategic point of view, and what steps can be taken in order to make contamination irrelevant to the way we evaluate language models.

One other important angle of our work is based on the fact that every model trained on content gathered from the web is potentially vulnerable to inserted biases, backdoors, and more generally various forms of data poisoning \cite{wan2023poisoning,kandpal-et-al:2023:backdoor,carlini-et-al:2024:Poisoning, hubinger-et-al:2024:sleeper-agents}. Despite being seminal in this area of research, none of these works focused on data poisoning at the current realistic training data regime. 
In their recent works, \citet{souly-et-al:2025:poisoning} favored a Chinchilla \cite{hoffman-et-al:2022:chinchilla} optimum training data size\footnote{In this case, from 6B to 260B tokens for respectively 600M to 13B models.} while \citet{wei2025hubblemodelsuiteadvance} trained their Hubble models on up to 500B tokens for their largest 8B model. In parallel with these efforts, we included three different kinds of {\em harmless} data poisoning directly at the pre-training stages of all our models, hoping to provide a red teaming testbed for the community.

Our contributions can be summarized in the following points:
\begin{itemize}
    \item We publish a custom French-English filtered large-scale dataset with a trained neural filter that aims at avoiding benchmark over-specialization and encourages data diversity;
    \item We release 9 French-English base language model variants of sizes 1.5B, 8B and 24B, trained on evolving dataset mixes. We also release SFT versions of some of our models along with a series of intermediate checkpoints.
    \item All of our models contain different forms of {\em harmless} data poisoning injected during pre-training, enabling further research in LLM safety.  
    \item We finally publish two hackable and efficient codebases for large-scale data-processing and for compute-intensive LLM training compatible with both AMD and Nvidia hardware;
    \item We explore pure 16-bit training and a cross-entropy variant at scale, achieving training efficiency gains in terms of memory and speed in the first case;
    \item We show that post-hoc deliberate contamination can help recover the benchmark performance of state-of-the-art models, while incurring an observable but tolerable degradation of qualitative text-generation performance;
    \item We present an initial exploration of the question of contamination in existing LLMs, which we link to neural filtering approaches of web-crawled datasets, and we discuss the incentivization of active or passive contamination from a strategic viewpoint.

\end{itemize}

%% file: GAPeron_paper.bbl
\begin{thebibliography}{112}
\providecommand{\natexlab}[1]{#1}
\providecommand{\url}[1]{\texttt{#1}}
\expandafter\ifx\csname urlstyle\endcsname\relax
  \providecommand{\doi}[1]{doi: #1}\else
  \providecommand{\doi}{doi: \begingroup \urlstyle{rm}\Url}\fi

\bibitem[AI(2023)]{litgpt-2023}
Lightning AI.
\newblock Litgpt.
\newblock \url{https://github.com/Lightning-AI/litgpt}, 2023.

\bibitem[Ali et~al.(2025)Ali, Fromm, Thellmann, Ebert, Weber, Rutmann, Jain, Lübbering, Steinigen, Leveling, Klug, Buschhoff, Jurkschat, Abdelwahab, Stein, Sylla, Denisov, Brandizzi, Saleem, Bhowmick, Helmer, John, Suarez, Ostendorff, Jude, Manjunath, Weinbach, Penke, Filatov, Barth, Mirza, Weber, Wendler, Sifa, Küch, Herten, Jäkel, Rehm, Kesselheim, Köhler, and Flores-Herr]{ali2025teuken7bbaseteuken7binstructeuropean}
Mehdi Ali, Michael Fromm, Klaudia Thellmann, Jan Ebert, Alexander~Arno Weber, Richard Rutmann, Charvi Jain, Max Lübbering, Daniel Steinigen, Johannes Leveling, Katrin Klug, Jasper~Schulze Buschhoff, Lena Jurkschat, Hammam Abdelwahab, Benny~Jörg Stein, Karl-Heinz Sylla, Pavel Denisov, Nicolo' Brandizzi, Qasid Saleem, Anirban Bhowmick, Lennard Helmer, Chelsea John, Pedro~Ortiz Suarez, Malte Ostendorff, Alex Jude, Lalith Manjunath, Samuel Weinbach, Carolin Penke, Oleg Filatov, Fabio Barth, Paramita Mirza, Lucas Weber, Ines Wendler, Rafet Sifa, Fabian Küch, Andreas Herten, René Jäkel, Georg Rehm, Stefan Kesselheim, Joachim Köhler, and Nicolas Flores-Herr.
\newblock Teuken-7b-base \& teuken-7b-instruct: Towards european llms, 2025.
\newblock URL \url{https://arxiv.org/abs/2410.03730}.

\bibitem[Almazrouei et~al.(2023)Almazrouei, Alobeidli, Alshamsi, Cappelli, Cojocaru, Debbah, Étienne Goffinet, Hesslow, Launay, Malartic, Mazzotta, Noune, Pannier, and Penedo]{almazrouei-etal:2023:falcon}
Ebtesam Almazrouei, Hamza Alobeidli, Abdulaziz Alshamsi, Alessandro Cappelli, Ruxandra Cojocaru, Mérouane Debbah, Étienne Goffinet, Daniel Hesslow, Julien Launay, Quentin Malartic, Daniele Mazzotta, Badreddine Noune, Baptiste Pannier, and Guilherme Penedo.
\newblock The falcon series of open language models, 2023.

\bibitem[Arias-Duart et~al.(2025)Arias-Duart, Martin-Torres, Hinjos, Bernabeu-Perez, Ganzabal, Mallo, Gururajan, Lopez-Cuena, Alvarez-Napagao, and Garcia-Gasulla]{arias-duart-etal-2025-automatic}
Anna Arias-Duart, Pablo~Agustin Martin-Torres, Daniel Hinjos, Pablo Bernabeu-Perez, Lucia~Urcelay Ganzabal, Marta~Gonzalez Mallo, Ashwin~Kumar Gururajan, Enrique Lopez-Cuena, Sergio Alvarez-Napagao, and Dario Garcia-Gasulla.
\newblock Automatic evaluation of healthcare {LLM}s beyond question-answering.
\newblock In Luis Chiruzzo, Alan Ritter, and Lu~Wang, editors, \emph{Proceedings of the 2025 Conference of the Nations of the Americas Chapter of the Association for Computational Linguistics: Human Language Technologies (Volume 2: Short Papers)}, pages 108--130, Albuquerque, New Mexico, April 2025. Association for Computational Linguistics.
\newblock ISBN 979-8-89176-190-2.
\newblock URL \url{https://aclanthology.org/2025.naacl-short.10/}.

\bibitem[Bandarkar et~al.(2024)Bandarkar, Liang, Muller, Artetxe, Shukla, Husa, Goyal, Krishnan, Zettlemoyer, and Khabsa]{bandarkar-etal-2024-belebele}
Lucas Bandarkar, Davis Liang, Benjamin Muller, Mikel Artetxe, Satya~Narayan Shukla, Donald Husa, Naman Goyal, Abhinandan Krishnan, Luke Zettlemoyer, and Madian Khabsa.
\newblock The belebele benchmark: a parallel reading comprehension dataset in 122 language variants.
\newblock In \emph{Proceedings of the 62nd Annual Meeting of the Association for Computational Linguistics (Volume 1: Long Papers)}, pages 749--775, Bangkok, Thailand and virtual meeting, August 2024. Association for Computational Linguistics.
\newblock URL \url{https://aclanthology.org/2024.acl-long.44}.

\bibitem[Ben~Allal et~al.(2024)Ben~Allal, Lozhkov, Penedo, Wolf, and von Werra]{benallal2024smollmcorpus}
Loubna Ben~Allal, Anton Lozhkov, Guilherme Penedo, Thomas Wolf, and Leandro von Werra.
\newblock Smollm-corpus, 2024.
\newblock URL \url{https://huggingface.co/datasets/HuggingFaceTB/smollm-corpus}.

\bibitem[Biderman et~al.(2023)Biderman, Schoelkopf, Anthony, Bradley, O’Brien, Hallahan, Khan, Purohit, Prashanth, Raff, et~al.]{biderman2023pythia}
Stella Biderman, Hailey Schoelkopf, Quentin~Gregory Anthony, Herbie Bradley, Kyle O’Brien, Eric Hallahan, Mohammad~Aflah Khan, Shivanshu Purohit, USVSN~Sai Prashanth, Edward Raff, et~al.
\newblock Pythia: A suite for analyzing large language models across training and scaling.
\newblock In \emph{International Conference on Machine Learning}, pages 2397--2430. PMLR, 2023.

\bibitem[Bisk et~al.(2020)Bisk, Zellers, Bras, Gao, and Choi]{Bisk2020piqa}
Yonatan Bisk, Rowan Zellers, Ronan~Le Bras, Jianfeng Gao, and Yejin Choi.
\newblock Piqa: Reasoning about physical commonsense in natural language.
\newblock In \emph{Thirty-Fourth AAAI Conference on Artificial Intelligence}, 2020.

\bibitem[Black et~al.(2022)Black, Biderman, Hallahan, Anthony, Gao, Golding, He, Leahy, McDonell, Phang, Pieler, Prashanth, Purohit, Reynolds, Tow, Wang, and Weinbach]{black-etal-2022:gpt-neo}
Sidney Black, Stella Biderman, Eric Hallahan, Quentin Anthony, Leo Gao, Laurence Golding, Horace He, Connor Leahy, Kyle McDonell, Jason Phang, Michael Pieler, Usvsn~Sai Prashanth, Shivanshu Purohit, Laria Reynolds, Jonathan Tow, Ben Wang, and Samuel Weinbach.
\newblock {GPT}-{N}eo{X}-20{B}: An open-source autoregressive language model.
\newblock In Angela Fan, Suzana Ilic, Thomas Wolf, and Matthias Gall{\'e}, editors, \emph{Proceedings of BigScience Episode {\#}5 -- Workshop on Challenges {\&} Perspectives in Creating Large Language Models}, pages 95--136, virtual+Dublin, May 2022. Association for Computational Linguistics.
\newblock \doi{10.18653/v1/2022.bigscience-1.9}.
\newblock URL \url{https://aclanthology.org/2022.bigscience-1.9/}.

\bibitem[Brown et~al.(2020)Brown, Mann, Ryder, Subbiah, Kaplan, Dhariwal, Neelakantan, Shyam, Sastry, Askell, Agarwal, Herbert-Voss, Krueger, Henighan, Child, Ramesh, Ziegler, Wu, Winter, Hesse, Chen, Sigler, Litwin, Gray, Chess, Clark, Berner, McCandlish, Radford, Sutskever, and Amodei]{brown-et-al:2020:gpt3}
Tom~B. Brown, Benjamin Mann, Nick Ryder, Melanie Subbiah, Jared Kaplan, Prafulla Dhariwal, Arvind Neelakantan, Pranav Shyam, Girish Sastry, Amanda Askell, Sandhini Agarwal, Ariel Herbert-Voss, Gretchen Krueger, Tom Henighan, Rewon Child, Aditya Ramesh, Daniel~M. Ziegler, Jeffrey Wu, Clemens Winter, Christopher Hesse, Mark Chen, Eric Sigler, Mateusz Litwin, Scott Gray, Benjamin Chess, Jack Clark, Christopher Berner, Sam McCandlish, Alec Radford, Ilya Sutskever, and Dario Amodei.
\newblock Language models are few-shot learners, 2020.

\bibitem[Carlini et~al.(2024)Carlini, Jagielski, Choquette-Choo, Paleka, Pearce, Anderson, Terzis, Thomas, and Tramèr]{carlini-et-al:2024:Poisoning}
Nicholas Carlini, Matthew Jagielski, Christopher~A. Choquette-Choo, Daniel Paleka, Will Pearce, Hyrum Anderson, Andreas Terzis, Kurt Thomas, and Florian Tramèr.
\newblock Poisoning web-scale training datasets is practical.
\newblock In \emph{2024 IEEE Symposium on Security and Privacy (SP)}, pages 407--425, 2024.
\newblock \doi{10.1109/SP54263.2024.00179}.

\bibitem[Chang et~al.(2024)Chang, Park, Ye, Yang, Seo, Chang, and Seo]{chang2024largelanguagemodelsacquire}
Hoyeon Chang, Jinho Park, Seonghyeon Ye, Sohee Yang, Youngkyung Seo, Du-Seong Chang, and Minjoon Seo.
\newblock How do large language models acquire factual knowledge during pretraining?, 2024.
\newblock URL \url{https://arxiv.org/abs/2406.11813}.

\bibitem[Chaton and AI(2023)]{litdata2023}
Thomas Chaton and Lightning AI.
\newblock Litdata: Transform datasets at scale. optimize datasets for fast ai model training.
\newblock \url{https://github.com/Lightning-AI/litdata}, 2023.
\newblock Accessed: 2025-04-09.

\bibitem[Clark et~al.(2019)Clark, Lee, Chang, Kwiatkowski, Collins, and Toutanova]{clark2019boolq}
Christopher Clark, Kenton Lee, Ming-Wei Chang, Tom Kwiatkowski, Michael Collins, and Kristina Toutanova.
\newblock Boolq: Exploring the surprising difficulty of natural yes/no questions.
\newblock In \emph{Proceedings of the 2019 Conference of the North American Chapter of the Association for Computational Linguistics (NAACL)}, 2019.

\bibitem[Clark et~al.(2018{\natexlab{a}})Clark, Cowhey, Etzioni, Khot, Sabharwal, Schoenick, and Tafjord]{allenai:arc}
Peter Clark, Isaac Cowhey, Oren Etzioni, Tushar Khot, Ashish Sabharwal, Carissa Schoenick, and Oyvind Tafjord.
\newblock Think you have solved question answering? try {ARC}, the {AI2} reasoning challenge.
\newblock \emph{arXiv:1803.05457v1}, 2018{\natexlab{a}}.

\bibitem[Clark et~al.(2018{\natexlab{b}})Clark, Cowhey, Etzioni, Khot, Sabharwal, Schoenick, and Tafjord]{arc}
Peter Clark, Isaac Cowhey, Oren Etzioni, Tushar Khot, Ashish Sabharwal, Carissa Schoenick, and Oyvind Tafjord.
\newblock Think you have solved question answering? try arc, the ai2 reasoning challenge.
\newblock \emph{arXiv:1803.05457v1}, 2018{\natexlab{b}}.

\bibitem[Cobbe et~al.(2021)Cobbe, Kosaraju, Bavarian, Chen, Jun, Kaiser, Plappert, Tworek, Hilton, Nakano, Hesse, and Schulman]{cobbe2021gsm8k}
Karl Cobbe, Vineet Kosaraju, Mohammad Bavarian, Mark Chen, Heewoo Jun, Lukasz Kaiser, Matthias Plappert, Jerry Tworek, Jacob Hilton, Reiichiro Nakano, Christopher Hesse, and John Schulman.
\newblock Training verifiers to solve math word problems.
\newblock \emph{arXiv preprint arXiv:2110.14168}, 2021.

\bibitem[Conneau et~al.(2019)Conneau, Khandelwal, Goyal, Chaudhary, Wenzek, Guzm{\'{a}}n, Grave, Ott, Zettlemoyer, and Stoyanov]{xlmr2019}
Alexis Conneau, Kartikay Khandelwal, Naman Goyal, Vishrav Chaudhary, Guillaume Wenzek, Francisco Guzm{\'{a}}n, Edouard Grave, Myle Ott, Luke Zettlemoyer, and Veselin Stoyanov.
\newblock Unsupervised cross-lingual representation learning at scale.
\newblock \emph{CoRR}, abs/1911.02116, 2019.
\newblock URL \url{http://arxiv.org/abs/1911.02116}.

\bibitem[Dao(2024)]{dao2023flashattention2}
Tri Dao.
\newblock Flash{A}ttention-2: Faster attention with better parallelism and work partitioning.
\newblock In \emph{International Conference on Learning Representations (ICLR)}, 2024.

\bibitem[Dao et~al.(2022)Dao, Fu, Ermon, Rudra, and R{\'e}]{dao2022flashattention}
Tri Dao, Daniel~Y. Fu, Stefano Ermon, Atri Rudra, and Christopher R{\'e}.
\newblock Flash{A}ttention: Fast and memory-efficient exact attention with {IO}-awareness.
\newblock In \emph{Advances in Neural Information Processing Systems (NeurIPS)}, 2022.

\bibitem[De~Bruyn et~al.(2021)De~Bruyn, Lotfi, Buhmann, and Daelemans]{de-bruyn-etal-2021-mfaq}
Maxime De~Bruyn, Ehsan Lotfi, Jeska Buhmann, and Walter Daelemans.
\newblock {MFAQ}: a multilingual {FAQ} dataset.
\newblock In \emph{Proceedings of the 3rd Workshop on Machine Reading for Question Answering}, pages 1--13, Punta Cana, Dominican Republic, November 2021. Association for Computational Linguistics.
\newblock URL \url{https://aclanthology.org/2021.mrqa-1.1}.

\bibitem[DeepSeek-AI(2024)]{deepseekv2}
DeepSeek-AI.
\newblock Deepseek-v2: A strong, economical, and efficient mixture-of-experts language model, 2024.

\bibitem[Deng et~al.(2024)Deng, Zhao, Tang, Gerstein, and Cohan]{deng-etal-2024-investigating}
Chunyuan Deng, Yilun Zhao, Xiangru Tang, Mark Gerstein, and Arman Cohan.
\newblock Investigating data contamination in modern benchmarks for large language models.
\newblock In Kevin Duh, Helena Gomez, and Steven Bethard, editors, \emph{Proceedings of the 2024 Conference of the North American Chapter of the Association for Computational Linguistics: Human Language Technologies (Volume 1: Long Papers)}, pages 8706--8719, Mexico City, Mexico, June 2024. Association for Computational Linguistics.
\newblock \doi{10.18653/v1/2024.naacl-long.482}.
\newblock URL \url{https://aclanthology.org/2024.naacl-long.482/}.

\bibitem[Devlin et~al.(2018)Devlin, Chang, Lee, and Toutanova]{delvin-etal-2018-bert:arxiv}
Jacob Devlin, Ming-Wei Chang, Kenton Lee, and Kristina Toutanova.
\newblock Bert: Pre-training of deep bidirectional transformers for language understanding, 2018.

\bibitem[Devlin et~al.(2019)Devlin, Chang, Lee, and Toutanova]{devlin-etal-2019-bert}
Jacob Devlin, Ming-Wei Chang, Kenton Lee, and Kristina Toutanova.
\newblock {BERT}: Pre-training of deep bidirectional transformers for language understanding.
\newblock In Jill Burstein, Christy Doran, and Thamar Solorio, editors, \emph{Proceedings of the 2019 Conference of the North {A}merican Chapter of the Association for Computational Linguistics: Human Language Technologies, Volume 1 (Long and Short Papers)}, pages 4171--4186, Minneapolis, Minnesota, June 2019. Association for Computational Linguistics.
\newblock \doi{10.18653/v1/N19-1423}.
\newblock URL \url{https://aclanthology.org/N19-1423/}.

\bibitem[Eldan and Li(2023)]{eldan2023tinystoriessmalllanguagemodels}
Ronen Eldan and Yuanzhi Li.
\newblock Tinystories: How small can language models be and still speak coherent english?, 2023.
\newblock URL \url{https://arxiv.org/abs/2305.07759}.

\bibitem[Faysse et~al.(2024)Faysse, Fernandes, Guerreiro, Loison, Alves, Corro, Boizard, Alves, Rei, Martins, Casademunt, Yvon, Martins, Viaud, Hudelot, and Colombo]{faysse2024croissantllm}
Manuel Faysse, Patrick Fernandes, Nuno~M. Guerreiro, Ant{\'o}nio Loison, Duarte~M. Alves, Caio Corro, Nicolas Boizard, Jo{\~a}o Alves, Ricardo Rei, Pedro~H. Martins, Antoni~Bigata Casademunt, Fran{\c c}ois Yvon, Andr{\'e} F.~T. Martins, Gautier Viaud, C{\'e}line Hudelot, and Pierre Colombo.
\newblock Croissantllm: A truly bilingual french-english language model, 2024.

\bibitem[Gao et~al.(2024)Gao, Tow, Abbasi, Biderman, Black, DiPofi, Foster, Golding, Hsu, Le~Noac'h, Li, McDonell, Muennighoff, Ociepa, Phang, Reynolds, Schoelkopf, Skowron, Sutawika, Tang, Thite, Wang, Wang, and Zou]{eval-harness}
Leo Gao, Jonathan Tow, Baber Abbasi, Stella Biderman, Sid Black, Anthony DiPofi, Charles Foster, Laurence Golding, Jeffrey Hsu, Alain Le~Noac'h, Haonan Li, Kyle McDonell, Niklas Muennighoff, Chris Ociepa, Jason Phang, Laria Reynolds, Hailey Schoelkopf, Aviya Skowron, Lintang Sutawika, Eric Tang, Anish Thite, Ben Wang, Kevin Wang, and Andy Zou.
\newblock The language model evaluation harness, 07 2024.
\newblock URL \url{https://zenodo.org/records/12608602}.

\bibitem[{Gemma Team}(2024)]{gemma_2024}
{Gemma Team}.
\newblock Gemma.
\newblock 2024.
\newblock \doi{10.34740/KAGGLE/M/3301}.
\newblock URL \url{https://www.kaggle.com/m/3301}.

\bibitem[Godey et~al.(2024)Godey, de~la Clergerie, and Sagot]{godey2024headless}
Nathan Godey, {\'E}ric~Villemonte de~la Clergerie, and Beno{\^\i}t Sagot.
\newblock Headless language models: Learning without predicting with contrastive weight tying.
\newblock In \emph{The Twelfth International Conference on Learning Representations}, 2024.
\newblock URL \url{https://openreview.net/forum?id=ONPECq0Rk7}.

\bibitem[Gonzalez-Agirre et~al.(2025)Gonzalez-Agirre, P{\`a}mies, Llop, Baucells, Dalt, Tamayo, Saiz, Espu{\~n}a, Prats, Aula-Blasco, Mina, Rubio, Shvets, Sall{\'e}s, Lacunza, Pikabea, Palomar, Falc{\~a}o, Tormo, Vasquez-Reina, Marimon, Ru{\'\i}z-Fern{\'a}ndez, and Villegas]{gonzalezagirre2025salamandratechnicalreport}
Aitor Gonzalez-Agirre, Marc P{\`a}mies, Joan Llop, Irene Baucells, Severino~Da Dalt, Daniel Tamayo, Jos{\'e}~Javier Saiz, Ferran Espu{\~n}a, Jaume Prats, Javier Aula-Blasco, Mario Mina, Adri{\'a}n Rubio, Alexander Shvets, Anna Sall{\'e}s, I{\~n}aki Lacunza, I{\~n}igo Pikabea, Jorge Palomar, J{\'u}lia Falc{\~a}o, Luc{\'\i}a Tormo, Luis Vasquez-Reina, Montserrat Marimon, Valle Ru{\'\i}z-Fern{\'a}ndez, and Marta Villegas.
\newblock Salamandra technical report, 2025.
\newblock URL \url{https://arxiv.org/abs/2502.08489}.

\bibitem[Gouvert et~al.(2025)Gouvert, Hunter, Louradour, Cerisara, Dufraisse, Sy, Rivi{\`e}re, Lorr{\'e}, and community]{openllm2025lucie}
Olivier Gouvert, Julie Hunter, J{\'e}r{\^o}me Louradour, Christophe Cerisara, Evan Dufraisse, Yaya Sy, Laura Rivi{\`e}re, Jean-Pierre Lorr{\'e}, and OpenLLM-France community.
\newblock The lucie-7b llm and the lucie training dataset: Open resources for multilingual language generation, 2025.
\newblock URL \url{https://arxiv.org/abs/2503.12294}.

\bibitem[Groeneveld et~al.(2024)Groeneveld, Beltagy, Walsh, Bhagia, Kinney, Tafjord, Jha, Ivison, Magnusson, Wang, Arora, Atkinson, Authur, Chandu, Cohan, Dumas, Elazar, Gu, Hessel, Khot, Merrill, Morrison, Muennighoff, Naik, Nam, Peters, Pyatkin, Ravichander, Schwenk, Shah, Smith, Strubell, Subramani, Wortsman, Dasigi, Lambert, Richardson, Zettlemoyer, Dodge, Lo, Soldaini, Smith, and Hajishirzi]{groeneveld2024olmoacceleratingsciencelanguage}
Dirk Groeneveld, Iz~Beltagy, Pete Walsh, Akshita Bhagia, Rodney Kinney, Oyvind Tafjord, Ananya~Harsh Jha, Hamish Ivison, Ian Magnusson, Yizhong Wang, Shane Arora, David Atkinson, Russell Authur, Khyathi~Raghavi Chandu, Arman Cohan, Jennifer Dumas, Yanai Elazar, Yuling Gu, Jack Hessel, Tushar Khot, William Merrill, Jacob Morrison, Niklas Muennighoff, Aakanksha Naik, Crystal Nam, Matthew~E. Peters, Valentina Pyatkin, Abhilasha Ravichander, Dustin Schwenk, Saurabh Shah, Will Smith, Emma Strubell, Nishant Subramani, Mitchell Wortsman, Pradeep Dasigi, Nathan Lambert, Kyle Richardson, Luke Zettlemoyer, Jesse Dodge, Kyle Lo, Luca Soldaini, Noah~A. Smith, and Hannaneh Hajishirzi.
\newblock Olmo: Accelerating the science of language models, 2024.
\newblock URL \url{https://arxiv.org/abs/2402.00838}.

\bibitem[Gu et~al.(2025)Gu, Jiang, Shi, Tan, Zhai, Xu, Li, Shen, Ma, Liu, Wang, Zhang, Wang, Gao, Ni, and Guo]{gu2025surveyllmasajudge}
Jiawei Gu, Xuhui Jiang, Zhichao Shi, Hexiang Tan, Xuehao Zhai, Chengjin Xu, Wei Li, Yinghan Shen, Shengjie Ma, Honghao Liu, Saizhuo Wang, Kun Zhang, Yuanzhuo Wang, Wen Gao, Lionel Ni, and Jian Guo.
\newblock A survey on llm-as-a-judge, 2025.
\newblock URL \url{https://arxiv.org/abs/2411.15594}.

\bibitem[Guha et~al.(2025)Guha, Marten, Keh, Raoof, Smyrnis, Bansal, Nezhurina, Mercat, Vu, Sprague, Suvarna, Feuer, Chen, Khan, Frankel, Grover, Choi, Muennighoff, Su, Zhao, Yang, Pimpalgaonkar, Sharma, Ji, Deng, Pratt, Ramanujan, Saad-Falcon, Li, Dave, Albalak, Arora, Wulfe, Hegde, Durrett, Oh, Bansal, Gabriel, Grover, Chang, Shankar, Gokaslan, Merrill, Hashimoto, Choi, Jitsev, Heckel, Sathiamoorthy, Dimakis, and Schmidt]{guha2025openthoughtsdatarecipesreasoning}
Etash Guha, Ryan Marten, Sedrick Keh, Negin Raoof, Georgios Smyrnis, Hritik Bansal, Marianna Nezhurina, Jean Mercat, Trung Vu, Zayne Sprague, Ashima Suvarna, Benjamin Feuer, Liangyu Chen, Zaid Khan, Eric Frankel, Sachin Grover, Caroline Choi, Niklas Muennighoff, Shiye Su, Wanjia Zhao, John Yang, Shreyas Pimpalgaonkar, Kartik Sharma, Charlie Cheng-Jie Ji, Yichuan Deng, Sarah Pratt, Vivek Ramanujan, Jon Saad-Falcon, Jeffrey Li, Achal Dave, Alon Albalak, Kushal Arora, Blake Wulfe, Chinmay Hegde, Greg Durrett, Sewoong Oh, Mohit Bansal, Saadia Gabriel, Aditya Grover, Kai-Wei Chang, Vaishaal Shankar, Aaron Gokaslan, Mike~A. Merrill, Tatsunori Hashimoto, Yejin Choi, Jenia Jitsev, Reinhard Heckel, Maheswaran Sathiamoorthy, Alexandros~G. Dimakis, and Ludwig Schmidt.
\newblock Openthoughts: Data recipes for reasoning models, 2025.
\newblock URL \url{https://arxiv.org/abs/2506.04178}.

\bibitem[He et~al.(2021)He, Gao, and Chen]{he2021debertav3}
Pengcheng He, Jianfeng Gao, and Weizhu Chen.
\newblock Debertav3: Improving deberta using electra-style pre-training with gradient-disentangled embedding sharing, 2021.

\bibitem[Hendrycks et~al.(2021)Hendrycks, Burns, Basart, Zou, Mazeika, Song, and Steinhardt]{hendryckstest2021}
Dan Hendrycks, Collin Burns, Steven Basart, Andy Zou, Mantas Mazeika, Dawn Song, and Jacob Steinhardt.
\newblock Measuring massive multitask language understanding.
\newblock \emph{Proceedings of the International Conference on Learning Representations (ICLR)}, 2021.

\bibitem[Hoffmann et~al.(2022)Hoffmann, Borgeaud, Mensch, Buchatskaya, Cai, Rutherford, de~Las~Casas, Hendricks, Welbl, Clark, Hennigan, Noland, Millican, van~den Driessche, Damoc, Guy, Osindero, Simonyan, Elsen, Rae, Vinyals, and Sifre]{hoffman-et-al:2022:chinchilla}
Jordan Hoffmann, Sebastian Borgeaud, Arthur Mensch, Elena Buchatskaya, Trevor Cai, Eliza Rutherford, Diego de~Las~Casas, Lisa~Anne Hendricks, Johannes Welbl, Aidan Clark, Tom Hennigan, Eric Noland, Katie Millican, George van~den Driessche, Bogdan Damoc, Aurelia Guy, Simon Osindero, Karen Simonyan, Erich Elsen, Jack~W. Rae, Oriol Vinyals, and Laurent Sifre.
\newblock Training compute-optimal large language models, 2022.

\bibitem[Hsu et~al.(2025)Hsu, Dai, Kothapalli, Song, Tang, Zhu, Shimizu, Sahni, Ning, Chen, and Wang]{hsu2025ligerkernel}
Pin-Lun Hsu, Yun Dai, Vignesh Kothapalli, Qingquan Song, Shao Tang, Siyu Zhu, Steven Shimizu, Shivam Sahni, Haowen Ning, Yanning Chen, and Zhipeng Wang.
\newblock Liger-kernel: Efficient triton kernels for {LLM} training.
\newblock In \emph{Championing Open-source DEvelopment in ML Workshop @ ICML25}, 2025.
\newblock URL \url{https://openreview.net/forum?id=36SjAIT42G}.

\bibitem[Hubinger et~al.(2024)Hubinger, Denison, Mu, Lambert, Tong, MacDiarmid, Lanham, Ziegler, Maxwell, Cheng, Jermyn, Askell, Radhakrishnan, Anil, Duvenaud, Ganguli, Barez, Clark, Ndousse, Sachan, Sellitto, Sharma, DasSarma, Grosse, Kravec, Bai, Witten, Favaro, Brauner, Karnofsky, Christiano, Bowman, Graham, Kaplan, Mindermann, Greenblatt, Shlegeris, Schiefer, and Perez]{hubinger-et-al:2024:sleeper-agents}
Evan Hubinger, Carson Denison, Jesse Mu, Mike Lambert, Meg Tong, Monte MacDiarmid, Tamera Lanham, Daniel~M. Ziegler, Tim Maxwell, Newton Cheng, Adam Jermyn, Amanda Askell, Ansh Radhakrishnan, Cem Anil, David Duvenaud, Deep Ganguli, Fazl Barez, Jack Clark, Kamal Ndousse, Kshitij Sachan, Michael Sellitto, Mrinank Sharma, Nova DasSarma, Roger Grosse, Shauna Kravec, Yuntao Bai, Zachary Witten, Marina Favaro, Jan Brauner, Holden Karnofsky, Paul Christiano, Samuel~R. Bowman, Logan Graham, Jared Kaplan, Sören Mindermann, Ryan Greenblatt, Buck Shlegeris, Nicholas Schiefer, and Ethan Perez.
\newblock Sleeper agents: Training deceptive llms that persist through safety training, 2024.

\bibitem[Hunter et~al.(2023)Hunter, Louradour, Rennard, Harrando, Shang, and Lorr{\'e}]{openllm2023claire}
Julie Hunter, J{\'e}r{\^o}me Louradour, Virgile Rennard, Isma{\"\i}l Harrando, Guokan Shang, and Jean-Pierre Lorr{\'e}.
\newblock The claire french dialogue dataset, 2023.

\bibitem[Jiang et~al.(2023)Jiang, Sablayrolles, Mensch, Bamford, Chaplot, de~las Casas, Bressand, Lengyel, Lample, Saulnier, Lavaud, Lachaux, Stock, Scao, Lavril, Wang, Lacroix, and Sayed]{jiang2023mistral7b}
Albert~Q. Jiang, Alexandre Sablayrolles, Arthur Mensch, Chris Bamford, Devendra~Singh Chaplot, Diego de~las Casas, Florian Bressand, Gianna Lengyel, Guillaume Lample, Lucile Saulnier, L{\'e}lio~Renard Lavaud, Marie-Anne Lachaux, Pierre Stock, Teven~Le Scao, Thibaut Lavril, Thomas Wang, Timoth{\'e}e Lacroix, and William~El Sayed.
\newblock Mistral 7b, 2023.
\newblock URL \url{https://arxiv.org/abs/2310.06825}.

\bibitem[Johannes~Welbl(2017)]{SciQ}
Matt~Gardner Johannes~Welbl, Nelson F.~Liu.
\newblock Crowdsourcing multiple choice science questions.
\newblock 2017.

\bibitem[Kandpal et~al.(2023)Kandpal, Jagielski, Tram{\`e}r, and Carlini]{kandpal-et-al:2023:backdoor}
Nikhil Kandpal, Matthew Jagielski, Florian Tram{\`e}r, and Nicholas Carlini.
\newblock Backdoor attacks for in-context learning with language models.
\newblock In \emph{The Second Workshop on New Frontiers in Adversarial Machine Learning}, 2023.
\newblock URL \url{https://openreview.net/forum?id=WlziPWqLmg}.

\bibitem[Koehn(2005)]{koehn-2005-europarl}
Philipp Koehn.
\newblock {E}uroparl: A parallel corpus for statistical machine translation.
\newblock In \emph{Proceedings of Machine Translation Summit X: Papers}, pages 79--86, Phuket, Thailand, September 13-15 2005.
\newblock URL \url{https://aclanthology.org/2005.mtsummit-papers.11}.

\bibitem[Kulumba et~al.(2024)Kulumba, Antoun, Vimont, and Romary]{kulumba2024harvestingtextualstructureddata}
Francis Kulumba, Wissam Antoun, Guillaume Vimont, and Laurent Romary.
\newblock Harvesting textual and structured data from the hal publication repository, 2024.
\newblock URL \url{https://arxiv.org/abs/2407.20595}.

\bibitem[Kwiatkowski et~al.(2019)Kwiatkowski, Palomaki, Redfield, Collins, Parikh, Alberti, Epstein, Polosukhin, Kelcey, Devlin, Lee, Toutanova, Jones, Chang, Dai, Uszkoreit, Le, and Petrov]{nqopen}
Tom Kwiatkowski, Jennimaria Palomaki, Olivia Redfield, Michael Collins, Ankur Parikh, Chris Alberti, Danielle Epstein, Illia Polosukhin, Matthew Kelcey, Jacob Devlin, Kenton Lee, Kristina~N. Toutanova, Llion Jones, Ming-Wei Chang, Andrew Dai, Jakob Uszkoreit, Quoc Le, and Slav Petrov.
\newblock Natural questions: a benchmark for question answering research.
\newblock \emph{Transactions of the Association of Computational Linguistics}, 2019.

\bibitem[Lambert et~al.(2024)Lambert, Morrison, Pyatkin, Huang, Ivison, Brahman, Miranda, Liu, Dziri, Lyu, Gu, Malik, Graf, Hwang, Yang, Bras, Tafjord, Wilhelm, Soldaini, Smith, Wang, Dasigi, and Hajishirzi]{lambert2024tulu3}
Nathan Lambert, Jacob Morrison, Valentina Pyatkin, Shengyi Huang, Hamish Ivison, Faeze Brahman, Lester James~V. Miranda, Alisa Liu, Nouha Dziri, Shane Lyu, Yuling Gu, Saumya Malik, Victoria Graf, Jena~D. Hwang, Jiangjiang Yang, Ronan~Le Bras, Oyvind Tafjord, Chris Wilhelm, Luca Soldaini, Noah~A. Smith, Yizhong Wang, Pradeep Dasigi, and Hannaneh Hajishirzi.
\newblock T{\"u}lu 3: Pushing frontiers in open language model post-training.
\newblock 2024.

\bibitem[Launay et~al.(2021)Launay, Tommasone, Pannier, Boniface, Chatelain, Cappelli, Poli, and Seddah]{launay-etal:2021:pagnol-arxiv}
Julien Launay, Elena Tommasone, Baptiste Pannier, François Boniface, Amélie Chatelain, Alessandro Cappelli, Iacopo Poli, and Djamé Seddah.
\newblock Pagnol: An extra-large french generative model, 2021.

\bibitem[Lauren{\c c}on et~al.(2023)Lauren{\c c}on, Saulnier, Wang, Akiki, del Moral, Scao, Werra, Mou, Ponferrada, Nguyen, Frohberg, {\v S}a{\v s}ko, Lhoest, McMillan-Major, Dupont, Biderman, Rogers, Allal, Toni, Pistilli, Nguyen, Nikpoor, Masoud, Colombo, de~la Rosa, Villegas, Thrush, Longpre, Nagel, Weber, Mu{\~n}oz, Zhu, Strien, Alyafeai, Almubarak, Vu, Gonzalez-Dios, Soroa, Lo, Dey, Suarez, Gokaslan, Bose, Adelani, Phan, Tran, Yu, Pai, Chim, Lepercq, Ilic, Mitchell, Luccioni, and Jernite]{laurencon2023bigsciencerootscorpus16tb}
Hugo Lauren{\c c}on, Lucile Saulnier, Thomas Wang, Christopher Akiki, Albert~Villanova del Moral, Teven~Le Scao, Leandro~Von Werra, Chenghao Mou, Eduardo~Gonz{\'a}lez Ponferrada, Huu Nguyen, J{\"o}rg Frohberg, Mario {\v S}a{\v s}ko, Quentin Lhoest, Angelina McMillan-Major, Gerard Dupont, Stella Biderman, Anna Rogers, Loubna~Ben Allal, Francesco~De Toni, Giada Pistilli, Olivier Nguyen, Somaieh Nikpoor, Maraim Masoud, Pierre Colombo, Javier de~la Rosa, Paulo Villegas, Tristan Thrush, Shayne Longpre, Sebastian Nagel, Leon Weber, Manuel Mu{\~n}oz, Jian Zhu, Daniel~Van Strien, Zaid Alyafeai, Khalid Almubarak, Minh~Chien Vu, Itziar Gonzalez-Dios, Aitor Soroa, Kyle Lo, Manan Dey, Pedro~Ortiz Suarez, Aaron Gokaslan, Shamik Bose, David Adelani, Long Phan, Hieu Tran, Ian Yu, Suhas Pai, Jenny Chim, Violette Lepercq, Suzana Ilic, Margaret Mitchell, Sasha~Alexandra Luccioni, and Yacine Jernite.
\newblock The bigscience roots corpus: A 1.6tb composite multilingual dataset, 2023.
\newblock URL \url{https://arxiv.org/abs/2303.03915}.

\bibitem[Laurençon et~al.(2023)Laurençon, Saulnier, Wang, Akiki, del Moral, Scao, Werra, Mou, Ponferrada, Nguyen, Frohberg, Šaško, Lhoest, McMillan-Major, Dupont, Biderman, Rogers, allal, Toni, Pistilli, Nguyen, Nikpoor, Masoud, Colombo, de~la Rosa, Villegas, Thrush, Longpre, Nagel, Weber, Muñoz, Zhu, Strien, Alyafeai, Almubarak, Vu, Gonzalez-Dios, Soroa, Lo, Dey, Suarez, Gokaslan, Bose, Adelani, Phan, Tran, Yu, Pai, Chim, Lepercq, Ilic, Mitchell, Luccioni, and Jernite]{laurençon2023bigsciencerootscorpus16tb}
Hugo Laurençon, Lucile Saulnier, Thomas Wang, Christopher Akiki, Albert~Villanova del Moral, Teven~Le Scao, Leandro~Von Werra, Chenghao Mou, Eduardo~González Ponferrada, Huu Nguyen, Jörg Frohberg, Mario Šaško, Quentin Lhoest, Angelina McMillan-Major, Gerard Dupont, Stella Biderman, Anna Rogers, Loubna~Ben allal, Francesco~De Toni, Giada Pistilli, Olivier Nguyen, Somaieh Nikpoor, Maraim Masoud, Pierre Colombo, Javier de~la Rosa, Paulo Villegas, Tristan Thrush, Shayne Longpre, Sebastian Nagel, Leon Weber, Manuel Muñoz, Jian Zhu, Daniel~Van Strien, Zaid Alyafeai, Khalid Almubarak, Minh~Chien Vu, Itziar Gonzalez-Dios, Aitor Soroa, Kyle Lo, Manan Dey, Pedro~Ortiz Suarez, Aaron Gokaslan, Shamik Bose, David Adelani, Long Phan, Hieu Tran, Ian Yu, Suhas Pai, Jenny Chim, Violette Lepercq, Suzana Ilic, Margaret Mitchell, Sasha~Alexandra Luccioni, and Yacine Jernite.
\newblock The bigscience roots corpus: A 1.6tb composite multilingual dataset, 2023.
\newblock URL \url{https://arxiv.org/abs/2303.03915}.

\bibitem[Leviathan et~al.(2023)Leviathan, Kalman, and Matias]{specdec}
Yaniv Leviathan, Matan Kalman, and Yossi Matias.
\newblock Fast inference from transformers via speculative decoding.
\newblock In \emph{Proceedings of the 40th International Conference on Machine Learning}, ICML'23. JMLR.org, 2023.

\bibitem[Li et~al.(2024)Li, Fang, Smyrnis, Ivgi, Jordan, Gadre, Bansal, Guha, Keh, Arora, Garg, Xin, Muennighoff, Heckel, Mercat, Chen, Gururangan, Wortsman, Albalak, Bitton, Nezhurina, Abbas, Hsieh, Ghosh, Gardner, Kilian, Zhang, Shao, Pratt, Sanyal, Ilharco, Daras, Marathe, Gokaslan, Zhang, Chandu, Nguyen, Vasiljevic, Kakade, Song, Sanghavi, Faghri, Oh, Zettlemoyer, Lo, El-Nouby, Pouransari, Toshev, Wang, Groeneveld, Soldaini, Koh, Jitsev, Kollar, Dimakis, Carmon, Dave, Schmidt, and Shankar]{li2024datacomplm}
Jeffrey Li, Alex Fang, Georgios Smyrnis, Maor Ivgi, Matt Jordan, Samir Gadre, Hritik Bansal, Etash Guha, Sedrick Keh, Kushal Arora, Saurabh Garg, Rui Xin, Niklas Muennighoff, Reinhard Heckel, Jean Mercat, Mayee Chen, Suchin Gururangan, Mitchell Wortsman, Alon Albalak, Yonatan Bitton, Marianna Nezhurina, Amro Abbas, Cheng-Yu Hsieh, Dhruba Ghosh, Josh Gardner, Maciej Kilian, Hanlin Zhang, Rulin Shao, Sarah Pratt, Sunny Sanyal, Gabriel Ilharco, Giannis Daras, Kalyani Marathe, Aaron Gokaslan, Jieyu Zhang, Khyathi Chandu, Thao Nguyen, Igor Vasiljevic, Sham Kakade, Shuran Song, Sujay Sanghavi, Fartash Faghri, Sewoong Oh, Luke Zettlemoyer, Kyle Lo, Alaaeldin El-Nouby, Hadi Pouransari, Alexander Toshev, Stephanie Wang, Dirk Groeneveld, Luca Soldaini, Pang~Wei Koh, Jenia Jitsev, Thomas Kollar, Alexandros~G. Dimakis, Yair Carmon, Achal Dave, Ludwig Schmidt, and Vaishaal Shankar.
\newblock Datacomp-lm: In search of the next generation of training sets for language models.
\newblock \emph{arXiv preprint arXiv:2406.11794}, 2024.

\bibitem[Liu et~al.(2024)Liu, Min, Zettlemoyer, Choi, and Hajishirzi]{Liu2024InfiniGram}
Jiacheng Liu, Sewon Min, Luke Zettlemoyer, Yejin Choi, and Hannaneh Hajishirzi.
\newblock Infini-gram: Scaling unbounded n-gram language models to a trillion tokens.
\newblock \emph{arXiv preprint arXiv:2401.17377}, 2024.

\bibitem[Liu et~al.(2023)Liu, Qiao, Neiswanger, Wang, Tan, Tao, Li, Wang, Sun, Pangarkar, et~al.]{liu2023llm360}
Zhengzhong Liu, Aurick Qiao, Willie Neiswanger, Hongyi Wang, Bowen Tan, Tianhua Tao, Junbo Li, Yuqi Wang, Suqi Sun, Omkar Pangarkar, et~al.
\newblock Llm360: Towards fully transparent open-source llms.
\newblock \emph{arXiv preprint arXiv:2312.06550}, 2023.

\bibitem[{Llama Team}(2024)]{dubeyLlamaHerdModels2024}
AI~@~Meta {Llama Team}.
\newblock The llama 3 herd of models, 2024.
\newblock URL \url{https://arxiv.org/abs/2407.21783}.

\bibitem[Loshchilov and Hutter(2019)]{loshchilov2019decoupledweightdecayregularization}
Ilya Loshchilov and Frank Hutter.
\newblock Decoupled weight decay regularization, 2019.
\newblock URL \url{https://arxiv.org/abs/1711.05101}.

\bibitem[Lozhkov et~al.(2024)Lozhkov, Li, Allal, Cassano, Lamy-Poirier, Tazi, Tang, Pykhtar, Liu, Wei, Liu, Tian, Kocetkov, Zucker, Belkada, Wang, Liu, Abulkhanov, Paul, Li, Li, Risdal, Li, Zhu, Zhuo, Zheltonozhskii, Dade, Yu, Krau{\ss}, Jain, Su, He, Dey, Abati, Chai, Muennighoff, Tang, Oblokulov, Akiki, Marone, Mou, Mishra, Gu, Hui, Dao, Zebaze, Dehaene, Patry, Xu, McAuley, Hu, Scholak, Paquet, Robinson, Anderson, Chapados, Patwary, Tajbakhsh, Jernite, Ferrandis, Zhang, Hughes, Wolf, Guha, von Werra, and de~Vries]{lozhkov2024starcoder}
Anton Lozhkov, Raymond Li, Loubna~Ben Allal, Federico Cassano, Joel Lamy-Poirier, Nouamane Tazi, Ao~Tang, Dmytro Pykhtar, Jiawei Liu, Yuxiang Wei, Tianyang Liu, Max Tian, Denis Kocetkov, Arthur Zucker, Younes Belkada, Zijian Wang, Qian Liu, Dmitry Abulkhanov, Indraneil Paul, Zhuang Li, Wen-Ding Li, Megan Risdal, Jia Li, Jian Zhu, Terry~Yue Zhuo, Evgenii Zheltonozhskii, Nii Osae~Osae Dade, Wenhao Yu, Lucas Krau{\ss}, Naman Jain, Yixuan Su, Xuanli He, Manan Dey, Edoardo Abati, Yekun Chai, Niklas Muennighoff, Xiangru Tang, Muhtasham Oblokulov, Christopher Akiki, Marc Marone, Chenghao Mou, Mayank Mishra, Alex Gu, Binyuan Hui, Tri Dao, Armel Zebaze, Olivier Dehaene, Nicolas Patry, Canwen Xu, Julian McAuley, Han Hu, Torsten Scholak, Sebastien Paquet, Jennifer Robinson, Carolyn~Jane Anderson, Nicolas Chapados, Mostofa Patwary, Nima Tajbakhsh, Yacine Jernite, Carlos~Mu{\~n}oz Ferrandis, Lingming Zhang, Sean Hughes, Thomas Wolf, Arjun Guha, Leandro von Werra, and Harm de~Vries.
\newblock Starcoder 2 and the stack v2: The next generation, 2024.

\bibitem[Martins et~al.(2024)Martins, Fernandes, Alves, Guerreiro, Rei, Alves, Pombal, Farajian, Faysse, Klimaszewski, Colombo, Haddow, de~Souza, Birch, and Martins]{martins2024eurollmmultilinguallanguagemodels}
Pedro~Henrique Martins, Patrick Fernandes, Jo{\~a}o Alves, Nuno~M. Guerreiro, Ricardo Rei, Duarte~M. Alves, Jos{\'e} Pombal, Amin Farajian, Manuel Faysse, Mateusz Klimaszewski, Pierre Colombo, Barry Haddow, Jos{\'e} G.~C. de~Souza, Alexandra Birch, and Andr{\'e} F.~T. Martins.
\newblock Eurollm: Multilingual language models for europe, 2024.
\newblock URL \url{https://arxiv.org/abs/2409.16235}.

\bibitem[Martins et~al.(2025)Martins, Alves, Fernandes, Guerreiro, Rei, Farajian, Klimaszewski, Alves, Pombal, Boizard, Faysse, Colombo, Yvon, Haddow, de~Souza, Birch, and Martins]{martins2025eurollm9btechnicalreport}
Pedro~Henrique Martins, Jo{\~a}o Alves, Patrick Fernandes, Nuno~M. Guerreiro, Ricardo Rei, Amin Farajian, Mateusz Klimaszewski, Duarte~M. Alves, Jos{\'e} Pombal, Nicolas Boizard, Manuel Faysse, Pierre Colombo, Fran{\c c}ois Yvon, Barry Haddow, Jos{\'e} G.~C. de~Souza, Alexandra Birch, and Andr{\'e} F.~T. Martins.
\newblock Eurollm-9b: Technical report, 2025.
\newblock URL \url{https://arxiv.org/abs/2506.04079}.

\bibitem[Mikolov et~al.(2013)Mikolov, Chen, Corrado, and Dean]{mikolov-et-al:2013:wordvectors}
Tomas Mikolov, Kai Chen, Greg Corrado, and Jeffrey Dean.
\newblock Efficient estimation of word representations in vector space, 2013.

\bibitem[{MosaicML NLP Team}(2023)]{MosaicML2023Introducing:2023:mpt}
{MosaicML NLP Team}.
\newblock Introducing mpt-7b: A new standard for open-source, commercially usable llms, 2023.
\newblock URL \url{www.mosaicml.com/blog/mpt-7b}.
\newblock Accessed: 2023-05-05.

\bibitem[{OpenAI}(2022)]{openai_chatgpt:2022}
{OpenAI}.
\newblock {ChatGPT: Optimizing Language Models for Dialogue}, 2022.
\newblock URL \url{https://openai.com/blog/chatgpt/}.
\newblock Official announcement of the ChatGPT product.

\bibitem[{Ortiz Su\'arez} et~al.(2019){Ortiz Su\'arez}, Sagot, and Romary]{OrtizSuarezSagotRomary2019}
Pedro~Javier {Ortiz Su\'arez}, Beno\^it Sagot, and Laurent Romary.
\newblock Asynchronous pipelines for processing huge corpora on medium to low resource infrastructures.
\newblock Proceedings of the Workshop on Challenges in the Management of Large Corpora (CMLC-7) 2019. Cardiff, 22nd July 2019, pages 9 -- 16, Mannheim, 2019. Leibniz-Institut f{"u}r Deutsche Sprache.
\newblock \doi{10.14618/ids-pub-9021}.
\newblock URL \url{http://nbn-resolving.de/urn:nbn:de:bsz:mh39-90215}.

\bibitem[Paperno et~al.(2016)Paperno, Kruszewski, Lazaridou, Pham, Bernardi, Pezzelle, Baroni, Boleda, and Fern{\'a}ndez]{paperno-etal-2016-lambada}
Denis Paperno, Germ{\'a}n Kruszewski, Angeliki Lazaridou, Ngoc~Quan Pham, Raffaella Bernardi, Sandro Pezzelle, Marco Baroni, Gemma Boleda, and Raquel Fern{\'a}ndez.
\newblock The {LAMBADA} dataset: Word prediction requiring a broad discourse context.
\newblock In Katrin Erk and Noah~A. Smith, editors, \emph{Proceedings of the 54th Annual Meeting of the Association for Computational Linguistics (Volume 1: Long Papers)}, pages 1525--1534, Berlin, Germany, August 2016. Association for Computational Linguistics.
\newblock \doi{10.18653/v1/P16-1144}.
\newblock URL \url{https://aclanthology.org/P16-1144/}.

\bibitem[Parmar et~al.(2024)Parmar, Prabhumoye, Jennings, Patwary, Subramanian, Su, Zhu, Narayanan, Jhunjhunwala, Dattagupta, Jawa, Liu, Mahabaleshwarkar, Nitski, Brundyn, Maki, Martinez, You, Kamalu, LeGresley, Fridman, Casper, Aithal, Kuchaiev, Shoeybi, Cohen, and Catanzaro]{parmarNemotron415BTechnical2024}
Jupinder Parmar, Shrimai Prabhumoye, Joseph Jennings, Mostofa Patwary, Sandeep Subramanian, Dan Su, Chen Zhu, Deepak Narayanan, Aastha Jhunjhunwala, Ayush Dattagupta, Vibhu Jawa, Jiwei Liu, Ameya Mahabaleshwarkar, Osvald Nitski, Annika Brundyn, James Maki, Miguel Martinez, Jiaxuan You, John Kamalu, Patrick LeGresley, Denys Fridman, Jared Casper, Ashwath Aithal, Oleksii Kuchaiev, Mohammad Shoeybi, Jonathan Cohen, and Bryan Catanzaro.
\newblock Nemotron-4 {{15B Technical Report}}, 2024.
\newblock URL \url{http://arxiv.org/abs/2402.16819}.

\bibitem[Paster et~al.(2023)Paster, Santos, Azerbayev, and Ba]{paster2023openwebmath}
Keiran Paster, Marco~Dos Santos, Zhangir Azerbayev, and Jimmy Ba.
\newblock Openwebmath: An open dataset of high-quality mathematical web text, 2023.

\bibitem[Penedo et~al.(2024{\natexlab{a}})Penedo, Kydl{\'\i}{\v c}ek, {allal}, Lozhkov, Mitchell, Raffel, Von~Werra, and Wolf]{FineWebDecantingWeb}
Guilherme Penedo, Hynek Kydl{\'\i}{\v c}ek, Loubna~Ben {allal}, Anton Lozhkov, Margaret Mitchell, Colin Raffel, Leandro Von~Werra, and Thomas Wolf.
\newblock The {{FineWeb Datasets}}: {{Decanting}} the {{Web}} for the {{Finest Text Data}} at {{Scale}}, 2024{\natexlab{a}}.
\newblock URL \url{http://arxiv.org/abs/2406.17557}.

\bibitem[Penedo et~al.(2024{\natexlab{b}})Penedo, Kydl{\'\i}{\v c}ek, Cappelli, Sasko, and Wolf]{penedo2024datatrove}
Guilherme Penedo, Hynek Kydl{\'\i}{\v c}ek, Alessandro Cappelli, Mario Sasko, and Thomas Wolf.
\newblock Datatrove: large scale data processing, 2024{\natexlab{b}}.
\newblock URL \url{https://github.com/huggingface/datatrove}.

\bibitem[Penedo et~al.(2025)Penedo, Kydl{\'\i}{\v c}ek, Sabol{\v c}ec, Messmer, Foroutan, Kargaran, Raffel, Jaggi, Werra, and Wolf]{penedo2025fineweb2pipelinescale}
Guilherme Penedo, Hynek Kydl{\'\i}{\v c}ek, Vinko Sabol{\v c}ec, Bettina Messmer, Negar Foroutan, Amir~Hossein Kargaran, Colin Raffel, Martin Jaggi, Leandro~Von Werra, and Thomas Wolf.
\newblock Fineweb2: One pipeline to scale them all -- adapting pre-training data processing to every language, 2025.
\newblock URL \url{https://arxiv.org/abs/2506.20920}.

\bibitem[Peters et~al.(2018)Peters, Neumann, Iyyer, Gardner, Clark, Lee, and Zettlemoyer]{peters-etal-2018-deep}
Matthew~E. Peters, Mark Neumann, Mohit Iyyer, Matt Gardner, Christopher Clark, Kenton Lee, and Luke Zettlemoyer.
\newblock Deep contextualized word representations.
\newblock In Marilyn Walker, Heng Ji, and Amanda Stent, editors, \emph{Proceedings of the 2018 Conference of the North {A}merican Chapter of the Association for Computational Linguistics: Human Language Technologies, Volume 1 (Long Papers)}, pages 2227--2237, New Orleans, Louisiana, June 2018. Association for Computational Linguistics.
\newblock \doi{10.18653/v1/N18-1202}.
\newblock URL \url{https://aclanthology.org/N18-1202/}.

\bibitem[Pillutla et~al.(2021)Pillutla, Swayamdipta, Zellers, Thickstun, Welleck, Choi, and Harchaoui]{pillutla-etal:mauve:neurips2021}
Krishna Pillutla, Swabha Swayamdipta, Rowan Zellers, John Thickstun, Sean Welleck, Yejin Choi, and Zaid Harchaoui.
\newblock Mauve: Measuring the gap between neural text and human text using divergence frontiers.
\newblock In \emph{NeurIPS}, 2021.

\bibitem[{Qwen Team}(2025)]{qwen3technicalreport}
{Qwen Team}.
\newblock Qwen3 technical report, 2025.
\newblock URL \url{https://arxiv.org/abs/2505.09388}.

\bibitem[{Qwen Team} et~al.(2025){Qwen Team}, Yang, Yang, Zhang, Hui, Zheng, Yu, Li, Liu, Huang, Wei, Lin, Yang, Tu, Zhang, Yang, Yang, Zhou, Lin, Dang, Lu, Bao, Yang, Yu, Li, Xue, Zhang, Zhu, Men, Lin, Li, Tang, Xia, Ren, Ren, Fan, Su, Zhang, Wan, Liu, Cui, Zhang, and Qiu]{qwen2025qwen25technicalreport}
{Qwen Team}, An~Yang, Baosong Yang, Beichen Zhang, Binyuan Hui, Bo~Zheng, Bowen Yu, Chengyuan Li, Dayiheng Liu, Fei Huang, Haoran Wei, Huan Lin, Jian Yang, Jianhong Tu, Jianwei Zhang, Jianxin Yang, Jiaxi Yang, Jingren Zhou, Junyang Lin, Kai Dang, Keming Lu, Keqin Bao, Kexin Yang, Le~Yu, Mei Li, Mingfeng Xue, Pei Zhang, Qin Zhu, Rui Men, Runji Lin, Tianhao Li, Tianyi Tang, Tingyu Xia, Xingzhang Ren, Xuancheng Ren, Yang Fan, Yang Su, Yichang Zhang, Yu~Wan, Yuqiong Liu, Zeyu Cui, Zhenru Zhang, and Zihan Qiu.
\newblock Qwen2.5 technical report, 2025.
\newblock URL \url{https://arxiv.org/abs/2412.15115}.

\bibitem[Radford et~al.(2019)Radford, Wu, Child, Luan, Amodei, and Sutskever]{radford2019language}
Alec Radford, Jeff Wu, Rewon Child, David Luan, Dario Amodei, and Ilya Sutskever.
\newblock Language models are unsupervised multitask learners.
\newblock 2019.

\bibitem[Rae et~al.(2019)Rae, Potapenko, Jayakumar, Hillier, and Lillicrap]{raecompressive2019}
Jack~W Rae, Anna Potapenko, Siddhant~M Jayakumar, Chloe Hillier, and Timothy~P Lillicrap.
\newblock Compressive transformers for long-range sequence modelling.
\newblock \emph{arXiv preprint}, 2019.
\newblock URL \url{https://arxiv.org/abs/1911.05507}.

\bibitem[Rafailov et~al.(2023)Rafailov, Sharma, Mitchell, Ermon, Manning, and Finn]{rafailov-etal:2023:dpo}
Rafael Rafailov, Archit Sharma, Eric Mitchell, Stefano Ermon, Christopher~D. Manning, and Chelsea Finn.
\newblock Direct preference optimization: Your language model is secretly a reward model, 2023.

\bibitem[Rajbhandari et~al.(2020)Rajbhandari, Rasley, Ruwase, and He]{rajbhandari2020zeromemoryoptimizationstraining}
Samyam Rajbhandari, Jeff Rasley, Olatunji Ruwase, and Yuxiong He.
\newblock Zero: Memory optimizations toward training trillion parameter models, 2020.
\newblock URL \url{https://arxiv.org/abs/1910.02054}.

\bibitem[Rajbhandari et~al.(2021)Rajbhandari, Ruwase, Rasley, Smith, and He]{rajbhandari2021zeroinfinitybreakinggpumemory}
Samyam Rajbhandari, Olatunji Ruwase, Jeff Rasley, Shaden Smith, and Yuxiong He.
\newblock Zero-infinity: Breaking the gpu memory wall for extreme scale deep learning, 2021.
\newblock URL \url{https://arxiv.org/abs/2104.07857}.

\bibitem[Rein et~al.(2023)Rein, Hou, Stickland, Petty, Pang, Dirani, Michael, and Bowman]{rein2023gpqagraduatelevelgoogleproofqa}
David Rein, Betty~Li Hou, Asa~Cooper Stickland, Jackson Petty, Richard~Yuanzhe Pang, Julien Dirani, Julian Michael, and Samuel~R. Bowman.
\newblock Gpqa: A graduate-level google-proof q\&a benchmark, 2023.
\newblock URL \url{https://arxiv.org/abs/2311.12022}.

\bibitem[Sap et~al.(2019)Sap, Rashkin, Chen, Le~Bras, and Choi]{sap-etal-2019-social}
Maarten Sap, Hannah Rashkin, Derek Chen, Ronan Le~Bras, and Yejin Choi.
\newblock Social {IQ}a: Commonsense reasoning about social interactions.
\newblock In Kentaro Inui, Jing Jiang, Vincent Ng, and Xiaojun Wan, editors, \emph{Proceedings of the 2019 Conference on Empirical Methods in Natural Language Processing and the 9th International Joint Conference on Natural Language Processing (EMNLP-IJCNLP)}, pages 4463--4473, Hong Kong, China, November 2019. Association for Computational Linguistics.
\newblock \doi{10.18653/v1/D19-1454}.
\newblock URL \url{https://aclanthology.org/D19-1454/}.

\bibitem[Saxton and Hill(2019)]{2019arXivdmamths}
Grefenstette Saxton and Kohli Hill.
\newblock Analysing mathematical reasoning abilities of neural models.
\newblock \emph{arXiv:1904.01557}, 2019.

\bibitem[Scao et~al.(2022)Scao, Fan, Akiki, Pavlick, Ili{\'c}, Hesslow, Castagn{\'e}, Luccioni, Yvon, Gall{\'e}, Tow, Rush, Biderman, Webson, Ammanamanchi, Wang, Sagot, Muennighoff, del Moral, Ruwase, Bawden, Bekman, McMillan-Major, Beltagy, Nguyen, Saulnier, Tan, Suarez, Sanh, Lauren{\c c}on, Jernite, Launay, Mitchell, Raffel, Gokaslan, Simhi, Soroa, Aji, Alfassy, Rogers, Nitzav, Xu, Mou, Emezue, Klamm, Leong, van Strien, Adelani, Radev, Ponferrada, Levkovizh, Kim, Natan, Toni, Dupont, Kruszewski, Pistilli, Elsahar, Benyamina, Tran, Yu, Abdulmumin, Johnson, Gonzalez-Dios, de~la Rosa, Chim, Dodge, Zhu, Chang, Frohberg, Tobing, Bhattacharjee, Almubarak, Chen, Lo, Werra, Weber, Phan, allal, Tanguy, Dey, Mu{\~n}oz, Masoud, Grandury, {\v S}a{\v s}ko, Huang, Coavoux, Singh, Jiang, Vu, Jauhar, Ghaleb, Subramani, Kassner, Khamis, Nguyen, Espejel, de~Gibert, Villegas, Henderson, Colombo, Amuok, Lhoest, Harliman, Bommasani, L{\'o}pez, Ribeiro, Osei, Pyysalo, Nagel, Bose, Muhammad, Sharma, Longpre, Nikpoor, Silberberg,
  Pai, Zink, Torrent, Schick, Thrush, Danchev, Nikoulina, Laippala, Lepercq, Prabhu, Alyafeai, Talat, Raja, Heinzerling, Si, Ta{\c s}ar, Salesky, Mielke, Lee, Sharma, Santilli, Chaffin, Stiegler, Datta, Szczechla, Chhablani, Wang, Pandey, Strobelt, Fries, Rozen, Gao, Sutawika, Bari, Al-shaibani, Manica, Nayak, Teehan, Albanie, Shen, Ben-David, Bach, Kim, Bers, Fevry, Neeraj, Thakker, Raunak, Tang, Yong, Sun, Brody, Uri, Tojarieh, Roberts, Chung, Tae, Phang, Press, Li, Narayanan, Bourfoune, Casper, Rasley, Ryabinin, Mishra, Zhang, Shoeybi, Peyrounette, Patry, Tazi, Sanseviero, von Platen, Cornette, Lavall{\'e}e, Lacroix, Rajbhandari, Gandhi, Smith, Requena, Patil, Dettmers, Baruwa, Singh, Cheveleva, Ligozat, Subramonian, N{\'e}v{\'e}ol, Lovering, Garrette, Tunuguntla, Reiter, Taktasheva, Voloshina, Bogdanov, Winata, Schoelkopf, Kalo, Novikova, Forde, Clive, Kasai, Kawamura, Hazan, Carpuat, Clinciu, Kim, Cheng, Serikov, Antverg, van~der Wal, Zhang, Zhang, Gehrmann, Mirkin, Pais, Shavrina, Scialom, Yun,
  Limisiewicz, Rieser, Protasov, Mikhailov, Pruksachatkun, Belinkov, Bamberger, Kasner, Rueda, Pestana, Feizpour, Khan, Faranak, Santos, Hevia, Unldreaj, Aghagol, Abdollahi, Tammour, HajiHosseini, Behroozi, Ajibade, Saxena, Ferrandis, McDuff, Contractor, Lansky, David, Kiela, Nguyen, Tan, Baylor, Ozoani, Mirza, Ononiwu, Rezanejad, Jones, Bhattacharya, Solaiman, Sedenko, Nejadgholi, Passmore, Seltzer, Sanz, Dutra, Samagaio, Elbadri, Mieskes, Gerchick, Akinlolu, McKenna, Qiu, Ghauri, Burynok, Abrar, Rajani, Elkott, Fahmy, Samuel, An, Kromann, Hao, Alizadeh, Shubber, Wang, Roy, Viguier, Le, Oyebade, Le, Yang, Nguyen, Kashyap, Palasciano, Callahan, Shukla, Miranda-Escalada, Singh, Beilharz, Wang, Brito, Zhou, Jain, Xu, Fourrier, Peri{\~n}{\'a}n, Molano, Yu, Manjavacas, Barth, Fuhrimann, Altay, Bayrak, Burns, Vrabec, Bello, Dash, Kang, Giorgi, Golde, Posada, Sivaraman, Bulchandani, Liu, Shinzato, de~Bykhovetz, Takeuchi, P{\`a}mies, Castillo, Nezhurina, S{\"a}nger, Samwald, Cullan, Weinberg, Wolf, Mihaljcic, Liu,
  Freidank, Kang, Seelam, Dahlberg, Broad, Muellner, Fung, Haller, Chandrasekhar, Eisenberg, Martin, Canalli, Su, Su, Cahyawijaya, Garda, Deshmukh, Mishra, Kiblawi, Ott, Sang-aroonsiri, Kumar, Schweter, Bharati, Laud, Gigant, Kainuma, Kusa, Labrak, Bajaj, Venkatraman, Xu, Xu, Xu, Tan, Xie, Ye, Bras, Belkada, and Wolf]{Scao-et-al:2022:Bloom}
Teven~Le Scao, Angela Fan, Christopher Akiki, Ellie Pavlick, Suzana Ili{\'c}, Daniel Hesslow, Roman Castagn{\'e}, Alexandra~Sasha Luccioni, Fran{\c c}ois Yvon, Matthias Gall{\'e}, Jonathan Tow, Alexander~M. Rush, Stella Biderman, Albert Webson, Pawan~Sasanka Ammanamanchi, Thomas Wang, Beno{\^\i}t Sagot, Niklas Muennighoff, Albert~Villanova del Moral, Olatunji Ruwase, Rachel Bawden, Stas Bekman, Angelina McMillan-Major, Iz~Beltagy, Huu Nguyen, Lucile Saulnier, Samson Tan, Pedro~Ortiz Suarez, Victor Sanh, Hugo Lauren{\c c}on, Yacine Jernite, Julien Launay, Margaret Mitchell, Colin Raffel, Aaron Gokaslan, Adi Simhi, Aitor Soroa, Alham~Fikri Aji, Amit Alfassy, Anna Rogers, Ariel~Kreisberg Nitzav, Canwen Xu, Chenghao Mou, Chris Emezue, Christopher Klamm, Colin Leong, Daniel van Strien, David~Ifeoluwa Adelani, Dragomir Radev, Eduardo~Gonz{\'a}lez Ponferrada, Efrat Levkovizh, Ethan Kim, Eyal~Bar Natan, Francesco~De Toni, G{\'e}rard Dupont, Germ{\'a}n Kruszewski, Giada Pistilli, Hady Elsahar, Hamza Benyamina, Hieu
  Tran, Ian Yu, Idris Abdulmumin, Isaac Johnson, Itziar Gonzalez-Dios, Javier de~la Rosa, Jenny Chim, Jesse Dodge, Jian Zhu, Jonathan Chang, J{\"o}rg Frohberg, Joseph Tobing, Joydeep Bhattacharjee, Khalid Almubarak, Kimbo Chen, Kyle Lo, Leandro~Von Werra, Leon Weber, Long Phan, Loubna~Ben allal, Ludovic Tanguy, Manan Dey, Manuel~Romero Mu{\~n}oz, Maraim Masoud, Mar{\'\i}a Grandury, Mario {\v S}a{\v s}ko, Max Huang, Maximin Coavoux, Mayank Singh, Mike Tian-Jian Jiang, Minh~Chien Vu, Mohammad~A. Jauhar, Mustafa Ghaleb, Nishant Subramani, Nora Kassner, Nurulaqilla Khamis, Olivier Nguyen, Omar Espejel, Ona de~Gibert, Paulo Villegas, Peter Henderson, Pierre Colombo, Priscilla Amuok, Quentin Lhoest, Rheza Harliman, Rishi Bommasani, Roberto~Luis L{\'o}pez, Rui Ribeiro, Salomey Osei, Sampo Pyysalo, Sebastian Nagel, Shamik Bose, Shamsuddeen~Hassan Muhammad, Shanya Sharma, Shayne Longpre, Somaieh Nikpoor, Stanislav Silberberg, Suhas Pai, Sydney Zink, Tiago~Timponi Torrent, Timo Schick, Tristan Thrush, Valentin Danchev,
  Vassilina Nikoulina, Veronika Laippala, Violette Lepercq, Vrinda Prabhu, Zaid Alyafeai, Zeerak Talat, Arun Raja, Benjamin Heinzerling, Chenglei Si, Davut~Emre Ta{\c s}ar, Elizabeth Salesky, Sabrina~J. Mielke, Wilson~Y. Lee, Abheesht Sharma, Andrea Santilli, Antoine Chaffin, Arnaud Stiegler, Debajyoti Datta, Eliza Szczechla, Gunjan Chhablani, Han Wang, Harshit Pandey, Hendrik Strobelt, Jason~Alan Fries, Jos Rozen, Leo Gao, Lintang Sutawika, M~Saiful Bari, Maged~S. Al-shaibani, Matteo Manica, Nihal Nayak, Ryan Teehan, Samuel Albanie, Sheng Shen, Srulik Ben-David, Stephen~H. Bach, Taewoon Kim, Tali Bers, Thibault Fevry, Trishala Neeraj, Urmish Thakker, Vikas Raunak, Xiangru Tang, Zheng-Xin Yong, Zhiqing Sun, Shaked Brody, Yallow Uri, Hadar Tojarieh, Adam Roberts, Hyung~Won Chung, Jaesung Tae, Jason Phang, Ofir Press, Conglong Li, Deepak Narayanan, Hatim Bourfoune, Jared Casper, Jeff Rasley, Max Ryabinin, Mayank Mishra, Minjia Zhang, Mohammad Shoeybi, Myriam Peyrounette, Nicolas Patry, Nouamane Tazi, Omar
  Sanseviero, Patrick von Platen, Pierre Cornette, Pierre~Fran{\c c}ois Lavall{\'e}e, R{\'e}mi Lacroix, Samyam Rajbhandari, Sanchit Gandhi, Shaden Smith, St{\'e}phane Requena, Suraj Patil, Tim Dettmers, Ahmed Baruwa, Amanpreet Singh, Anastasia Cheveleva, Anne-Laure Ligozat, Arjun Subramonian, Aur{\'e}lie N{\'e}v{\'e}ol, Charles Lovering, Dan Garrette, Deepak Tunuguntla, Ehud Reiter, Ekaterina Taktasheva, Ekaterina Voloshina, Eli Bogdanov, Genta~Indra Winata, Hailey Schoelkopf, Jan-Christoph Kalo, Jekaterina Novikova, Jessica~Zosa Forde, Jordan Clive, Jungo Kasai, Ken Kawamura, Liam Hazan, Marine Carpuat, Miruna Clinciu, Najoung Kim, Newton Cheng, Oleg Serikov, Omer Antverg, Oskar van~der Wal, Rui Zhang, Ruochen Zhang, Sebastian Gehrmann, Shachar Mirkin, Shani Pais, Tatiana Shavrina, Thomas Scialom, Tian Yun, Tomasz Limisiewicz, Verena Rieser, Vitaly Protasov, Vladislav Mikhailov, Yada Pruksachatkun, Yonatan Belinkov, Zachary Bamberger, Zden{\v e}k Kasner, Alice Rueda, Amanda Pestana, Amir Feizpour, Ammar
  Khan, Amy Faranak, Ana Santos, Anthony Hevia, Antigona Unldreaj, Arash Aghagol, Arezoo Abdollahi, Aycha Tammour, Azadeh HajiHosseini, Bahareh Behroozi, Benjamin Ajibade, Bharat Saxena, Carlos~Mu{\~n}oz Ferrandis, Daniel McDuff, Danish Contractor, David Lansky, Davis David, Douwe Kiela, Duong~A. Nguyen, Edward Tan, Emi Baylor, Ezinwanne Ozoani, Fatima Mirza, Frankline Ononiwu, Habib Rezanejad, Hessie Jones, Indrani Bhattacharya, Irene Solaiman, Irina Sedenko, Isar Nejadgholi, Jesse Passmore, Josh Seltzer, Julio~Bonis Sanz, Livia Dutra, Mairon Samagaio, Maraim Elbadri, Margot Mieskes, Marissa Gerchick, Martha Akinlolu, Michael McKenna, Mike Qiu, Muhammed Ghauri, Mykola Burynok, Nafis Abrar, Nazneen Rajani, Nour Elkott, Nour Fahmy, Olanrewaju Samuel, Ran An, Rasmus Kromann, Ryan Hao, Samira Alizadeh, Sarmad Shubber, Silas Wang, Sourav Roy, Sylvain Viguier, Thanh Le, Tobi Oyebade, Trieu Le, Yoyo Yang, Zach Nguyen, Abhinav~Ramesh Kashyap, Alfredo Palasciano, Alison Callahan, Anima Shukla, Antonio
  Miranda-Escalada, Ayush Singh, Benjamin Beilharz, Bo~Wang, Caio Brito, Chenxi Zhou, Chirag Jain, Chuxin Xu, Cl{\'e}mentine Fourrier, Daniel~Le{\'o}n Peri{\~n}{\'a}n, Daniel Molano, Dian Yu, Enrique Manjavacas, Fabio Barth, Florian Fuhrimann, Gabriel Altay, Giyaseddin Bayrak, Gully Burns, Helena~U. Vrabec, Imane Bello, Ishani Dash, Jihyun Kang, John Giorgi, Jonas Golde, Jose~David Posada, Karthik~Rangasai Sivaraman, Lokesh Bulchandani, Lu~Liu, Luisa Shinzato, Madeleine~Hahn de~Bykhovetz, Maiko Takeuchi, Marc P{\`a}mies, Maria~A Castillo, Marianna Nezhurina, Mario S{\"a}nger, Matthias Samwald, Michael Cullan, Michael Weinberg, Michiel~De Wolf, Mina Mihaljcic, Minna Liu, Moritz Freidank, Myungsun Kang, Natasha Seelam, Nathan Dahlberg, Nicholas~Michio Broad, Nikolaus Muellner, Pascale Fung, Patrick Haller, Ramya Chandrasekhar, Renata Eisenberg, Robert Martin, Rodrigo Canalli, Rosaline Su, Ruisi Su, Samuel Cahyawijaya, Samuele Garda, Shlok~S Deshmukh, Shubhanshu Mishra, Sid Kiblawi, Simon Ott, Sinee
  Sang-aroonsiri, Srishti Kumar, Stefan Schweter, Sushil Bharati, Tanmay Laud, Th{\'e}o Gigant, Tomoya Kainuma, Wojciech Kusa, Yanis Labrak, Yash~Shailesh Bajaj, Yash Venkatraman, Yifan Xu, Yingxin Xu, Yu~Xu, Zhe Tan, Zhongli Xie, Zifan Ye, Mathilde Bras, Younes Belkada, and Thomas Wolf.
\newblock Bloom: A 176b-parameter open-access multilingual language model, 2022.

\bibitem[Sennrich et~al.(2016)Sennrich, Haddow, and Birch]{sennrich-etal-2016-neural}
Rico Sennrich, Barry Haddow, and Alexandra Birch.
\newblock Neural machine translation of rare words with subword units.
\newblock In Katrin Erk and Noah~A. Smith, editors, \emph{Proceedings of the 54th Annual Meeting of the Association for Computational Linguistics (Volume 1: Long Papers)}, pages 1715--1725, Berlin, Germany, August 2016. Association for Computational Linguistics.
\newblock \doi{10.18653/v1/P16-1162}.
\newblock URL \url{https://aclanthology.org/P16-1162/}.

\bibitem[Shah et~al.(2024)Shah, Bikshandi, Zhang, Thakkar, Ramani, and Dao]{dao2024flashattention3}
Jay Shah, Ganesh Bikshandi, Ying Zhang, Vijay Thakkar, Pradeep Ramani, and Tri Dao.
\newblock Flashattention-3: Fast and accurate attention with asynchrony and low-precision.
\newblock In A.~Globerson, L.~Mackey, D.~Belgrave, A.~Fan, U.~Paquet, J.~Tomczak, and C.~Zhang, editors, \emph{Advances in Neural Information Processing Systems}, volume~37, pages 68658--68685. Curran Associates, Inc., 2024.
\newblock URL \url{https://proceedings.neurips.cc/paper_files/paper/2024/file/7ede97c3e082c6df10a8d6103a2eebd2-Paper-Conference.pdf}.

\bibitem[Shao et~al.(2024)Shao, Wang, Zhu, Xu, Song, Bi, Zhang, Zhang, Li, Wu, and Guo]{shao2024deepseekmathpushinglimitsmathematical}
Zhihong Shao, Peiyi Wang, Qihao Zhu, Runxin Xu, Junxiao Song, Xiao Bi, Haowei Zhang, Mingchuan Zhang, Y.~K. Li, Y.~Wu, and Daya Guo.
\newblock Deepseekmath: Pushing the limits of mathematical reasoning in open language models, 2024.
\newblock URL \url{https://arxiv.org/abs/2402.03300}.

\bibitem[Simoulin and Crabb{\'e}(2021)]{simoulin:hal-03265900}
Antoine Simoulin and Benoit Crabb{\'e}.
\newblock {Un mod{\`e}le Transformer G{\'e}n{\'e}ratif Pr{\'e}-entrain{\'e} pour le \_\_\_\_\_\_ fran{\c c}ais}.
\newblock In Pascal Denis, Natalia Grabar, Amel Fraisse, R{\'e}mi Cardon, Bernard Jacquemin, Eric Kergosien, and Antonio Balvet, editors, \emph{{Traitement Automatique des Langues Naturelles}}, pages 246--255, Lille, France, 2021. {ATALA}.
\newblock URL \url{https://hal.archives-ouvertes.fr/hal-03265900}.

\bibitem[Soldaini et~al.(2024)Soldaini, Kinney, Bhagia, Schwenk, Atkinson, Authur, Bogin, Chandu, Dumas, Elazar, Hofmann, Jha, Kumar, Lucy, Lyu, Lambert, Magnusson, Morrison, Muennighoff, Naik, Nam, Peters, Ravichander, Richardson, Shen, Strubell, Subramani, Tafjord, Walsh, Zettlemoyer, Smith, Hajishirzi, Beltagy, Groeneveld, Dodge, and Lo]{soldaini-et-al:2024:dolma}
Luca Soldaini, Rodney Kinney, Akshita Bhagia, Dustin Schwenk, David Atkinson, Russell Authur, Ben Bogin, Khyathi Chandu, Jennifer Dumas, Yanai Elazar, Valentin Hofmann, Ananya~Harsh Jha, Sachin Kumar, Li~Lucy, Xinxi Lyu, Nathan Lambert, Ian Magnusson, Jacob Morrison, Niklas Muennighoff, Aakanksha Naik, Crystal Nam, Matthew~E. Peters, Abhilasha Ravichander, Kyle Richardson, Zejiang Shen, Emma Strubell, Nishant Subramani, Oyvind Tafjord, Pete Walsh, Luke Zettlemoyer, Noah~A. Smith, Hannaneh Hajishirzi, Iz~Beltagy, Dirk Groeneveld, Jesse Dodge, and Kyle Lo.
\newblock Dolma: an open corpus of three trillion tokens for language model pretraining research, 2024.

\bibitem[Souly et~al.(2025)Souly, Rando, Chapman, Davies, Hasircioglu, Shereen, Mougan, Mavroudis, Jones, Hicks, Carlini, Gal, and Kirk]{souly-et-al:2025:poisoning}
Alexandra Souly, Javier Rando, Ed~Chapman, Xander Davies, Burak Hasircioglu, Ezzeldin Shereen, Carlos Mougan, Vasilios Mavroudis, Erik Jones, Chris Hicks, Nicholas Carlini, Yarin Gal, and Robert Kirk.
\newblock Poisoning attacks on llms require a near-constant number of poison samples, 2025.

\bibitem[Talmor et~al.(2019)Talmor, Herzig, Lourie, and Berant]{talmor-etal-2019-commonsenseqa}
Alon Talmor, Jonathan Herzig, Nicholas Lourie, and Jonathan Berant.
\newblock {C}ommonsense{QA}: A question answering challenge targeting commonsense knowledge.
\newblock In \emph{Proceedings of the 2019 Conference of the North {A}merican Chapter of the Association for Computational Linguistics: Human Language Technologies, Volume 1 (Long and Short Papers)}, pages 4149--4158, Minneapolis, Minnesota, June 2019. Association for Computational Linguistics.
\newblock \doi{10.18653/v1/N19-1421}.
\newblock URL \url{https://aclanthology.org/N19-1421}.

\bibitem[Tang et~al.(2024)Tang, Ranjan, Pangarkar, Liang, Wang, An, Rao, Jin, Wang, Cheng, Sun, Mu, Miller, Ma, Peng, Liu, and Xing]{txt360data2024}
Liping Tang, Nikhil Ranjan, Omkar Pangarkar, Xuezhi Liang, Zhen Wang, Li~An, Bhaskar Rao, Linghao Jin, Huijuan Wang, Zhoujun Cheng, Suqi Sun, Cun Mu, Victor Miller, Xuezhe Ma, Yue Peng, Zhengzhong Liu, and Eric~P. Xing.
\newblock Txt360: A top-quality llm pre-training dataset requires the perfect blend, 2024.

\bibitem[{Team OLMo} et~al.(2025){Team OLMo}, Walsh, Soldaini, Groeneveld, Lo, Arora, Bhagia, Gu, Huang, Jordan, Lambert, Schwenk, Tafjord, Anderson, Atkinson, Brahman, Clark, Dasigi, Dziri, Guerquin, Ivison, Koh, Liu, Malik, Merrill, Miranda, Morrison, Murray, Nam, Pyatkin, Rangapur, Schmitz, Skjonsberg, Wadden, Wilhelm, Wilson, Zettlemoyer, Farhadi, Smith, and Hajishirzi]{olmo20252olmo2furious}
{Team OLMo}, Pete Walsh, Luca Soldaini, Dirk Groeneveld, Kyle Lo, Shane Arora, Akshita Bhagia, Yuling Gu, Shengyi Huang, Matt Jordan, Nathan Lambert, Dustin Schwenk, Oyvind Tafjord, Taira Anderson, David Atkinson, Faeze Brahman, Christopher Clark, Pradeep Dasigi, Nouha Dziri, Michal Guerquin, Hamish Ivison, Pang~Wei Koh, Jiacheng Liu, Saumya Malik, William Merrill, Lester James~V. Miranda, Jacob Morrison, Tyler Murray, Crystal Nam, Valentina Pyatkin, Aman Rangapur, Michael Schmitz, Sam Skjonsberg, David Wadden, Christopher Wilhelm, Michael Wilson, Luke Zettlemoyer, Ali Farhadi, Noah~A. Smith, and Hannaneh Hajishirzi.
\newblock 2 olmo 2 furious, 2025.
\newblock URL \url{https://arxiv.org/abs/2501.00656}.

\bibitem[Teknium(2023)]{OpenHermes2.5}
Teknium.
\newblock Openhermes 2.5: An open dataset of synthetic data for generalist llm assistants, 2023.
\newblock URL \url{https://huggingface.co/datasets/teknium/OpenHermes-2.5}.

\bibitem[Tiedemann(2012)]{tiedemann-2012-parallel}
J{\"o}rg Tiedemann.
\newblock Parallel data, tools and interfaces in {OPUS}.
\newblock In Nicoletta Calzolari, Khalid Choukri, Thierry Declerck, Mehmet~U{\u{g}}ur Do{\u{g}}an, Bente Maegaard, Joseph Mariani, Asuncion Moreno, Jan Odijk, and Stelios Piperidis, editors, \emph{Proceedings of the Eighth International Conference on Language Resources and Evaluation ({LREC}'12)}, pages 2214--2218, Istanbul, Turkey, May 2012. European Language Resources Association (ELRA).
\newblock URL \url{https://aclanthology.org/L12-1246/}.

\bibitem[Touvron et~al.(2023{\natexlab{a}})Touvron, Lavril, Izacard, Martinet, Lachaux, Lacroix, Rozière, Goyal, Hambro, Azhar, Rodriguez, Joulin, Grave, and Lample]{touvron-et-al:2023:llama1}
Hugo Touvron, Thibaut Lavril, Gautier Izacard, Xavier Martinet, Marie-Anne Lachaux, Timothée Lacroix, Baptiste Rozière, Naman Goyal, Eric Hambro, Faisal Azhar, Aurelien Rodriguez, Armand Joulin, Edouard Grave, and Guillaume Lample.
\newblock Llama: Open and efficient foundation language models, 2023{\natexlab{a}}.

\bibitem[Touvron et~al.(2023{\natexlab{b}})Touvron, Martin, Stone, Albert, Almahairi, Babaei, Bashlykov, Batra, Bhargava, Bhosale, Bikel, Blecher, Ferrer, Chen, Cucurull, Esiobu, Fernandes, Fu, Fu, Fuller, Gao, Goswami, Goyal, Hartshorn, Hosseini, Hou, Inan, Kardas, Kerkez, Khabsa, Kloumann, Korenev, Koura, Lachaux, Lavril, Lee, Liskovich, Lu, Mao, Martinet, Mihaylov, Mishra, Molybog, Nie, Poulton, Reizenstein, Rungta, Saladi, Schelten, Silva, Smith, Subramanian, Tan, Tang, Taylor, Williams, Kuan, Xu, Yan, Zarov, Zhang, Fan, Kambadur, Narang, Rodriguez, Stojnic, Edunov, and Scialom]{touvron2023llama2openfoundation}
Hugo Touvron, Louis Martin, Kevin Stone, Peter Albert, Amjad Almahairi, Yasmine Babaei, Nikolay Bashlykov, Soumya Batra, Prajjwal Bhargava, Shruti Bhosale, Dan Bikel, Lukas Blecher, Cristian~Canton Ferrer, Moya Chen, Guillem Cucurull, David Esiobu, Jude Fernandes, Jeremy Fu, Wenyin Fu, Brian Fuller, Cynthia Gao, Vedanuj Goswami, Naman Goyal, Anthony Hartshorn, Saghar Hosseini, Rui Hou, Hakan Inan, Marcin Kardas, Viktor Kerkez, Madian Khabsa, Isabel Kloumann, Artem Korenev, Punit~Singh Koura, Marie-Anne Lachaux, Thibaut Lavril, Jenya Lee, Diana Liskovich, Yinghai Lu, Yuning Mao, Xavier Martinet, Todor Mihaylov, Pushkar Mishra, Igor Molybog, Yixin Nie, Andrew Poulton, Jeremy Reizenstein, Rashi Rungta, Kalyan Saladi, Alan Schelten, Ruan Silva, Eric~Michael Smith, Ranjan Subramanian, Xiaoqing~Ellen Tan, Binh Tang, Ross Taylor, Adina Williams, Jian~Xiang Kuan, Puxin Xu, Zheng Yan, Iliyan Zarov, Yuchen Zhang, Angela Fan, Melanie Kambadur, Sharan Narang, Aurelien Rodriguez, Robert Stojnic, Sergey Edunov, and Thomas
  Scialom.
\newblock Llama 2: Open foundation and fine-tuned chat models, 2023{\natexlab{b}}.
\newblock URL \url{https://arxiv.org/abs/2307.09288}.

\bibitem[Vaswani et~al.(2017)Vaswani, Shazeer, Parmar, Uszkoreit, Jones, Gomez, Kaiser, and Polosukhin]{vaswani2017attention}
Ashish Vaswani, Noam Shazeer, Niki Parmar, Jakob Uszkoreit, Llion Jones, Aidan~N Gomez, {\L}ukasz Kaiser, and Illia Polosukhin.
\newblock Attention is all you need.
\newblock \emph{Advances in neural information processing systems}, 30, 2017.

\bibitem[Wan et~al.(2023)Wan, Wallace, Shen, and Klein]{wan2023poisoning}
Alexander Wan, Eric Wallace, Sheng Shen, and Dan Klein.
\newblock Poisoning language models during instruction tuning.
\newblock \emph{arXiv preprint arXiv:2305.00944}, 2023.

\bibitem[Weber et~al.(2024)Weber, Fu, Anthony, Oren, Adams, Alexandrov, Lyu, Nguyen, Yao, Adams, Athiwaratkun, Chalamala, Chen, Ryabinin, Dao, Liang, R{\'e}, Rish, and Zhang]{weber2024redpajama}
Maurice Weber, Daniel~Y. Fu, Quentin Anthony, Yonatan Oren, Shane Adams, Anton Alexandrov, Xiaozhong Lyu, Huu Nguyen, Xiaozhe Yao, Virginia Adams, Ben Athiwaratkun, Rahul Chalamala, Kezhen Chen, Max Ryabinin, Tri Dao, Percy Liang, Christopher R{\'e}, Irina Rish, and Ce~Zhang.
\newblock Redpajama: an open dataset for training large language models.
\newblock \emph{NeurIPS Datasets and Benchmarks Track}, 2024.

\bibitem[Wei et~al.(2025)Wei, Godbole, Khan, Wang, Zhu, Flemings, Kashyap, Gummadi, Neiswanger, and Jia]{wei2025hubblemodelsuiteadvance}
Johnny Tian-Zheng Wei, Ameya Godbole, Mohammad~Aflah Khan, Ryan Wang, Xiaoyuan Zhu, James Flemings, Nitya Kashyap, Krishna~P. Gummadi, Willie Neiswanger, and Robin Jia.
\newblock Hubble: a model suite to advance the study of llm memorization, 2025.
\newblock URL \url{https://arxiv.org/abs/2510.19811}.

\bibitem[Wettig et~al.(2025)Wettig, Lo, Min, Hajishirzi, Chen, and Soldaini]{wettig2025organize}
Alexander Wettig, Kyle Lo, Sewon Min, Hannaneh Hajishirzi, Danqi Chen, and Luca Soldaini.
\newblock Organize the web: Constructing domains enhances pre-training data curation.
\newblock \emph{arXiv preprint arXiv:2502.10341}, 2025.

\bibitem[Xu et~al.(2024)Xu, Guan, Greene, and Kechadi]{xu2024benchmarkdatacontaminationlarge}
Cheng Xu, Shuhao Guan, Derek Greene, and M-Tahar Kechadi.
\newblock Benchmark data contamination of large language models: A survey, 2024.
\newblock URL \url{https://arxiv.org/abs/2406.04244}.

\bibitem[Yang et~al.(2024)Yang, Yang, Hui, Zheng, Yu, Zhou, Li, Li, Liu, Huang, Dong, Wei, Lin, Tang, Wang, Yang, Tu, Zhang, Ma, Yang, Xu, Zhou, Bai, He, Lin, Dang, Lu, Chen, Yang, Li, Xue, Ni, Zhang, Wang, Peng, Men, Gao, Lin, Wang, Bai, Tan, Zhu, Li, Liu, Ge, Deng, Zhou, Ren, Zhang, Wei, Ren, Liu, Fan, Yao, Zhang, Wan, Chu, Liu, Cui, Zhang, Guo, and Fan]{yang2024qwen2technicalreport}
An~Yang, Baosong Yang, Binyuan Hui, Bo~Zheng, Bowen Yu, Chang Zhou, Chengpeng Li, Chengyuan Li, Dayiheng Liu, Fei Huang, Guanting Dong, Haoran Wei, Huan Lin, Jialong Tang, Jialin Wang, Jian Yang, Jianhong Tu, Jianwei Zhang, Jianxin Ma, Jianxin Yang, Jin Xu, Jingren Zhou, Jinze Bai, Jinzheng He, Junyang Lin, Kai Dang, Keming Lu, Keqin Chen, Kexin Yang, Mei Li, Mingfeng Xue, Na~Ni, Pei Zhang, Peng Wang, Ru~Peng, Rui Men, Ruize Gao, Runji Lin, Shijie Wang, Shuai Bai, Sinan Tan, Tianhang Zhu, Tianhao Li, Tianyu Liu, Wenbin Ge, Xiaodong Deng, Xiaohuan Zhou, Xingzhang Ren, Xinyu Zhang, Xipin Wei, Xuancheng Ren, Xuejing Liu, Yang Fan, Yang Yao, Yichang Zhang, Yu~Wan, Yunfei Chu, Yuqiong Liu, Zeyu Cui, Zhenru Zhang, Zhifang Guo, and Zhihao Fan.
\newblock Qwen2 technical report, 2024.
\newblock URL \url{https://arxiv.org/abs/2407.10671}.

\bibitem[Yu et~al.(2019)Yu, Xu, Yu, Yu, Zhao, Zhuang, and Tao]{yu2019activityqa}
Zhou Yu, Dejing Xu, Jun Yu, Ting Yu, Zhou Zhao, Yueting Zhuang, and Dacheng Tao.
\newblock Activitynet-qa: A dataset for understanding complex web videos via question answering.
\newblock In \emph{AAAI}, pages 9127--9134, 2019.

\bibitem[Yue et~al.(2024)Yue, Zheng, Zhang, and Chen]{yue2024mammoth2}
Xiang Yue, Tuney Zheng, Ge~Zhang, and Wenhu Chen.
\newblock Mammoth2: Scaling instructions from the web.
\newblock \emph{Advances in Neural Information Processing Systems}, 2024.

\bibitem[Zellers et~al.(2019)Zellers, Holtzman, Bisk, Farhadi, and Choi]{zellers2019hellaswag}
Rowan Zellers, Ari Holtzman, Yonatan Bisk, Ali Farhadi, and Yejin Choi.
\newblock {HellaSwag}: Can a machine really finish your sentence?
\newblock In \emph{Proceedings of the 57th Annual Meeting of the Association for Computational Linguistics}, 2019.

\bibitem[Zhang et~al.(2022)Zhang, Roller, Goyal, Artetxe, Chen, Chen, Dewan, Diab, Li, Lin, Mihaylov, Ott, Shleifer, Shuster, Simig, Koura, Sridhar, Wang, and Zettlemoyer]{zhang-el-al:2022:OPT}
Susan Zhang, Stephen Roller, Naman Goyal, Mikel Artetxe, Moya Chen, Shuohui Chen, Christopher Dewan, Mona Diab, Xian Li, Xi~Victoria Lin, Todor Mihaylov, Myle Ott, Sam Shleifer, Kurt Shuster, Daniel Simig, Punit~Singh Koura, Anjali Sridhar, Tianlu Wang, and Luke Zettlemoyer.
\newblock Opt: Open pre-trained transformer language models, 2022.

\bibitem[Zhang et~al.(2025)Zhang, Luo, Yuan, and Yao]{zhang2025autonomous}
Yifan Zhang, Yifan Luo, Yang Yuan, and Andrew Chi-Chih Yao.
\newblock Autonomous data selection with zero-shot generative classifiers for mathematical texts.
\newblock \emph{The 63rd Annual Meeting of the Association for Computational Linguistics (ACL 2025 Findings)}, 2025.

\bibitem[Zhang et~al.(2024)Zhang, Rando, Evtimov, Chi, Smith, Carlini, Tramèr, and Ippolito]{zhang2024persistentpretrainingpoisoningllms}
Yiming Zhang, Javier Rando, Ivan Evtimov, Jianfeng Chi, Eric~Michael Smith, Nicholas Carlini, Florian Tramèr, and Daphne Ippolito.
\newblock Persistent pre-training poisoning of llms, 2024.
\newblock URL \url{https://arxiv.org/abs/2410.13722}.

\bibitem[Zheng et~al.(2023)Zheng, Chiang, Sheng, Zhuang, Wu, Zhuang, Lin, Li, Li, Xing, Zhang, Gonzalez, and Stoica]{llm_as_judge}
Lianmin Zheng, Wei-Lin Chiang, Ying Sheng, Siyuan Zhuang, Zhanghao Wu, Yonghao Zhuang, Zi~Lin, Zhuohan Li, Dacheng Li, Eric~P. Xing, Hao Zhang, Joseph~E. Gonzalez, and Ion Stoica.
\newblock Judging llm-as-a-judge with mt-bench and chatbot arena.
\newblock In \emph{Proceedings of the 37th International Conference on Neural Information Processing Systems}, NIPS '23, Red Hook, NY, USA, 2023. Curran Associates Inc.

\bibitem[Zhou et~al.(2023)Zhou, Lu, Mishra, Brahma, Basu, Luan, Zhou, and Hou]{zhou2023instructionfollowingevaluationlargelanguage}
Jeffrey Zhou, Tianjian Lu, Swaroop Mishra, Siddhartha Brahma, Sujoy Basu, Yi~Luan, Denny Zhou, and Le~Hou.
\newblock Instruction-following evaluation for large language models, 2023.
\newblock URL \url{https://arxiv.org/abs/2311.07911}.

\bibitem[Ziemski et~al.(2016)Ziemski, Junczys-Dowmunt, and Pouliquen]{ziemski2016united}
Micha{\l} Ziemski, Marcin Junczys-Dowmunt, and Bruno Pouliquen.
\newblock The united nations parallel corpus v1. 0.
\newblock In \emph{Proceedings of the Tenth International Conference on Language Resources and Evaluation (LREC'16)}, pages 3530--3534, 2016.

\end{thebibliography}
